\documentclass{macro/neu_msthesis}

\usepackage{physics}
\usepackage{siunitx}
\usepackage{graphics} % for pdf, bitmapped graphics files
\usepackage{graphicx} % for epx
\usepackage{epsfig} % for postscript graphics files
\usepackage{lipsum} 
\usepackage{amsmath, bm} % assumes amsmath package installed
\usepackage{amssymb}  % assumes amsmath package installed
\usepackage{upgreek}
\usepackage{url}
\usepackage{caption}
\usepackage{subcaption}
\usepackage{pgfplots}
\usepackage{floatrow}
\usepackage{float}
\usepackage{multicol}
\usepackage{tikz}
\usepackage{booktabs}
\usepackage{longtable}
\usepackage{algorithm}
\usepackage{algpseudocode}
\usepackage{comment}
\usetikzlibrary{shapes, arrows}
\usepackage{amsmath}
\newfloatcommand{capbtabbox}{table}[][\FBwidth]

\title{Traversability Aware Autonomous Navigation for Multi-Modal Mobility Morphobot (M4)}

\author{Hrigved Mahesh Suryawanshi}

\dept{Electrical and Computer Engineering}

\degree{Masters of Science}

\degreename{Robotics}

\field{Robotics}

\submitdate{December 2025}

\numberofmembers{3}

\principaladviser{Dr. Alireza Ramezani}
\firstreader{Dr. Aksaray Derya}
\secondreader{Dr. Michael Everett}
\chairman{Dr. xxx}

\dean{Dr. xxx}

\usepackage{amsfonts}			% special math characters
\usepackage{times}		% use Postscript fonts

\newcommand{\ifno}[1]{}

\usepackage{multirow}
\makeindex
\usepackage{makeidx}
\usepackage{acronym}

\usepackage{url}
\usepackage{graphicx}
\ifnum\pdfoutput>0
\usepackage[pdftex,%                    % hyper-references for ps2pdf
bookmarks=true,%                        % generate bookmarks ...
bookmarksnumbered=true,%                % ... with numbers
hypertexnames=false,%                   % needed for correct links to figures
breaklinks=true           %Allows link text to break across lines;
]{hyperref}
\else
\usepackage[hypertex,%                  % hyper-references for ps2pdf
bookmarks=true,%                        % generate bookmarks ...
bookmarksnumbered=true,%                % ... with numbers
hypertexnames=false,%                   % needed for correct links to figures
breaklinks=true           %Allows link text to break across lines;
]{hyperref}
\fi

\hypersetup{				% setup PDF information fields
pdfauthor = {\authorRef},
pdftitle = {\titleRef},
pdfsubject = {\expandafter{\degreeRef} thesis submitted to Northeastern University},
pdfkeywords = {Multimodal robots,M4,Traversability,3D path planning}
pdfcreator = {LaTeX with hyperref package},
pdfproducer = {dvips + ps2pdf}}

\begin{document}

% add a pdf bookmark to the cover page
\pdfbookmark[1]{Cover}{cover}

% --- title page ---
\titlepage

% --- front matter ---
\begin{frontmatter}
% print signature page
%\signaturepage
% dedication
% dedication.tex:

\begin{dedication}
To my family.
\end{dedication}

% table of content (add bookmark for convenience)
\pdfbookmark[1]{Table of Contents}{contents}
\tableofcontents
\listoffigures
\newpage\ssp
\listoftables

% include a list of Acronyms (comment out if no acronyms are specified)
%\input{tex/acronyms.tex}

% include any of the front matter files that contain text
% attention the input does cause a page break, the include on 
% the other hand does not
% acknowledgements.tex:

\begin{acknowledgements}

I extend my heartfelt gratitude to all the members of the SiliconSynapse Lab for creating a collaborative and supportive research environment throughout this work. I am especially thankful to Adarsh Salagame, who served as my mentor in the lab and provided invaluable guidance on technical challenges and research direction. I also thank Kevin Wu, my appointed thesis mentor, for his consistent support and feedback that helped shape this thesis. 
I express my sincere appreciation to Prof. Alireza Ramezani, my principal adviser, for his mentorship and for providing me the opportunity to work on this project. I would like to thank my labmate, Henry Noyes, for his help with various aspects of the work, and Bibek Gupta for his assistance whenever needed. A special thanks to my good friend Irfan Ali for always being there to help with doubts and for the many valuable discussions. Finally, I am deeply grateful to my girlfriend Prajoti and my loving family for their unwavering support, patience, and encouragement throughout this journey. This work would not have been possible without them.

\end{acknowledgements}
% abstract.tex:
\begin{abstract}

Autonomous navigation in unstructured environments requires robots to assess terrain difficulty in real-time and plan paths that balance efficiency with safety. This thesis presents a traversability-aware navigation framework for the M4 robot platform that uses learned terrain analysis to generate energy-efficient paths avoiding difficult terrain.Our approach uses FAST-LIO for real-time localization, generating 2.5D elevation maps from LiDAR point clouds. A CNN-based model processes these elevation maps to estimate traversability scores, which are converted into navigation costs for path planning. A custom A* planner incorporates these costs alongside geometric distance and energy consumption to find paths that trade modest distance increases for substantial terrain quality improvements. Before system development, a platform-agnostic study compared LiDAR-based and camera-based SLAM using OptiTrack ground truth. Point cloud comparison through ICP alignment and cloud-to-mesh distance analysis demonstrated that LiDAR-based mapping achieves centimeter-level precision essential for elevation mapping, while camera-based approaches exhibited significantly higher geometric error. These findings directly resulted in the selection of LiDAR as the primary sensor to generate elevation maps. The complete pipeline integrates FAST-LIO localization, GPU-accelerated elevation mapping, CNN-based traversability estimation, and Nav2 navigation with a custom traversability-aware planner. Experimental results demonstrate that the system successfully avoids low traversability regions and accepts a few longer paths to achieve a reduction in terrain cost. This work establishes a foundation for intelligent terrain-aware navigation applicable to multi-modal robotic platforms.

\end{abstract}

\end{frontmatter}

% --- body of the document ---

%\pagestyle{plain}
\pagestyle{headings}

% include each chapter like below
\chapter{Introduction}
\label{chap:introduction}

\section{Background and Motivation}
\label{sec:background-motivation}

Autonomous mobile robots are increasingly being deployed in environments that are hazardous, remote, or simply not feasible for continuous human operation. Applications ranging from disaster response and industrial inspection to planetary exploration ask for a long-term autonomy solution under uncertain terrain and sensing conditions. However, a central drawback of most field robots is that they rely on a single locomotion mode, or if multimodal, they don't efficiently know when to change modes. Wheeled platforms are energy-efficient and stable on smooth surfaces, but easily defeated by stairs, deep rubble, loose soil, or steep slopes. Aerial robots bypass many geometric obstacles, but with an enormous energy penalty: flight endurance on the M4 scale is typically measured in minutes rather than hours, and safe takeoff and landing require additional clearance and control overhead. These complementary weaknesses motivate platforms that can adapt their locomotion to the environment rather than forcing the environment to match a single mobility strategy.

Several research groups have explored ground-aerial hybrid platforms to address these challenges. LEONARDO demonstrated bipedal walking combined with thrust-assisted balance and flight capabilities~\cite{kim2021bipedal}, while MTABot introduced an efficient morphable design using two transformable wheels for terrestrial-aerial transitions~\cite{shi2024mtabot}. More recently, ATMO presented dynamic ground-aerial transition capabilities through aerially transforming morphology~\cite{mandralis2025atmo}. These platforms demonstrate growing interest in multimodal locomotion, though each faces unique trade-offs between payload capacity, energy efficiency, and mechanical complexity.

This thesis builds on that motivation through the Multi-Modal Mobility Morphobot (M4), originally developed at Caltech's Center for Autonomous Systems and Technologies~\cite{sihite2023multi}. M4 is a bio-inspired ground–aerial platform whose mechanical design is rooted in the repurposing of appendages similar to birds such as Chukars and Hoatzins that use the same limbs for climbing, balancing, and flight. M4 contains four transformable appendages that can act as wheels, legs, or propellers that generate thrust, enabling multiple locomotion configurations or modes within a compact, mass-efficient system. Unlike many hybrid robots that carry isolated actuation chains for each mode, M4 reuses its structures to avoid the redundancy and payload costs that typically narrow the real-world usefulness of multi-modal systems.

\begin{figure}[htbp]
\centering
\includegraphics[width=0.95\textwidth]{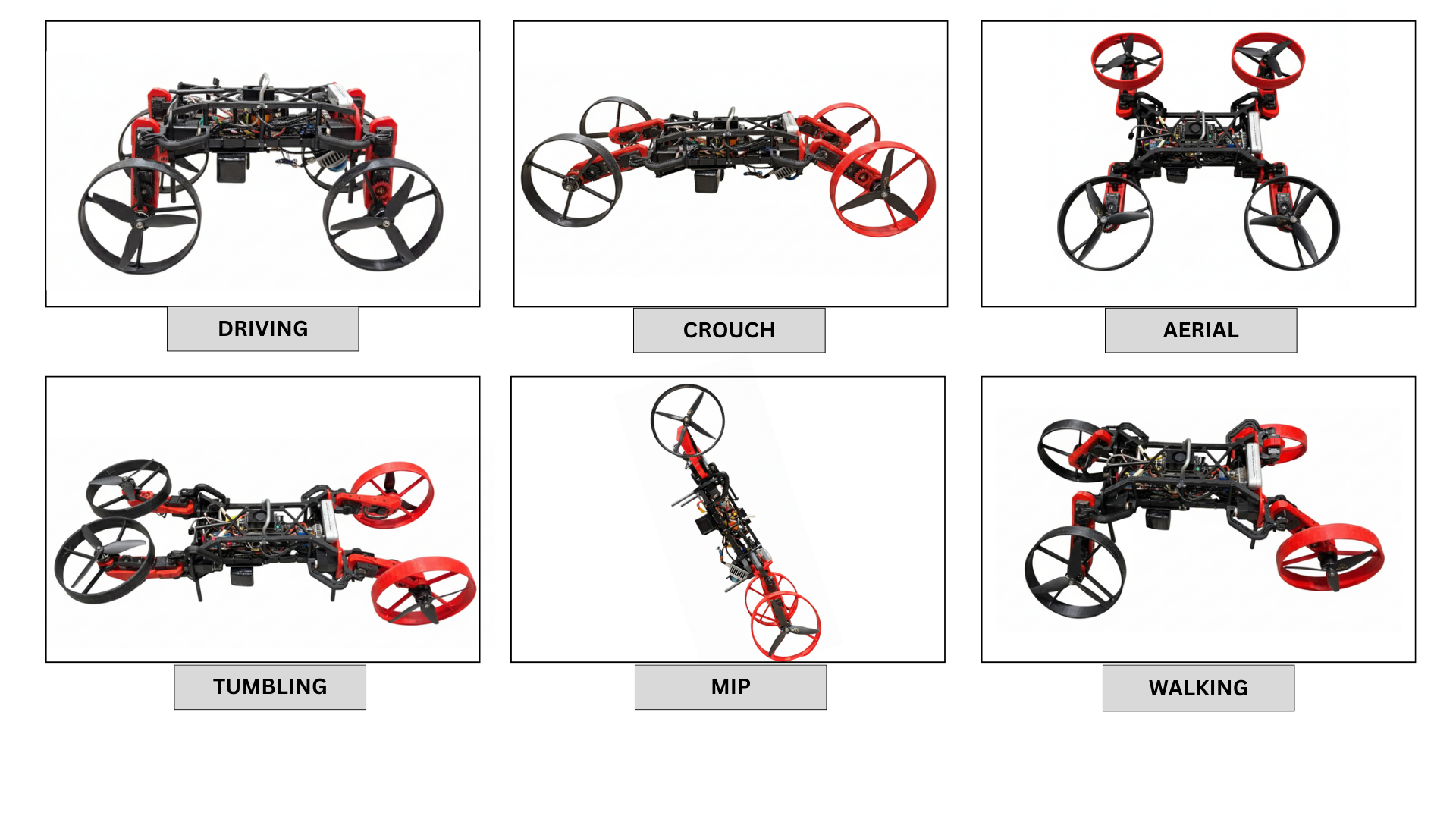}
\caption{Illustrates M4 Locomotion modes.~\cite{sihite2023multi}} 
\label{fig:m4-original}
\end{figure}

A signature aspect of the robot is that it has a shrouded propeller and wheel mechanism where each appendage houses a propeller for thrust, while the shroud itself acts as a driven wheel via external gearing. This provides for hybrid maneuvers such as thruster-assisted rolling and wing-assisted incline running and pushes the platform beyond a simple "drive or fly" structure. Such mobility is compelling in planetary missions, where rovers have suffered major setbacks from terrain traps and traction loss. The Perseverance–Ingenuity mission has demonstrated the value of combining ground and aerial assets~\cite{farley2020mars, tzanetos2022ingenuity}, but these systems remain physically separate and manually coordinated. A unified platform such as M4 provides a path toward continuous autonomy without reliance on multi-vehicle planning or human-in-the-loop orchestration.

Previous work on M4 focused on establishing the mechanical platform and exploring multi-modal navigation. Sihite et al.~\cite{sihite2023multi} demonstrated autonomous 3D path planning using SLAM and multi-modal probabilistic roadmaps, showing the robot flying over obstacles and transitioning between ground and aerial modes. Rajput~\cite{rajput2023m4} demonstrated energy-aware multi-modal planning using a 2.5D terrain representation and a learned traversability model that drove mode switching in Gazebo simulation. More recently, Gherold et al.~\cite{gherold2025self} introduced self-supervised cost of transport estimation for multimodal path planning, further advancing terrain-aware decision making. Although these works validated the idea of terrain-guided decisions, they assumed idealized sensing and ground-truth state estimates. Physical deployment introduces different realities: LiDAR and IMU data are noisy, calibration errors and motion blur cause gaps in elevation maps, and localization in GPS-denied settings must be handled onboard rather than inherited from simulation.

The M4 platform has since been rebuilt at Northeastern University, with new sensors including Livox MID-360 LiDAR, an RGB camera, and embedded compute integration with Jetson Orin, but without a field-ready autonomy stack. This thesis is thus motivated not only to demonstrate the capability of multi-modal mobility in principle but also to provide the perception and mapping decision pipeline required for reliable deployment in unstructured environments. In practice, the work focuses on robust terrain understanding, real-time elevation mapping, learned traversability prediction, and integrating these components into a navigation system that selects between ground driving and aerial bypass when the latter becomes necessary.

\begin{figure}[htbp]
\centering
\includegraphics[width=0.75\textwidth]{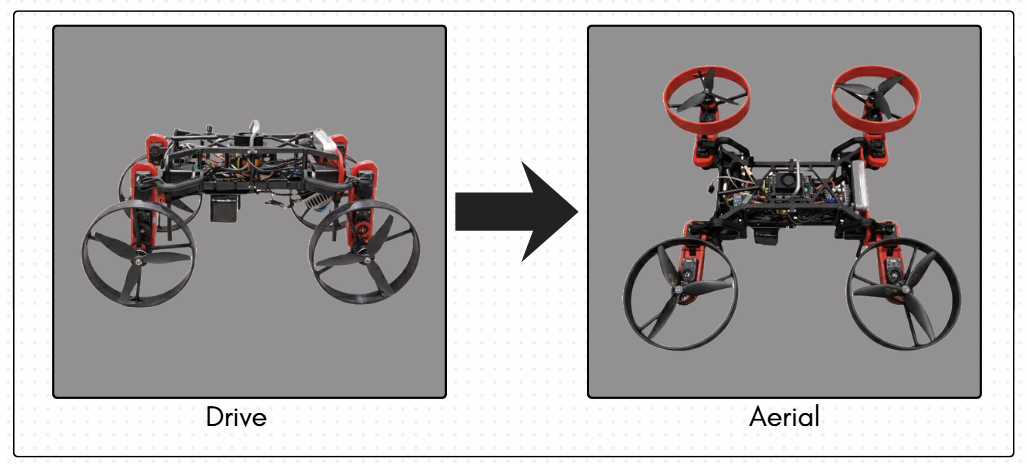}
\caption{illustrates the Transformation between wheeled ground locomotion and aerial flight modes.}
\label{fig:m4-rebuilt}
\end{figure}

\section{Defining Traversability}
\label{sec:defining-traversability}

The term ``traversability'' appears frequently in mobile robotics literature, yet its meaning varies considerably across different research communities and applications. Before proceeding with the technical contributions of this thesis, it is worth examining what traversability actually means and how this work interprets the concept.

\subsection{Origin and Evolving Perspectives}

The concept of terrain traversability was formally introduced to the robotics community by Seraji in 1999~\cite{seraji1999traversability}, who proposed the ``Traversability Index'' for planetary rovers at NASA's Jet Propulsion Laboratory. This early work defined traversability through fuzzy logic rules based on terrain characteristics such as roughness, slope, and discontinuity. Howard and Seraji~\cite{howard2001vision} later extended this to vision-based systems, establishing traversability as a measurable quantity derived from sensor observations.

Since then, the definition of traversability has evolved along multiple directions. Some researchers treat it as a binary classification: terrain is either traversable or not, similar to traditional obstacle detection. Others define it as a continuous measure reflecting the ``ease of traversal'' across a region. More recent work has framed traversability in terms of semantic terrain classes (grass, gravel, asphalt), learned features from self-supervised navigation experience, or even proprioceptive signals like vibration and wheel slip. The diversity of definitions reflects the fact that traversability is not a property of the terrain alone; it emerges from the interaction between a specific robot and a specific environment.

In this thesis, traversability is understood as the ability of a particular robot to navigate across a particular region of terrain, given what that robot can physically do. This definition is inherently robot-specific: the same terrain can have vastly different traversability values for different platforms.

Consider a staircase. For a legged robot capable of climbing steps, stairs represent highly traversable terrain; the robot can ascend or descend with relative ease. For a wheeled robot like M4 in ground mode, the same staircase is completely impassable; no amount of path planning will allow the wheels to climb discrete steps that exceed the platform's ground clearance. The terrain itself has not changed, but its traversability depends entirely on the capabilities of the robot evaluating it.

This robot-centric view extends beyond binary passability. Even among regions that a robot \textit{can} traverse, some are easier than others. A flat concrete floor and a grassy slope may both be traversable for M4's wheeled mode, but navigating the slope requires more energy, risks wheel slip, and places greater stress on the motors. Traversability, therefore, is not just about whether the robot can go somewhere, but how easily and efficiently it can do so.

\subsection{Traversability for Comparison and Decision Making}

Beyond labeling regions as traversable or not, traversability serves a comparative function in navigation. When multiple paths exist between a start and goal location, the planner must decide which path to take. If both paths are geometrically feasible, the question becomes: which path is \textit{more} traversable? Which requires less effort, less energy, less risk?

This comparative aspect is central to how traversability is used in this thesis. The CNN-based traversability estimator assigns a continuous score $T \in [0, 1]$ to each cell of the elevation map. Higher values indicate terrain that the M4 robot can traverse more easily; lower values indicate terrain that is more difficult, risky, or potentially impassable. These scores are then converted into navigation costs, allowing the path planner to optimize not just for geometric distance, but for traversability along the entire route.

In essence, traversability answers two questions for the M4 navigation system:
\begin{enumerate}
    \item \textbf{Can the robot go there?} If traversability is below a critical threshold, the region is treated as an obstacle for ground mode navigation.
    \item \textbf{How hard is it to go there?} Among traversable regions, the relative traversability scores guide the planner toward paths that are smoother, safer, and more energy-efficient.
\end{enumerate}

For a multimodal robot like M4, traversability also informs mode selection. When ground traversability along all candidate paths falls below acceptable levels, the decision framework considers whether aerial locomotion offers a better alternative, trading the energy cost of flight against the difficulty or impossibility of ground traversal.

\section{Overview of the Thesis}
\label{sec:thesis-overview}

This thesis develops a complete autonomous navigation framework for the physical M4 platform centered on terrain and traversability-aware decision making. The central premise is that intelligent path choosing and mode avoiding difficult paths continuously, along with predictive evaluation of terrain and locomotion difficulty, rather than an isolated geometry-based pipeline. M4 therefore uses LiDAR-based terrain mapping, a CNN traversability estimator, and a planner that weighs ground path feasibility against the energy cost of aerial bypass.

The work proceeds in two layers. First, foundational perception validation is performed to build confidence in the terrain representation required for traversability estimation. This began with a platform-agnostic study comparing point clouds generated from two independent sensing pipelines, a vision-based depth reconstruction and a LiDAR-based mapping method~\cite{salagame2025crater}. By benchmarking against motion-capture ground truth, we verified that LiDAR produced more consistent geometry, fewer surface discontinuities, and lower drift, making it reliable for constructing accurate elevation maps. These results established that LiDAR-centric elevation mapping would provide the stable geometric substrate needed for downstream terrain reasoning.

Second, the main system contribution integrates elevation mapping, learned traversability, and navigation into an autonomy stack that executes on the M4 platform in real time. Traversability scores are mapped into navigation costs for the ground planner, encouraging safe and energy-efficient path planning. When no feasible or cost-effective ground path exists, the decision module triggers aerial mode to bypass the difficult region.

The remainder of this document is organized as follows. Chapter~\ref{chap:literature-review} presents the literature review covering multimodal robotics, traversability estimation, elevation mapping, and terrain-aware navigation. Chapter~\ref{chap:PerceptionValidation} presents the platform-agnostic perception study, where point clouds generated from independent sensing systems are compared and validated for use in elevation mapping. Chapter~\ref{chap:SystemArchitecture} details the full autonomy pipeline including the M4 platform, elevation map construction, CNN-based traversability inference, costmap generation, and the mode-selection logic. Chapter~\ref{chap:results} presents experimental evaluation of the perception pipeline and navigation behavior, and Chapter~\ref{chap:results} concludes with a summary of contributions and directions for future work.

% Path planning
\chapter{Literature Review}
\label{chap:literature-review}

Autonomous navigation for multimodal ground--aerial robots sits at the intersection of hybrid mobility design, terrain perception, traversability reasoning, localization, and energy-aware planning. Unlike single-mode systems, multimodal robots must decide not only \textit{where to go} but also \textit{how to move}, balancing endurance, safety, and feasibility across distinct locomotion regimes. This chapter surveys the most relevant work that informs the autonomy stack developed for M4. We organize the discussion around five themes: (1) multimodal robots and hybrid mobility systems; (2) terrain perception, traversability estimation, and elevation mapping; (3) LiDAR--inertial localization and mapping; (4) energy-aware and multi-objective planning for hybrid platforms; and (5) integration of perception and planning within navigation frameworks.

\section{Multimodal Robots and Hybrid Mobility Systems}
\label{sec:lr-multimodal}

The primary motivation for multimodal and morphing robots emerges from the shortcomings of single-mode mobility. Ground robots may offer stable contact-based locomotion and long endurance, but often fail on highly cluttered, discontinuous, or deformable terrain such as rubble, sand, loose gravel, or steep step fields. Aerial robots circumvent many of these geometric constraints, yet flight remains energetically expensive and payload-limited, especially for platforms in the sub-10 kg class~\cite{sihite2023multi}. Hybrid systems thus attempt to combine ground efficiency with aerial reach, enabling continuous navigation across heterogeneous terrain.

Early ground--aerial hybrids largely relied on mechanically distinct ``drive then fly'' paradigms. The HyTAQ platform embedded a quadrotor within a protective rolling cage, enabling energy-efficient ground travel with opportunistic flight for obstacles or gaps~\cite{kalantari2014hytaq}. More recent platforms such as MTABot introduced efficient morphable terrestrial-aerial designs using transformable wheels~\cite{shi2024mtabot}. These systems demonstrated the feasibility of hybrid locomotion but often required pre-scripted transition logic rather than terrain-driven autonomy.

\begin{figure}[htbp]
    \centering
    \includegraphics[width=0.95\textwidth]{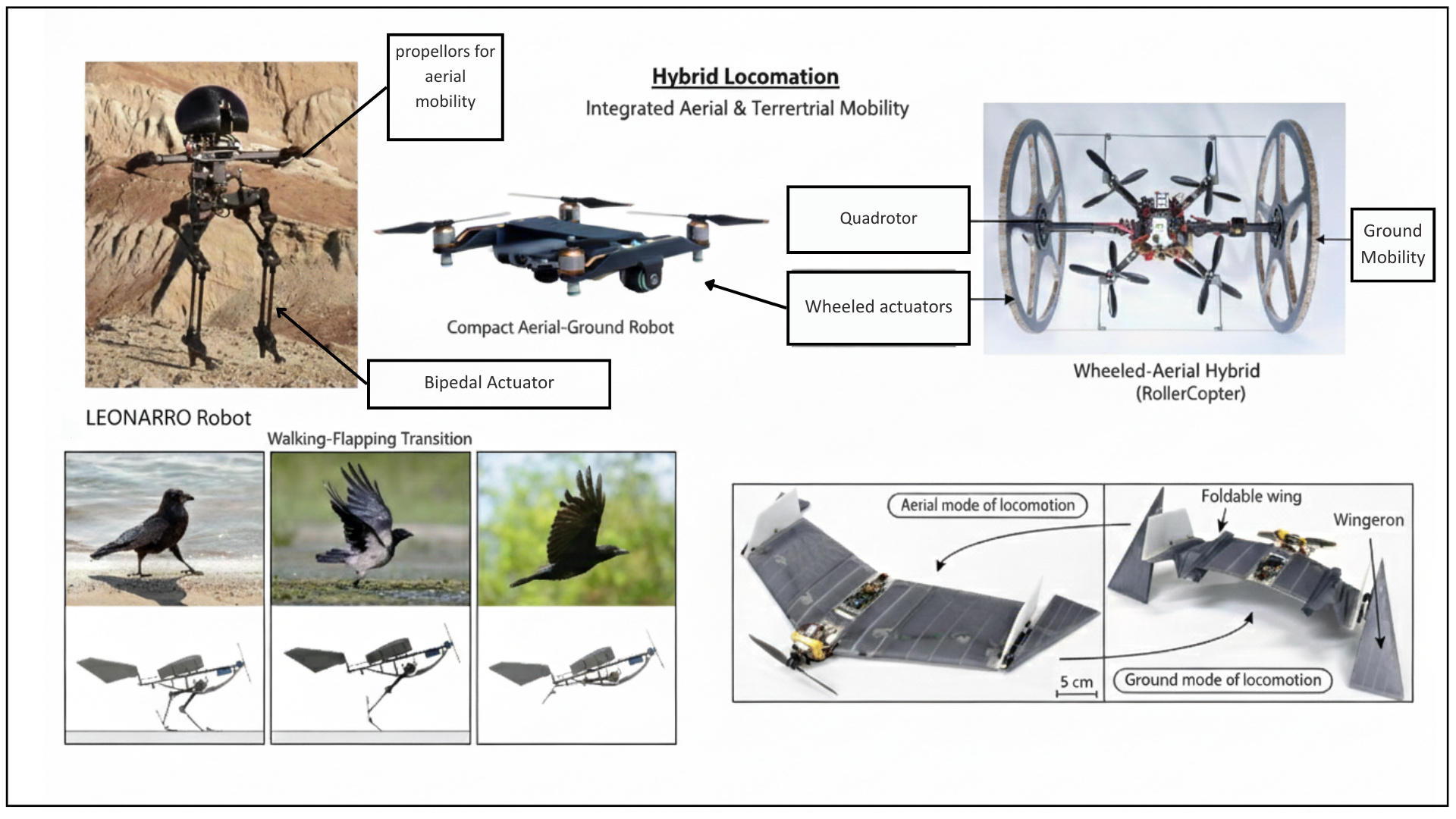}
    \caption{Examples of multi-modal ground-aerial robot platforms: LEO~\cite{kim2021bipedal}, DALER~\cite{daler2015bioinspired}, RAVEN~\cite{shin2024fast}, Duawlfin~\cite{tang2025duawlfin}, and Rollocopter~\cite{sabet2019rollocopter}.}
    \label{fig:multimodal-robots}
\end{figure}

Beyond rolling systems, multimodality has also been explored through walking--flying or legged--aerial combinations. Caltech's LEONARDO demonstrated dynamic bipedal walking paired with flight stabilization, enabling behaviors such as slacklining and hybrid locomotion over discontinuous terrain~\cite{kim2021bipedal}. Bio-inspired platforms have shown that aerial thrust can extend terrestrial mobility through assisted climbing and balancing. Ramezani et al.\ developed a biomimetic robotic platform to study flight specializations of bats, demonstrating sophisticated aerial maneuvers inspired by biological systems~\cite{ramezani2017biomimetic}. The Aerobat platform further explored integrated mechanical intelligence and control for bio-inspired flight~\cite{sihite2021integrated, sihite2020enforcing}. Thruster-assisted legged locomotion has been extensively studied, with work on unilateral ground contact force regulations~\cite{sihite2021unilateral, liang2021rough} and capture point control for bipedal systems~\cite{pitroda2024capture}.

More recently, ATMO demonstrated dynamic ground-aerial transitions through aerially transforming morphology~\cite{mandralis2025atmo}. Work on dynamic quadrupedal legged and aerial locomotion via structure repurposing has shown the potential for unified platforms that leverage appendage multifunctionality~\cite{wang2025dynamic}. Thruster-assisted incline walking using reduced-order models~\cite{krishnamurthy2025thruster} and narrow-path dynamic walking with integrated posture manipulation~\cite{krishnamurthy2024narrow} have advanced the control of hybrid legged-aerial systems on challenging terrain.

Mode switching has also been studied from a planning and control perspective. Energy-aware planning for multimodal systems requires reasoning about transition costs and mode-specific capabilities. Sihite et al.\ demonstrated efficient path planning using integrated probabilistic roadmaps (PRM) and reference governors for multi-modal legged-aerial locomotion~\cite{sihite2022efficient}. Posture manipulation during dynamic maneuvers, including wall-jumping~\cite{sihite2024posture} and tumbling for heading control~\cite{salagame2024dynamic, salagame2024heading}, enables robust transitions between locomotion modes.

In planetary and field exploration contexts, where terrain uncertainty is high, hybrid mobility is particularly valuable. Thruster-assisted legged mobility has been proposed for Mars exploration~\cite{ramezani2022thruster}, and morphing micro aerial vehicle designs offer efficient aerial mobility in thin atmospheres~\cite{ramezani2022efficient}. The Perseverance--Ingenuity mission demonstrated the value of combining ground and aerial assets~\cite{farley2020mars, tzanetos2022ingenuity}, though these systems remain physically separate and manually coordinated. These studies reinforce that multimodal autonomy must be guided by terrain understanding rather than solely by geometric obstacle triggers.

The M4 design builds directly on these motivations. Its appendage repurposing architecture enables reconfiguration without redundant actuators, improving mass efficiency and reducing the mechanical overhead that limits many hybrid designs~\cite{sihite2023multi}. Prior work on M4 established the mechanical system and explored simulation-based multimodal planning. Sihite et al.\ demonstrated autonomous 3D path planning on the M4 platform using SLAM and multi-modal probabilistic roadmaps~\cite{sihite2023demonstrating}. Dynamic modeling of wing-assisted inclined running further expanded M4's locomotion repertoire~\cite{sihite2024dynamic}. This body of work motivates the specific autonomy gap addressed in this thesis: bringing multimodal planning from simulation into real-world deployment through robust terrain perception and traversability-driven decision making.

\section{Terrain Perception, Traversability Estimation, and Elevation Mapping}
\label{sec:lr-trav}

Accurate terrain understanding forms the backbone of autonomous navigation in unstructured environments. For M4, terrain perception is critical not only for safe ground planning, but also for deciding when ground locomotion becomes inefficient or infeasible, warranting an aerial bypass. Traversability provides the metric that links terrain perception to decision-level autonomy.

Traversability is inherently \textit{robot-specific}: it quantifies the feasibility or cost of moving through a region given a robot's drivetrain, clearance, stability limits, and contact mechanics. A mild grass slope can be safe for a large rover but destabilizing for a smaller platform; loose gravel may be benign for tracked robots yet produce severe slip for wheels~\cite{howard2001vision, papadakis2013traversability}. As emphasized by modern surveys, traversability estimates must incorporate both geometric and physical terrain properties relative to robot capability. Waibel et al.\ addressed the question of terrain roughness estimation for both local and global path planning~\cite{waibel2022rough}.

\subsection*{Geometric Traversability and Classical Terrain Reasoning}

Classic approaches to traversability rely on geometric cues extracted from range sensors. Early work by Howard and Seraji used vision-derived terrain parameters to compute slope and roughness-based feasibility scores~\cite{howard2001vision, seraji2001traversability}. These methods typically discretize the environment into grid cells and compute local features such as slope, curvature, roughness, and obstacle height, then compare them against robot-dependent thresholds.

\begin{figure}[htbp]
    \centering
    \includegraphics[width=0.85\textwidth]{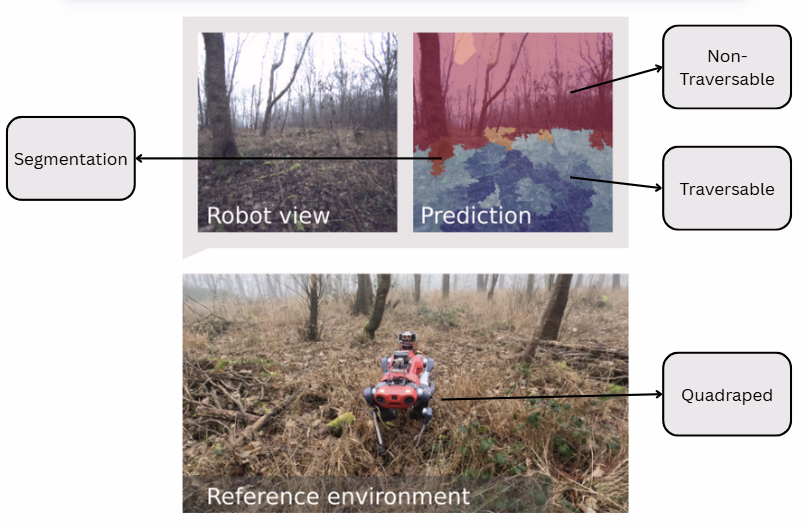}
    \caption{Example of vision-based traversability estimation~\cite{frey2023fast}.}
    \label{fig:vision-trav}
\end{figure}

While interpretable and efficient, geometric methods require manual tuning and struggle when terrain material dominates geometric structure. Multiple surveys identify brittleness as a recurring limitation, particularly in real terrains with mixed substrates~\cite{papadakis2013traversability}.

\subsection*{Vision-Based Traversability and Semantic Terrain Classification}

Vision-based methods augment geometry by using appearance cues to infer terrain type and physical properties. Learning-based approaches use RGB or RGB-D imagery for terrain classification, often mapping textures to classes like pavement, grass, mud, gravel, or water. Valada and Burgard demonstrated deep spatiotemporal models for robust proprioceptive terrain classification~\cite{valada2017deep}. Self-supervised frameworks further relate appearance to observed traversal outcomes, allowing robots to learn traversability directly from experience. Z{\"u}rn et al.\ combined self-supervised visual terrain classification with unsupervised acoustic feature learning~\cite{zurn2020self}. Frey et al.\ developed fast traversability estimation for wild visual navigation~\cite{frey2023fast}.

\begin{comment}
\begin{figure}[htbp]
    \centering
    \includegraphics[width=0.55\textwidth]{figs/wayfaster.png}
    \caption{WayFASTER traversability estimation model architecture for real-time terrain analysis~\cite{gasparino2024wayfaster}.}
    \label{fig:wayfaster}
\end{figure}
\end{comment}
Vision methods are particularly valuable for distinguishing materials that geometric profiles cannot, such as wet mud versus dry soil or shallow water versus solid ground. Yet their sensitivity to lighting, shadows, and occlusions remains a practical limitation in field deployments, motivating multi-modal sensing and elevation-informed learning~\cite{frey2023fast}.

\subsection*{Learning-Based Traversability from Elevation}

Deep learning has increasingly been used to estimate traversability from elevation maps or 3D terrain representations. CNN-based estimators treat a local elevation patch as an image-like input and output a continuous traversability score, learning nonlinear terrain features beyond hand-engineered geometry. WayFAST demonstrated predictive traversability for outdoor navigation, showing strong generalization across unseen terrain and improved global planner performance~\cite{gasparino2022wayfast}. WayFASTER extended this with self-supervised prediction for increased navigation awareness~\cite{gasparino2024wayfaster}.

Simulation-to-real transfer for learned traversability has shown promise, where terrain labels are derived from simulation but adapted using real-world driving outcomes. Chavez-Garcia et al.\ demonstrated learning ground traversability from simulations~\cite{chavez2018learning}. Vecchio et al.\ combined self-supervised learning with unsupervised domain adaptation on synthetic data for terrain traversability prediction~\cite{vecchio2024terrain}.

For M4, using elevation-guided learning is especially appropriate because LiDAR provides a lighting-invariant geometric basis for traversability prediction. A CNN can then infer robot-specific difficulty from spatial patterns linked to slope onset, step structures, or roughness clusters, traits that are hard to threshold robustly across heterogeneous real terrain.

\subsection*{Proprioceptive Traversability and Reactive Adaptation}

Proprioceptive methods estimate traversability from the robot's internal sensor response during motion. IMU roll/pitch dynamics, wheel slip, or motor current spikes provide strong signals of terrain difficulty, particularly in deformable ground or high-resistance regions. Learning quadrupedal locomotion on deformable terrain has shown how proprioceptive feedback enables adaptation to challenging substrates~\cite{choi2023learning}. High-speed control and navigation for quadrupedal robots on complex terrain further demonstrates the role of proprioception in dynamic locomotion~\cite{kim2025high}. For wheeled-legged robots, learning robust autonomous navigation and locomotion combines multiple sensing modalities~\cite{lee2024learning}.

Such systems excel at online adaptation but are reactive: the robot must experience difficulty before classifying it. For multimodal switching, predictive traversability (before committing to a path) is preferable, though proprioceptive feedback remains valuable for confirming model predictions and refining confidence. Snake robots with tactile perception have demonstrated navigation on large-scale challenging terrain using reactive sensing~\cite{jiang2024snake}, while hierarchical RL-guided approaches enable large-scale navigation~\cite{jiang2024hierarchical}.

\subsection*{Elevation Mapping as the Terrain Substrate}

Elevation maps provide the practical 2.5D terrain representation that supports both traversability estimation and planning. Unlike full volumetric occupancy grids, elevation maps store a height scalar (often with variance) per horizontal cell, retaining the terrain surface structure most relevant for ground robots while remaining computationally tractable~\cite{fankhauser2014elevation, miki2022elevation}. Fankhauser et al.\ introduced robot-centric elevation mapping with uncertainty propagation, enabling stable updates despite pose drift~\cite{fankhauser2014elevation}. A universal grid map library was developed for rough terrain navigation applications~\cite{fankhauser2016grid}. GPU-accelerated variants allow real-time updates over large local windows, supporting dynamic replanning~\cite{miki2022elevation}.

Foundational terrain mapping work also includes octree-based 3D representations such as OctoMap, which introduced probabilistic occupancy fusion for LiDAR-based mapping~\cite{hornung2013octomap}. These methods collectively motivate the multi-layer elevation representation used in M4, which stores height, variance, temporal confidence, and CNN traversability output within a unified grid.

Taken together, the literature indicates a clear trajectory: geometric terrain reasoning provides interpretability and efficiency, but learning-based traversability from elevation maps offers robustness and generalization required for real-world hybrid autonomy. This thesis adopts that elevation-driven CNN framework for M4 and uses its output as the primary terrain signal for both ground navigation and multimodal switching.

\section{LiDAR--Inertial Localization and Mapping}
\label{sec:lr-localization}

Terrain-aware autonomy depends on localization precision. If the robot's pose estimate drifts, elevation maps smear and traversability scores become unreliable. Field robots therefore require high-rate, drift-resilient localization that functions in GPS-denied environments.

LiDAR--inertial odometry (LIO) has emerged as a strong solution for outdoor robotics. FAST-LIO introduced a tightly coupled LiDAR--IMU fusion framework that achieves real-time (100+ Hz) state estimation with strong drift control even in high-dynamics motion~\cite{xu2021fastlio}. FAST-LIO2 further improved direct LiDAR-inertial odometry with enhanced efficiency~\cite{xu2022fastlio2}. LIO-SAM incorporated factor-graph optimization and loop closure to improve long-term consistency~\cite{shan2020liosam}. LOAM established early scan-matching pipelines for LiDAR odometry~\cite{zhang2014loam}, later extended with LeGO-LOAM for lightweight ground-optimized operation on variable terrain~\cite{shan2018lego}.

Compared to purely visual SLAM, LiDAR-based systems are less sensitive to illumination changes and can maintain mapping quality in dust, foliage, or low-texture environments. A comprehensive survey by Cadena et al.\ examined the past, present, and future of simultaneous localization and mapping toward the robust-perception age~\cite{cadena2016slam}. Visual--inertial systems such as ORB-SLAM3 remain valuable baselines for multi-sensor autonomy, offering loop closure and dense mapping in structured environments~\cite{campos2021orb}. RTAB-Map provides an open-source library for visual and LiDAR SLAM suitable for large-scale and long-term operation~\cite{labbe2019rtabmap}.

For M4, reliable LIO is essential because elevation mapping and traversability prediction depend on accurate point cloud deskewing and consistent frame alignment. The literature suggests that LiDAR--IMU fusion provides the most stable backbone for unstructured terrain autonomy, particularly for platforms that switch between motion regimes. This thesis therefore builds on FAST-LIO for state estimation, using its output to drive elevation map updates and terrain reasoning downstream.

\section{Energy-Aware and Multi-Objective Planning for Hybrid Platforms}
\label{sec:lr-planning}

Hybrid robots require planners that reason about more than geometric shortest paths. For multimodal systems, a ground path that is geometrically feasible might be energetically expensive due to rough terrain, while a short aerial bypass may be impossible due to limited battery or transition cost. Mode selection therefore requires multi-objective planning.

Classical shortest-path methods such as Dijkstra~\cite{shu2012dijkstra} and A*~\cite{hart1968astar} underpin most modern field planners. The dynamic window approach provides local collision avoidance for reactive navigation~\cite{fox1997dwa}. Probabilistic roadmap (PRM) methods discretize configuration spaces into graphs, enabling efficient path search in high-dimensional spaces~\cite{kavraki1998prm}. For multimodal systems, these methods must be extended to incorporate mode-specific costs and transition penalties.

Energy cost modeling itself is an active research area. Self-supervised cost of transport estimation enables accurate mode-dependent energy prediction from experience, improving switching accuracy over hand-crafted power approximations~\cite{gherold2025self}. PRM-based multimodal planning with reference governors has demonstrated energy-aware mode selection in simulation~\cite{sihite2022efficient}.

Multi-objective planners often combine distance, traversability, energy, and risk. Quadrupedal locomotion control on inclined surfaces using collocation methods demonstrates terrain-aware planning for legged systems~\cite{salagame2024quadrupedal}. Contact-implicit motion planning enables morpho-functional loco-manipulation with non-impulsive transitions~\cite{salagame2024non, salagame2024loco}. Optimal trajectory planning using contact-implicit optimization has been applied to snake robot locomotion~\cite{salagame2025optimal}, while reduced-order model-based gait generation using NMPC enables real-time planning~\cite{salagame2025reduced}.

For fault-tolerant operation, NMPC-based unified posture manipulation and thrust vectoring enables fault recovery in multimodal systems~\cite{salagame2025nmpc}. Such planners align closely with M4's goal: ground travel is preferred when terrain cost remains low, while aerial bypass is triggered when traversability-based ground cost exceeds the energetic penalty of flight plus transition overhead.

The literature indicates that multimodal systems benefit from explicit cost comparison between candidate paths across different locomotion regimes, rather than reactive obstacle-triggered switching. This thesis adopts that principle by embedding traversability-derived costs into the ground planner and jointly evaluating aerial bypass paths within a unified energy objective.

\section{Integration of Perception and Navigation Frameworks}
\label{sec:lr-integration}

Even with strong perception and planning components, robust autonomy requires careful integration into a navigation framework. Modern robotics stacks increasingly emphasize modular architectures, enabling perception-driven costmaps, replanning, and safety recovery.

The DARPA Subterranean Challenge demonstrated the importance of integrated autonomy stacks for complex environments. Team CERBERUS won the challenge using sophisticated perception and navigation integration~\cite{tranzatto2022cerberus}. For terrain-aware autonomy, the key integration challenge is creating costmaps that encode not only obstacles but also continuous terrain difficulty. Several systems have incorporated elevation-derived costs into layered costmaps, enabling path planners to route through safer terrain rather than merely avoiding collisions~\cite{fankhauser2016grid, miki2022elevation}.

Hybrid robots further require transition monitoring and failure reasoning. Vision-guided loco-manipulation with snake robots demonstrates integrated perception and planning for complex manipulation tasks~\cite{salagame2025vision}. Legged and hybrid locomotion controllers often integrate stability observers or momentum-based safety monitors to prevent unsafe transitions on steep or irregular ground. Rough-terrain locomotion with unilateral contact force regulations enables robust operation on challenging substrates~\cite{liang2021rough}. Thruster-assisted locomotion using decoupled model predictive control with learned contact residuals provides robust control during dynamic maneuvers~\cite{wang2025thruster}.

Bio-inspired approaches offer additional insights for integration. Work on dynamic morphing wing flight, including actuation and flight control by shifting structure response~\cite{sihite2023actuation}, bounding flight control~\cite{gupta2024bounding}, and banking turns using optimization~\cite{gupta2024banking}, demonstrates sophisticated integration of sensing, planning, and control. Morphology-centered views of dynamically versatile wing conformations provide design principles for adaptable systems~\cite{sihite2025morphology}. Efficient modeling using neural networks and cubature rules enables real-time computation for complex dynamics~\cite{ghanem2021efficient}.

The generative design of morpho-functional legged robots such as Husky Carbon demonstrates integrated design and control approaches~\cite{ramezani2021generative}. Hovering control of flapping wings in tandem with multi-rotors shows hybrid aerial system integration~\cite{dhole2023hovering}. Trajectory planning for bat-like flapping wing robots~\cite{hoff2019trajectory} and optimization of robotic bat structure and movement with biological kinematic synergies~\cite{hoff2018optimizing} provide bio-inspired foundations for multimodal system design.

In M4's context, integration means translating CNN traversability predictions into costs compatible with Nav2's global and local planning threads, while continuously evaluating whether a ground path remains feasible relative to flight energy. The COBRA platform, developed for crater observation, demonstrates similar integration principles for challenging terrain navigation~\cite{salagame2025crater}. The literature supports this architecture: traversability-driven costmaps provide richer planning signals, and recovery monitors enable robust switching when ground navigation degrades. This thesis implements those principles within a ROS2-based autonomy stack for physical M4 deployment.

% Methodology

% 
% CHAPTER 3: PERCEPTION SYSTEM VALIDATION
% 
\chapter{Perception System Validation}
\label{chap:PerceptionValidation}

Accurate terrain perception forms the foundation of traversability-based navigation for multimodal robots. Before implementing the complete autonomy pipeline on the M4 platform, a critical design decision must be addressed: which sensing modality, camera-based depth reconstruction or LiDAR, provides sufficient geometric fidelity for elevation mapping and subsequent terrain reasoning? This chapter presents a systematic, platform-agnostic study comparing LiDAR-based and camera-based SLAM under controlled conditions to establish an evidence-based sensor selection for the M4 navigation system described in Chapter~\ref{chap:SystemArchitecture}.

\section{Experimental Motivation and Objectives}
\label{sec:experimental-motivation}

Elevation mapping serves as the foundation for traversability-based mode selection for the M4 robot. The quality of these height maps directly influences the reliability of terrain difficulty estimation, which ultimately determines whether locomotion mode transitions are justified or potentially unsafe. Before developing the M4 autonomy pipeline, we conducted an independent, platform-agnostic perception study to answer a fundamental question: "Is camera-based depth reconstruction sufficiently accurate for elevation mapping, or is LiDAR required to obtain the geometric fidelity needed for stable terrain reasoning?" The main reason for comparison is the tradeoff of using lidar, which consumes more power and adds to the extra weight which significantly impacting the flight time when M4 switches mode. And even when traversing as a wheeled robot

This study was originally performed as part of prior work on the Cobra snake-robot platform~\cite{salagame2025crater}, but the analysis itself intentionally avoided assumptions tied to any specific robot geometry or locomotion mode. Instead, the focus was on comparing sensing modalities, camera SLAM versus LiDAR SLAM, under identical conditions to understand their suitability for generating reliable 3D terrain structure. Because the M4 robot relies heavily on elevation maps for traversability prediction and mode selection, the insights from this earlier study provide a direct, evidence-based justification for selecting LiDAR as the primary mapping modality for this thesis.

Camera-based systems offer well-known advantages: low cost, lightweight packaging, and the ability to provide visual texture useful for appearance-based terrain classification. Stereo and RGB-D sensors are widely used in indoor mobile robotics, and visual SLAM systems such as RTAB-Map can provide dense depth reconstructions alongside localization~\cite{labbe2019rtabmap}. However, monocular systems suffer from scale ambiguity~\cite{cadena2016slam}, stereo systems struggle on low-texture or repetitive surfaces, and depth noise increases rapidly with distance. These limitations can degrade the quality of elevation maps, particularly for outdoor or unstructured environments where texture can be sparse.

LiDAR, in contrast, provides direct time-of-flight measurements with centimeter-level accuracy, independent of lighting or visual texture. Although LiDAR sensors come with higher cost and weight, their accuracy and range characteristics make them strong candidates for geometric mapping, especially for robots like M4, that depend on precise height estimates to discriminate between navigable terrain and obstacles as small as 5–10 cm.

To evaluate these trade-offs systematically, we performed a comparison of LiDAR-based SLAM (LIO-SAM) and camera-based SLAM (RTAB-Map) using an OptiTrack motion capture system as ground truth~\cite{optitrack2023} for odometry and lidar as a reference point-cloud for as compared to the point-cloud obtained by VIO. The study involved two primary objectives:

\begin{enumerate}
    \item \textbf{Check point cloud quality.} Examine the 3D point clouds produced by both systems and measure how much they differ from a known reference surface. This helps us understand how clean, accurate, or noisy each reconstruction is.
    
    \item \textbf{Compare trajectory accuracy.} Evaluate how closely each SLAM pipeline tracks the robot's actual motion by comparing its pose estimates against OptiTrack ground truth. This shows how much drift or inconsistency each method accumulates during typical runs.
\end{enumerate}

The controlled indoor setting and the use of OptiTrack ground truth allowed us to isolate the compare the SLAM performance from different environmental setups. The results demonstrated that LiDAR-based SLAM consistently produced smoother, denser, and more accurate point clouds, with substantially lower drift than camera-based SLAM. These findings were expected, but to understand what extent the camera-based point cloud results differ, these comparisons were necessary. These findings directly helped the design of M4's autonomy stack, because traversability and mode switching rely on elevation maps that must reflect height differences on the order of only a few centimeters. LiDAR turned out to be unavoidable as the only sensing modality capable of meeting the accuracy requirements for safe operation.

\section{Test Rig Design and Fabrication}
\label{sec:test-rig-design}

To enable fair and controlled comparison between LiDAR-based and camera-based SLAM, a custom test rig was designed and fabricated with the following requirements: rigid mechanical coupling between sensors to ensure they traverse identical trajectories, precise and repeatable sensor mounting to enable accurate transformation calibration, compact and portable form factor for maneuverability in indoor test environments, and sufficient structural stability to prevent sensor vibration or relative motion during data collection.

\begin{figure}[htbp]
    \centering
    \begin{subfigure}[b]{0.38\textwidth}
        \centering
        \includegraphics[width=\textwidth]{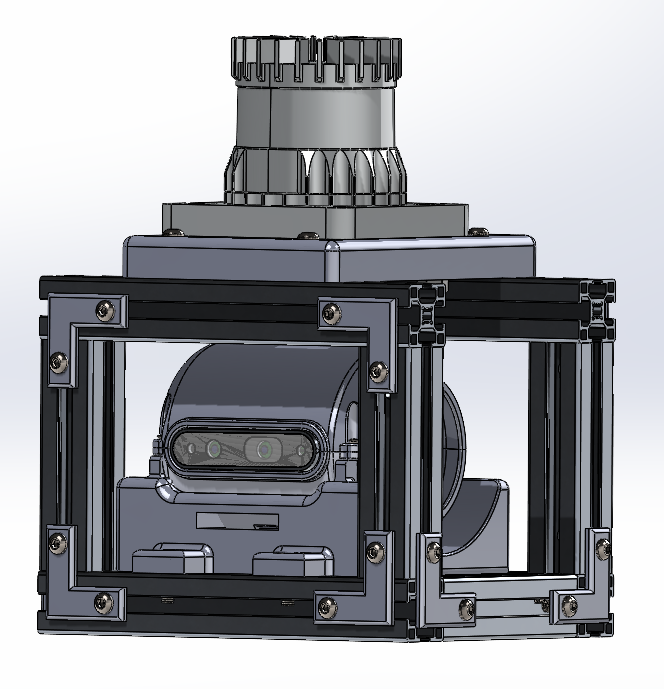}
        \caption{CAD model}
        \label{fig:rig-cad-iso}
    \end{subfigure}
    \hfill
    \begin{subfigure}[b]{0.38\textwidth}
        \centering
        \includegraphics[width=\textwidth]{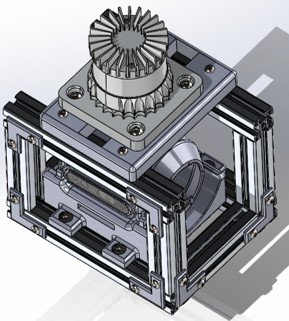}
        \caption{CAD model - Top view}
        \label{fig:rig-cad-side}
    \end{subfigure}
    
    \vspace{0.6cm}
    
    \begin{subfigure}[b]{0.38\textwidth}
        \centering
        \includegraphics[width=\textwidth]{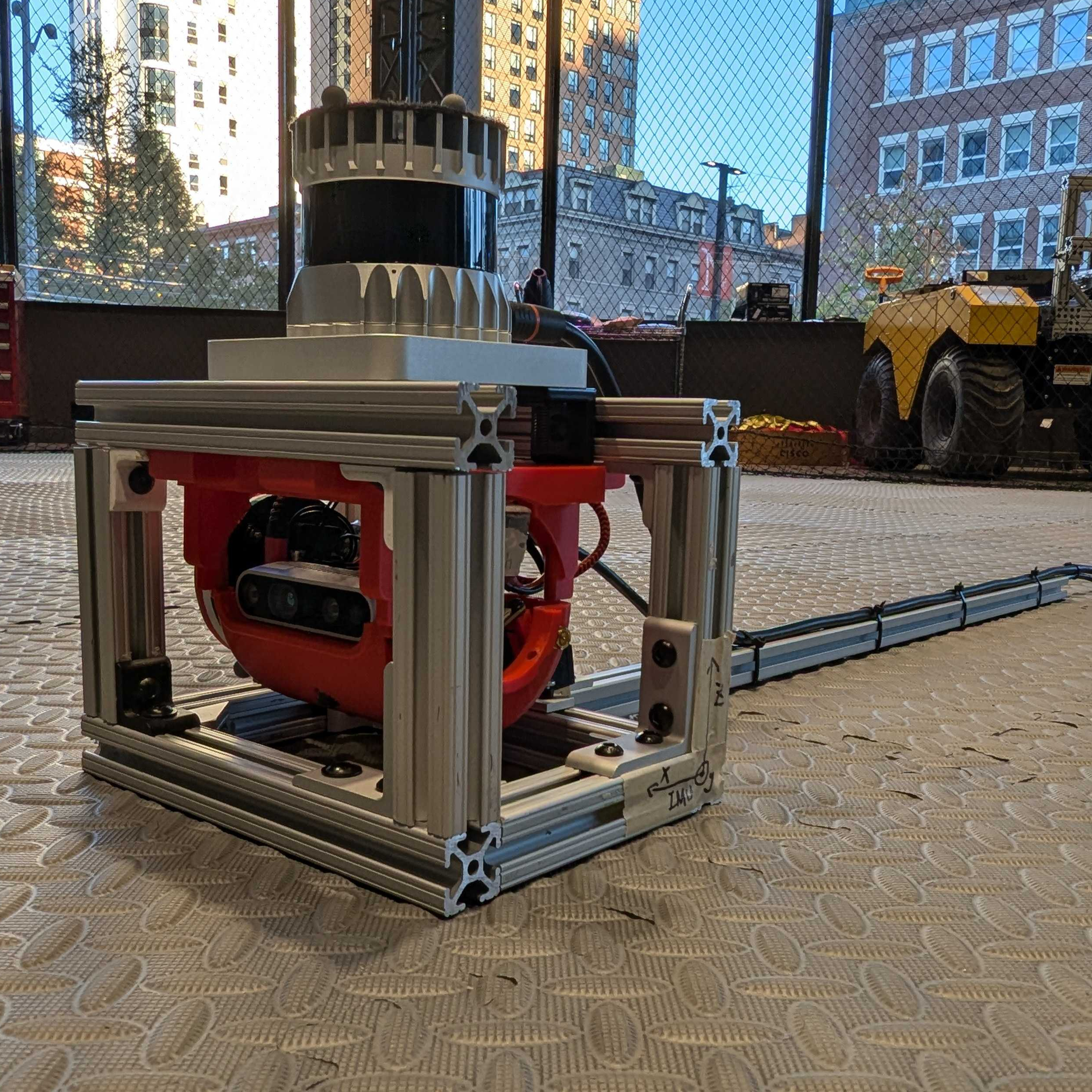}
        \caption{Physical assembly - view 1}
        \label{fig:rig-real-1}
    \end{subfigure}
    \hfill
    \begin{subfigure}[b]{0.38\textwidth}
        \centering
        \includegraphics[width=\textwidth]{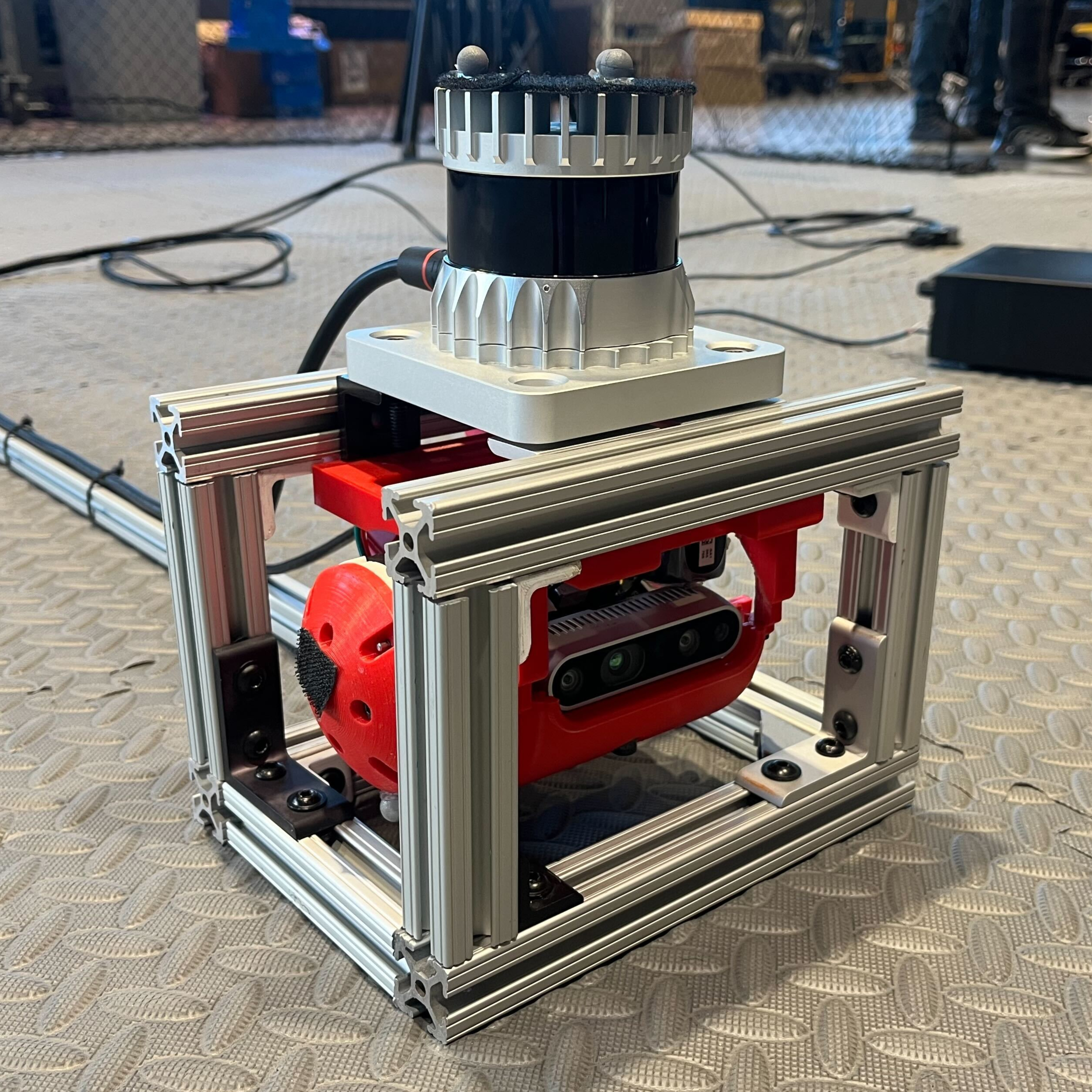}
        \caption{Physical assembly - view 2}
        \label{fig:rig-real-2}
    \end{subfigure}
    
    \caption{Illustrates the design of the custom test rig for SLAM benchmarking.}
    \label{fig:test-rig}
\end{figure}

The resulting test rig, shown in Fig.~\ref{fig:test-rig}, measures 9 inches in length, 6 inches in width, and 6 inches in height, constructed primarily from aluminum extrusions forming a rigid rectangular frame. Aluminum was selected for its favorable strength-to-weight ratio, ease of machining, and thermal stability, as the lidar heats up quite a bit after excessive use and in the absence of heat shrink, aluminum extrusions were crucial to dissipate heat. The extrusion-based design allowed rapid prototyping and simple modifications during development, with ease of assembling and dissembling, with T-slot and L-slot connections providing adjustable mounting positions for sensor placement optimization.

Sensor mounting was accomplished through custom 3D-printed brackets designed to interface between the sensors' bolt patterns and the aluminum extrusion T-slots. The Ouster REV3 LiDAR was positioned at the top of the rig to maximize its vertical field of view and minimize occlusions from the rig structure itself. The RealSense depth camera was mounted below the LiDAR in a vertically stacked configuration, with both sensors oriented to face forward along the rig's primary axis. This arrangement ensured that both sensors observed approximately the same scene volume during data collection, though with different perspectives due to the vertical offset and hardware limits.

The test rig was originally conceived to support mapping validation for the COBRA snake robot, a serpentine locomotion platform under development at the SiliconSynapse Lab. However, the SLAM benchmarking framework is platform-agnostic; the performance characteristics of mapping algorithms and the geometric accuracy of different sensors apply broadly to any mobile robot requiring terrain representation. The validation insights and sensor selection conclusions drawn from this work directly informed the M4 system design, particularly the decision to employ LiDAR as the primary sensor for elevation mapping despite the additional weight and power requirements compared to camera-only solutions.

The compact dimensions and handheld portability of the test rig helped the data collection across diverse indoor environments fairly easily while maintaining the consistent sensor configuration necessary for controlled comparison. OptiTrack reflective markers were fixated to the top of the lidar sensor to enable motion capture tracking during experiments, providing the ground truth pose reference required for trajectory accuracy validation described in Section~\ref{sec:methodology-setup}.

\section{Sensor Calibration and SLAM Setup}
\label{sec:methodology-sensing}

This section describes the setup, calibration procedures, and processing pipeline used throughout the benchmarking study. Although the study predates the current M4 implementation, the analysis directly shaped and affirmed the choice of sensor selection for elevation mapping.

The Ouster REV3 128-beam LiDAR was operated at 20\,Hz using the official Ouster ROS driver. For indoor experiments, the range was manually set at 10\,m to avoid boundary reflections and to reduce unnecessary load during registration and also because our test setup was also limited to 5m by 5m. Point cloud data were streamed on \texttt{/ouster/points} and IMU packets on \texttt{/ouster/imu}.

Accurate IMU noise characterization is essential for tightly-coupled LiDAR-inertial odometry systems. A four-hour static dataset was recorded from the LiDAR's integrated IMU. Allan variance analysis~\cite{allan1966statistics} was performed using the \texttt{imu\_allan\_ros} package, which computed accelerometer and gyroscope noise characteristics including angle random walk, velocity random walk, and bias instability. These values were later used to configure IMU pre-integration in LIO-SAM.

Only the procedure is described here; the extracted parameters and Allan deviation results are reported in Chapter 6.

A Madgwick filter was applied afterward to generate an orientation estimate required for FAST-LIO integration.

\section{SLAM Implementations and pointcloud pre-processing}
\label{subsec:slam-implementations}

\paragraph{LIO-SAM.}
LIO-SAM~\cite{shan2020liosam} was configured with the corrected IMU noise parameters. Indoor mapping required reducing voxel grid size (from the outdoor default 0.4–0.5\,m down to approximately 0.15\,m) to suppress NaN returns from near objects.

\paragraph{RTAB-Map.}
RTAB-Map~\cite{labbe2019rtabmap} ran in RGB-D mode using a RealSense camera. This appearance-based SLAM system combines visual odometry with bag-of-words loop closure detection. Loop closure was enabled with conservative thresholds. Feature-tracking degraded when the camera moved too rapidly, so the robot was manually guided at a steady, slow pace. RTAB-Map published its colored point clouds on \texttt{/rtabmap/cloud\_map}.

All geometric analysis was conducted in CloudCompare~\cite{cloudcompare}. Color from RTAB-Map clouds was removed and transformed into a mono coloured cloud manually in CloudCompare so that registration relied strictly on geometry.

\subsection{ICP Alignment}
\label{subsec:icp-alignment}

Iterative Closest Point (ICP)~\cite{besl1992icp} registration was used to align point clouds from the two SLAM systems. While a CAD-defined transform existed for sensor placement, we instead applied dataset-specific ICP refinement to accurately recover the true sensor alignment. This compensated for assembly tolerances and minor deviations from nominal mounting positions. ICP consistently converged with sub-25\,cm RMS error and estimated offsets close to the measured sensor placement (approximately --10.8\,cm vertical, +1.7\,cm forward).

\subsection{Segmentation and reasoning}
\label{subsec:segmentation}

Analyzing an entire environment as one cloud was misleading because:
\begin{itemize}
    \item distant walls naturally appear as large geometric deviations (despite being correct),
    \item occluded regions inflate global error scales,
    \item objects with very different geometric scale (flat walls vs small boxes) distort color-map interpretation.
\end{itemize}
Therefore, each scenario (planar surface, box, occluded planks, full cage) was isolated and evaluated independently.

\subsection{Triangulation and Cloud-to-Cloud Distance Computation}
\label{subsec:c2c-computation}

LIO-SAM clouds were triangulated using Delaunay meshing. For each RTAB-Map point $p_i$, the perpendicular distance to the neaREST triangle was computed:

\begin{equation}
d_i = \min_{\triangle \in M} \| p_i - \mathrm{proj}_{\triangle}(p_i) \|
\label{eq:c2c-distance}
\end{equation}

This allowed fine-grained, geometry-aware comparison against a dense LiDAR surface rather than point-to-point heuristics.

\begin{figure}[htbp]
    \centering
    \includegraphics[width=0.85\textwidth]{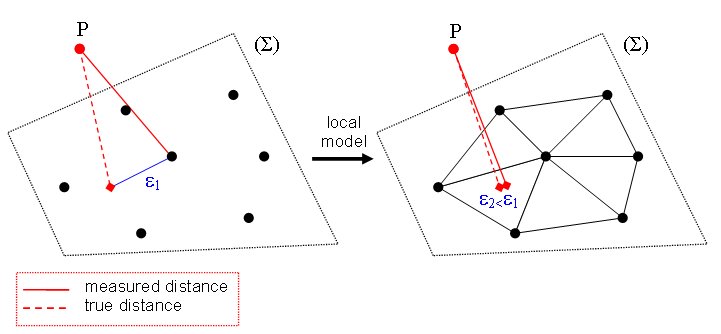}
    \caption{Illustrates the cloud-to-mesh triangular distance computation.}
    \label{fig:delaunay-mesh}
\end{figure}

\section{Experimental Setup and Test Scenarios}
\label{sec:methodology-setup}

All experiments were performed inside a motion-capture-equipped room with OptiTrack ground truth. The OptiTrack system provided sub-millimeter accuracy pose estimates at 120\,Hz, serving as the reference for trajectory evaluation~\cite{optitrack2023}.

\begin{figure}[htbp]
    \centering
    \includegraphics[width=0.85\textwidth]{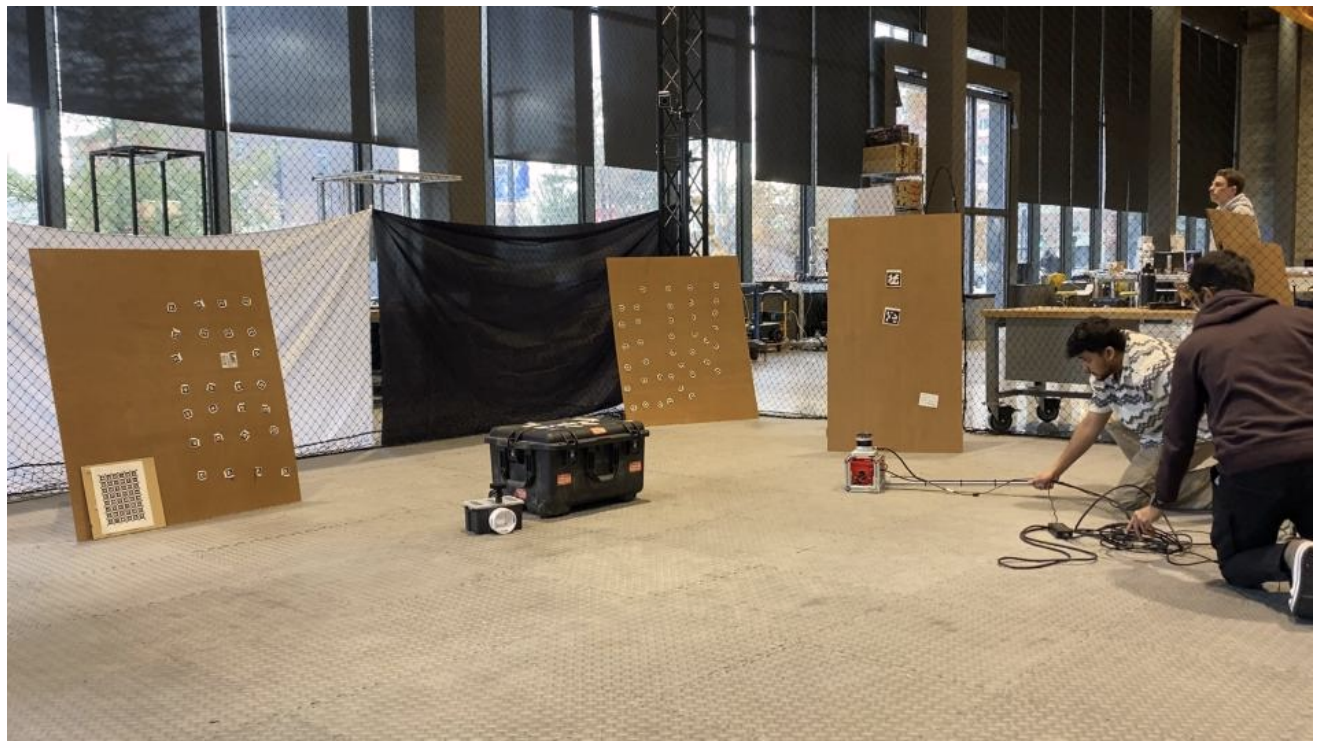}
    \caption{Setup for SLAM benchmarking.}
    \label{fig:experiment-setup}
\end{figure}

Four scenarios were collected and processed to evaluate different aspects of geometric reconstruction quality:

\begin{table}[htbp]
\centering
\caption{Test scenarios used for geometric quality evaluation.}
\label{tab:test-scenarios}
\begin{tabular}{|l|p{6cm}|p{5cm}|}
\hline
\textbf{Scenario} & \textbf{Description} & \textbf{Purpose} \\
\hline
Planar surface & Large flat cloth surface mounted vertically & Evaluate height consistency and planar reconstruction \\
\hline
Isolated box & Single cardboard box placed at center of cage & Test object geometry, edge reconstruction, and discontinuity handling \\
\hline
Occluded planks & Wooden planks arranged to partially block camera view & Assess robustness in cluttered or partially occluded scenes \\
\hline
Full cage environment & Entire indoor test cage with mesh walls and objects & Evaluate overall map structure and combined reconstruction behavior \\
\hline
\end{tabular}
\end{table}

Trajectory error metrics were computed using the \texttt{evo} evaluation tool~\cite{grupp2017evo}, which provides Absolute Trajectory Error (ATE) and Relative Pose Error (RPE) metrics following established SLAM benchmarking conventions. Geometric deviation metrics were obtained from CloudCompare's cloud-to-cloud (C2C) distance operator.

The combination of trajectory accuracy assessment (comparing SLAM odometry against OptiTrack ground truth) and point cloud quality evaluation (measuring geometric deviation between LiDAR and camera reconstructions) provides a comprehensive characterization of each sensing modality's suitability for elevation mapping applications. Results from these experiments are presented in Chapter~\ref{sec:results}.
% =============================================================
%          CHAPTER 4: SYSTEM ARCHITECTURE AND IMPLEMENTATION
% =============================================================
\chapter{System Architecture and Implementation}
\label{chap:SystemArchitecture}

Building upon the sensor selection validated in Chapter~\ref{chap:PerceptionValidation}, this chapter presents the complete navigation system architecture implemented on the M4 multimodal robot. The system integrates LiDAR-based localization, GPU-accelerated elevation mapping with learned traversability estimation, and energy-aware path planning within a distributed ROS2 framework. Each component is designed to support the ultimate goal of autonomous locomotion mode selection based on real-time terrain analysis.

\section{M4 Platform Overview and Hardware Integration}
\label{sec:m4-platform}

\subsection{Platform overview and LiDAR Installation}
\label{subsec:platform-architecture}

\begin{figure}[htbp]
    \centering
    \includegraphics[width=0.9\textwidth]{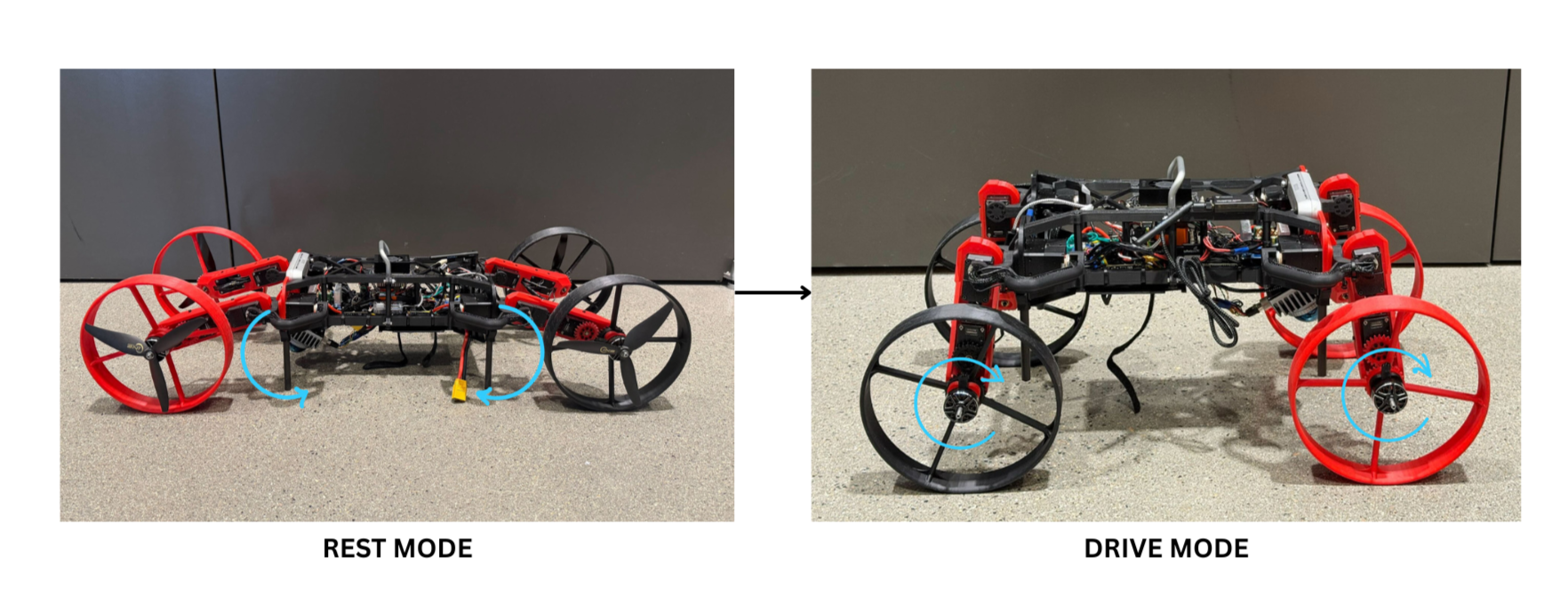}
    \caption{illustrates M4 platform in rest mode (left) and drive mode (right) configuration.}
    \label{fig:m4_modes}
\end{figure}

The ground locomotion system employs four Dynamixel XM540-W270-R servo motors configured in a mechanically coupled differential drive arrangement~\cite{dynamixel2023}. The front-left and back-left wheels operate synchronously, as do the front-right and back-right wheels, creating a two-sided differential drive system that provides four-wheel traction while maintaining kinematic simplicity. Each servo provides 270° rotation with 4096-step resolution and integrated velocity control, enabling precise odometry feedback for dead-reckoning estimation. Figure~\ref{fig:m4_modes} illustrates the platform in both rest and drive configurations.

\begin{figure}[htbp]
    \centering
    \begin{subfigure}[b]{0.48\textwidth}
        \centering
        \includegraphics[height=5cm]{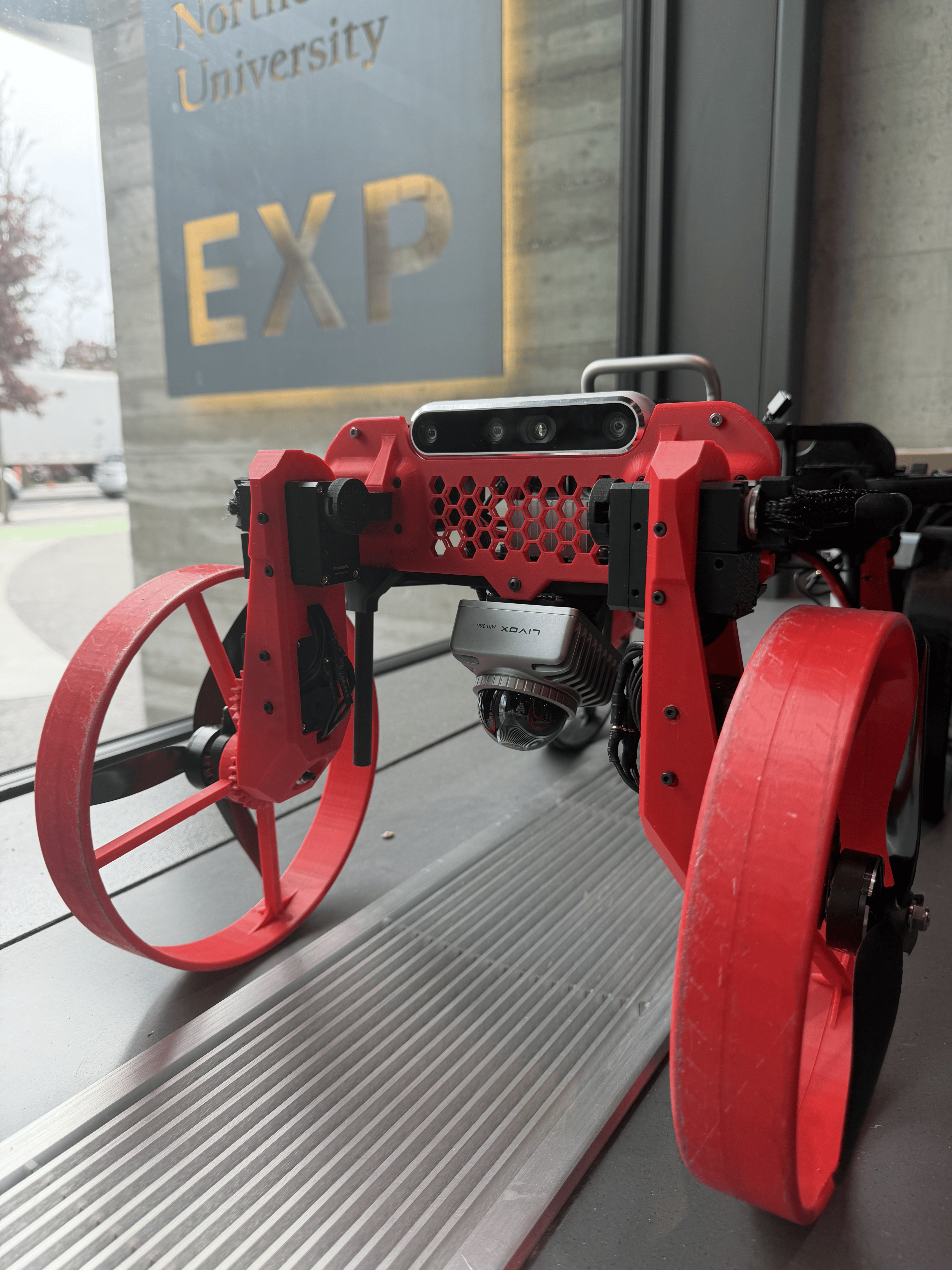}
        \caption{Actual sensor installation}
        \label{fig:lidar_physical}
    \end{subfigure}
    \hfill
    \begin{subfigure}[b]{0.48\textwidth}
        \centering
        \includegraphics[width=\textwidth]{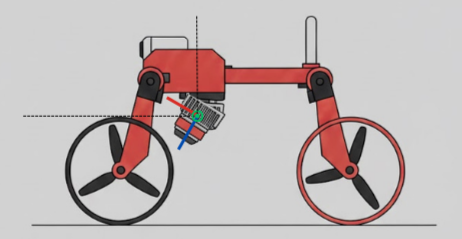}
        \caption{Coordinate frame representation}
        \label{fig:lidar_axis}
    \end{subfigure}
    \caption{Illustrates the mounting orientation of Lidar.}
    \label{fig:lidar_mount}
\end{figure}

Perception hardware centers on a Livox MID-360 solid-state LiDAR sensor selected for its combination of high point density (200,000 points/second), compact form factor, and non-repetitive scanning pattern that mitigates motion distortion during dynamic movement~\cite{livox2020mid360}. The sensor mounts at a 30° angle from the inverted vertical axis (equivalently, 60° from the horizontal plane) to maximize ground coverage while maintaining sufficient forward and lateral field-of-view for obstacle detection. This orientation balances the competing requirements of elevation mapping which demands dense ground point returns and obstacle avoidance, which requires forward-looking perception. Figure~\ref{fig:lidar_mount} depicts the LiDAR mounting geometry and coordinate frame orientation.

The onboard computing architecture distributes processing across an NVIDIA Jetson Orin (robot-mounted) and an external laptop workstation. The Jetson handles real-time navigation control and sensor data acquisition, while the laptop executes computationally intensive elevation mapping and CNN-based traversability estimation. This distributed approach enables the system to meet real-time constraints while leveraging machine learning methods that exceed embedded GPU capabilities.

\subsection{Software Integration and Architecture}
\label{subsec:software-architecture}

The navigation system implements a distributed ROS2 (Robot Operating System 2) architecture~\cite{macenski2022ros2} with components communicating via DDS multicast networking. This design decouples sensor processing, mapping, planning, and control into independent nodes that can execute on different hardware platforms while maintaining synchronized operation through a shared ROS2 network.

The software stack uses Docker containerization for the elevation mapping and traversability estimation pipeline, ensuring reproducible deployment and dependency isolation. The Docker container executes on the external laptop and publishes traversability costmaps and elevation data to the ROS2 network, where they are consumed by navigation nodes running natively on the Jetson platform. This containerized approach simplifies integration of Python-based machine learning frameworks with the C++-dominated ROS2 ecosystem.

Core system components include:
\begin{itemize}
    \item \textbf{FAST-LIO}: LiDAR-inertial odometry providing drift-free localization at 100+ Hz
    \item \textbf{Elevation Mapping}: Grid-based terrain height estimation with CNN traversability analysis
    \item \textbf{Nav2}: Navigation framework providing path planning and obstacle avoidance
    \item \textbf{Ground Control Node}: Custom differential drive controller interfacing with Dynamixel servos
\end{itemize}

\begin{figure}[htbp]
    \centering
    \includegraphics[width=0.8\textwidth]{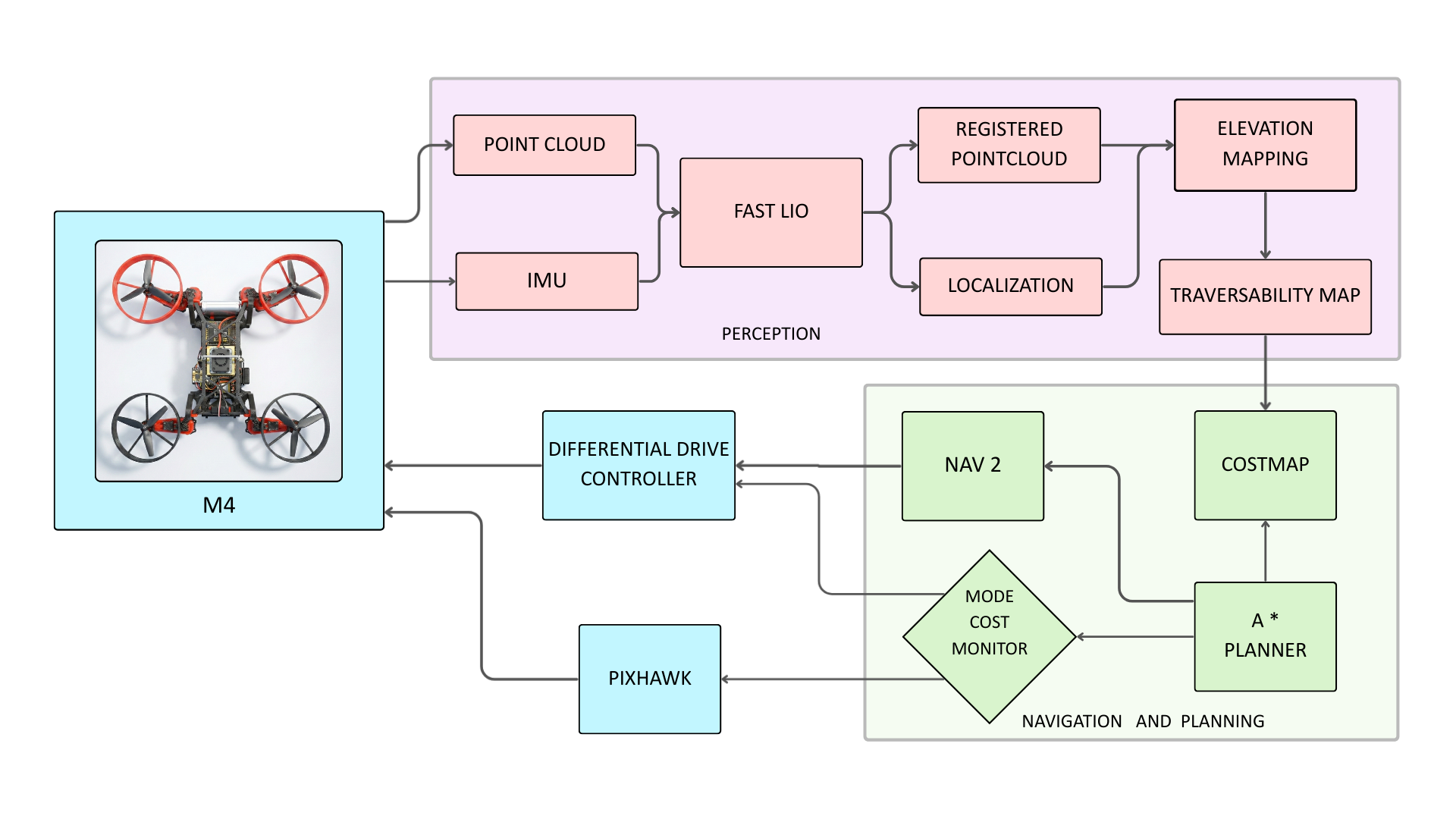}
    \caption{Illustrates the System architecture.}
    \label{fig:system_architecture}
\end{figure}

All nodes operate with standardized coordinate frames (\texttt{odom}, \texttt{base\_link}, along with the transformed \texttt{camera\_init\_corrected}) following REP-105 conventions. The system maintains two parallel localization estimates: FAST-LIO provides the primary \texttt{map} $\rightarrow$ \texttt{odom} transform for global accuracy, while wheel odometry from the ground control node publishes the \texttt{odom} $\rightarrow$ \texttt{base\_link} transform required by Nav2. FAST-LIO's frame is transformed using a static transform running simultaneously to overcome the lidar's mounting position and perceive the data in a corrected frame. Figure~\ref{fig:system_architecture} illustrates the complete system architecture showing node distribution, communication topology, and data flow between components.

\subsection{Robot Control}
\label{subsec:robot-control}

Ground mode operation utilizes a custom ROS2 node (\texttt{dyn4wd\_gs}) that implements the differential drive kinematic model and communicates with the Dynamixel servo motors via the Dynamixel SDK. The node subscribes to velocity commands from the navigation stack (\texttt{/cmd\_vel}, \texttt{geometry\_msgs/Twist} message type) and translates these into synchronized motor velocity setpoints for all four wheels.

\subsubsection{Inverse Kinematics}
\label{subsubsec:inverse-kinematics}

\begin{figure}[htbp]
    \centering
    \includegraphics[width=0.7\textwidth]{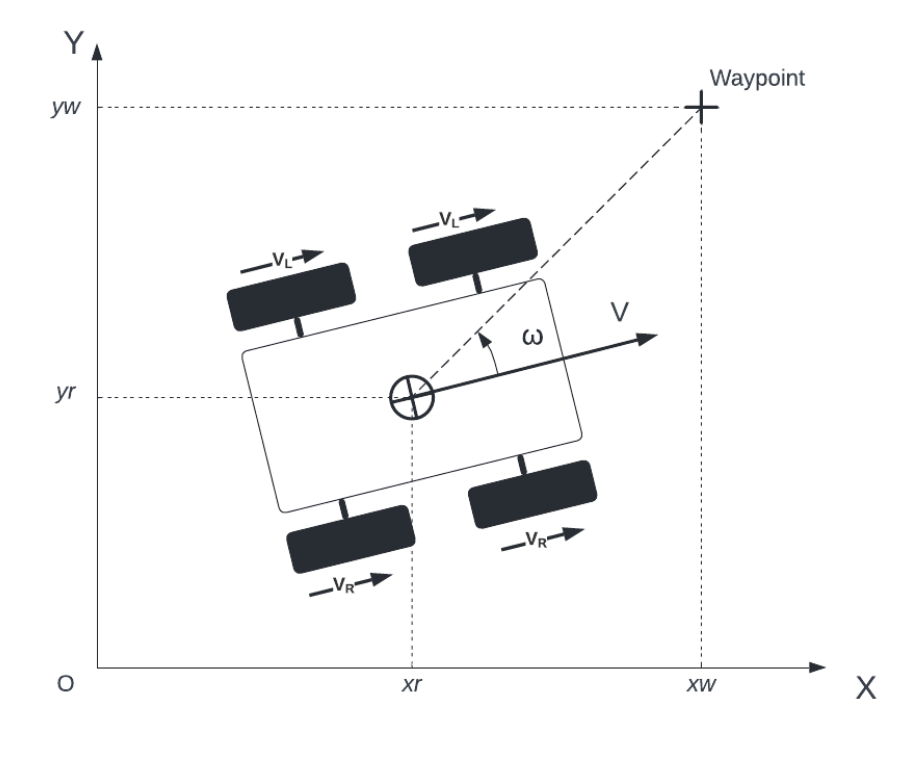}
    \caption{Differential drive kinematic model representation.}
    \label{fig:diff_drive_kinematics}
\end{figure}

The kinematic model follows standard differential drive formulation~\cite{siegwart2011robotics} as illustrated in Figure~\ref{fig:diff_drive_kinematics}. Given commanded linear velocity $v$ (m/s) and angular velocity $\omega$ (rad/s), left and right wheel velocities are computed as:

\begin{equation}
    v_L = v - \frac{\omega L}{2}, \quad v_R = v + \frac{\omega L}{2}
    \label{eq:diff-drive-velocities}
\end{equation}

\noindent where $L = 0.47$ m represents the track width between wheel centers. These velocities convert to motor RPM accounting for wheel radius $r = 0.076$ m and gear ratio $G = 1.0$:

\begin{equation}
    \text{RPM}_{\text{motor}} = \frac{60 \cdot v_{\text{wheel}}}{2\pi r G}
    \label{eq:velocity-to-rpm}
\end{equation}

Motor communication employs GroupSyncWrite for simultaneous velocity commands to all four servos (IDs: FL=9, FR=10, BL=11, BR=12) in a single bus transaction, minimizing communication latency. The control loop operates at 50 Hz with safety features including velocity limit enforcement ($\pm$1.0 m/s linear, $\pm$1.0 rad/s angular), command timeout monitoring (300 ms), and RPM deadband filtering (1.0 RPM threshold) to eliminate zero-velocity drift.

\subsubsection{Forward Kinematics and Wheel Odometry}
\label{subsubsec:forward-kinematics}

The node implements forward kinematics for wheel-based odometry estimation through dead-reckoning integration of encoder feedback~\cite{borenstein1996odometry}. At each timestep, the arc lengths traveled by left and right wheel sides are averaged to compute linear displacement, while their difference yields angular displacement:

\begin{equation}
    \Delta s = \frac{s_L + s_R}{2}, \quad \Delta \theta = \frac{s_R - s_L}{L}
    \label{eq:displacement}
\end{equation}

The robot pose $(x, y, \theta)$ is then updated using:

\begin{align}
    x_{k+1} &= x_k + \Delta s \cos\left(\theta_k + \frac{\Delta\theta}{2}\right) \label{eq:odom-x} \\
    y_{k+1} &= y_k + \Delta s \sin\left(\theta_k + \frac{\Delta\theta}{2}\right) \label{eq:odom-y} \\
    \theta_{k+1} &= \theta_k + \Delta \theta \label{eq:odom-theta}
\end{align}

This odometry estimate is published on the \texttt{/odom} topic at 50 Hz. Due to wheel slip, unmodeled dynamics, and encoder quantization, wheel odometry accumulates drift over time~\cite{borenstein1996odometry}. For this reason, the system relies on FAST-LIO's drift-free LiDAR-inertial odometry for global localization, while wheel odometry provides short-term velocity estimates and satisfies Nav2's transform requirements.

\section{FAST-LIO Localization System}
\label{sec:fastlio-localization}

The system employs FAST-LIO (Fast LiDAR-Inertial Odometry)~\cite{xu2021fastlio, xu2022fastlio2} for primary robot localization, replacing the previously considered LIO-SAM framework. This transition was motivated by two key factors: FAST-LIO's lightweight computational footprint better suited the embedded Jetson platform, and its tightly-coupled iterated extended Kalman filter (iEKF) approach proved more robust to the non-repetitive scan patterns characteristic of the Livox MID-360 solid-state LiDAR. Unlike spinning mechanical LiDARs that LIO-SAM was originally designed for, the MID-360's irregular scanning pattern benefits from FAST-LIO's point-wise correspondence method rather than feature extraction approaches.

\begin{figure}[htbp]
    \centering
    \includegraphics[width=0.65\textwidth]{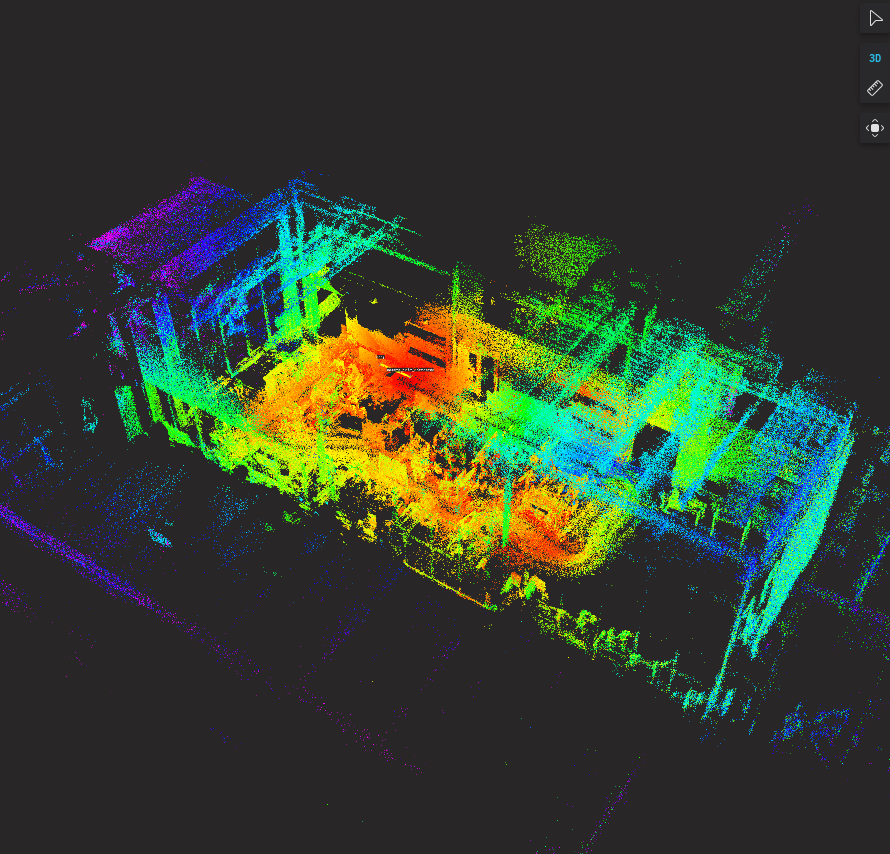}
    \caption{Cloud map generated by FAST-LIO.}
    \label{fig:fastlio_map_sample}
\end{figure}

System configuration prioritizes real-time performance while maintaining localization accuracy. Point cloud downsampling uses a 0.5 m voxel filter to reduce computational load without sacrificing terrain detail necessary for elevation mapping. The ikd-tree (incremental k-d tree) map representation~\cite{cai2021ikdtree} maintains a local map of 500 m radius, providing sufficient context for loop closure detection while preventing unbounded memory growth during extended operation. FAST-LIO publishes the \texttt{map} $\rightarrow$ \texttt{odom} coordinate transformation at 100+ Hz, providing drift-free global localization for the navigation stack.

For integration with downstream mapping and planning modules, the system utilizes two primary FAST-LIO outputs: the \texttt{/cloud\_registered\_body} topic provides motion-compensated point clouds in the robot body frame for elevation map construction, while the odometry output on \texttt{/Odometry} supplies high-rate robot pose estimates. These inputs are required by the elevation package which enables the elevation mapping node to construct accurate terrain representations aligned with the robot's coordinate frame, while the navigation stack receives continuous pose updates for trajectory tracking. Validation of FAST-LIO's localization performance through loop closure analysis and comparison with wheel odometry drift is presented in Chapter~\ref{chap:results}.

\section{Elevation Mapping and Traversability Estimation}
\label{sec:elevation-mapping}

A GPU-accelerated elevation mapping method is implemented to construct a 2.5D representation of the terrain surrounding the robot. This framework, based on the work of Miki et al.~\cite{miki2022elevation}, processes point clouds from FAST-LIO's \texttt{/cloud\_registered\_body} topic to generate a multi-layer grid map containing height estimates, uncertainty measures, and learned traversability predictions.

\subsection{Elevation Map Generation}
\label{subsec:elevation-generation}

The elevation map maintains a robot-centric 2D grid where each cell $(i,j)$ stores a height estimate $\hat{h}_{ij}$ and its associated variance $\sigma^2_{h,ij}$. The map frame is defined such that its z-axis remains aligned with the inertial vertical, while the x-y plane follows the robot's horizontal orientation. Map parameters include a 5.0 m $\times$ 5.0 m coverage area with 0.04 m resolution (125$\times$125 cells), providing sufficient local context for navigation planning while maintaining real-time update rates.

\subsubsection{Height Cell Update}
\label{subsubsec:height-update}

Each incoming point cloud from FAST-LIO undergoes transformation to the map frame, followed by height fusion using a Kalman filter formulation~\cite{fankhauser2014elevation}. For a point measurement $p_z$ falling into cell $(i,j)$, the height estimate is updated as:

\begin{equation}
    \hat{h}_{ij}^+ = \frac{\sigma_p^2 \hat{h}_{ij}^- + \sigma_{h,ij}^{2-} p_z}{\sigma_p^2 + \sigma_{h,ij}^{2-}}
    \label{eq:kalman-height}
\end{equation}

where the superscripts $-$ and $+$ denote estimates before and after the update, respectively. The measurement variance $\sigma_p^2$ accounts for sensor noise characteristics, modeled as $\sigma_p^2 = \alpha_d d^2$, where $d$ represents the distance from the sensor to the point and $\alpha_d = 0.002$ is an empirically determined noise coefficient for the MID-360 LiDAR.

The cell variance is updated through:

\begin{equation}
    \sigma_{h,ij}^{2+} = \frac{\sigma_{h,ij}^{2-} \sigma_p^2}{\sigma_{h,ij}^{2-} + \sigma_p^2}
    \label{eq:kalman-variance}
\end{equation}

To handle unobserved cells and account for potential drift in robot localization, a time-based variance inflation is applied at 0.1 Hz:

\begin{equation}
    \sigma_{h,ij}^{2} \leftarrow \sigma_{h,ij}^{2} + \sigma_t^2 \Delta t
    \label{eq:variance-inflation}
\end{equation}

where $\sigma_t^2 = 0.01$ m$^2$/s represents the variance growth rate and $\Delta t$ is the time elapsed since the last update.

\subsubsection{GPU Processing Pipeline}
\label{subsubsec:gpu-pipeline}

The elevation mapping implementation leverages CUDA kernels for parallel processing, enabling real-time operation despite the high point cloud density from the MID-360. The processing pipeline executes entirely on GPU memory, iterating through point clouds to perform coordinate transformation, outlier rejection based on Mahalanobis distance (threshold of 2.5), and simultaneous height updates across all affected cells. Point cloud downsampling uses a 0.5 m voxel filter to balance computational load against terrain detail preservation.

\begin{figure}[htbp]
    \centering
    \includegraphics[width=0.7\textwidth]{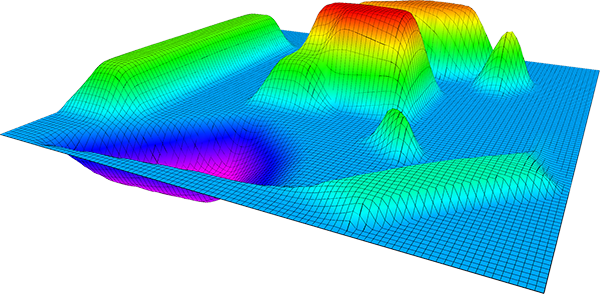}
    \caption{ Illustrates a sample elevation map.}
    \label{fig:elevation_sample}
\end{figure}

\subsection{Traversability Estimation}
\label{subsec:traversability-estimation}

Rather than relying solely on geometric features such as slope or roughness~\cite{papadakis2013traversability}, the system employs a learned traversability estimation approach using a convolutional neural network. This model, trained on expert demonstrations of successful navigation, captures complex terrain characteristics that may not be adequately represented solely by geometric methods.

\subsubsection{CNN Architecture}
\label{subsubsec:cnn-architecture}

The traversability estimation network consists of a lightweight CNN that processes local elevation map patches to produce traversability scores. The architecture comprises:

\begin{itemize}
    \item Input layer: 48$\times$48 pixel elevation patch centered on each cell
    \item Three convolutional layers (32, 64, 64 filters) with 3$\times$3 kernels and ReLU activations
    \item Two fully connected layers (128, 64 neurons) with dropout (p=0.5)
    \item Output layer: Single neuron with sigmoid activation producing traversability $\in [0,1]$
\end{itemize}

The network operates directly on GPU-resident elevation data using PyTorch, eliminating data transfer overhead between processing and inference. Inference executes at the full map update rate (10-20 Hz depending on point cloud density), providing real-time traversability estimates synchronized with terrain perception.

\begin{figure}[htbp]
    \centering
    \includegraphics[width=0.8\textwidth]{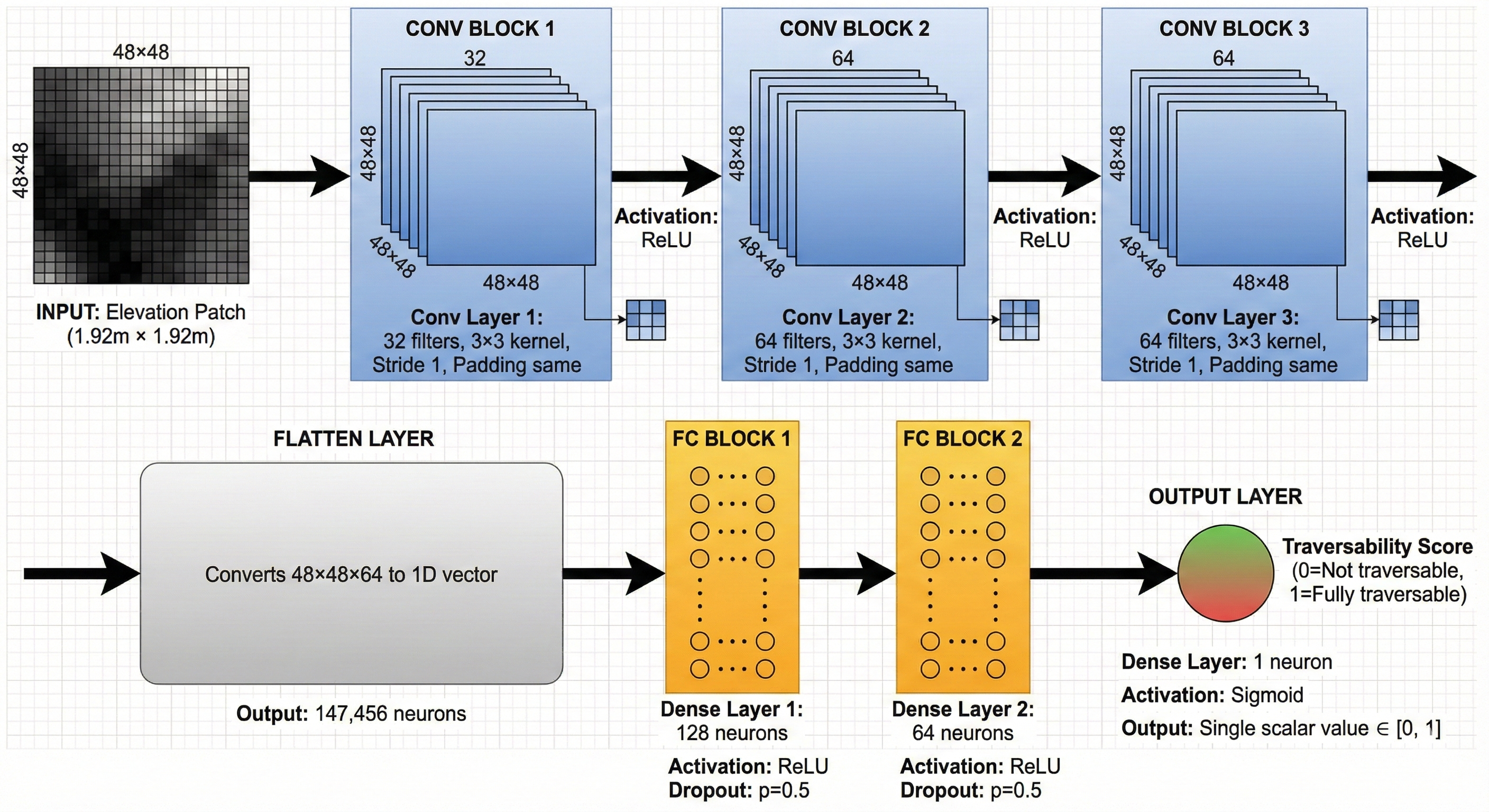}
    \caption{Illustrates the CNN architecture}
    \label{fig:cnn_arch}
\end{figure}

\subsubsection{Traversability Calculation}
\label{subsubsec:traversability-calc}

For navigation planning, traversability values are computed as:

\begin{equation}
    T_{ij} = \text{CNN}(\hat{h}_{(i-24):(i+24), (j-24):(j+24)})
    \label{eq:traversability}
\end{equation}

where $T_{ij} \in [0, 1]$ represents the predicted traversability for cell $(i,j)$ based on its local 1.92 m $\times$ 1.92 m neighborhood. Higher values indicate terrain suitable for ground-based navigation, while lower values suggest obstacles, steep slopes, or uncertain regions requiring aerial mode or path replanning.

The CNN model was trained on trajectory data collected during manual teleoperation over varied terrain, with traversable regions (successfully navigated paths) labeled as positive examples and obstacle-adjacent or failed approach regions as negative examples. This supervised learning approach captures implicit expert knowledge about terrain negotiability specific to the M4 platform's ground clearance (0.15 m) and kinematic constraints.

\subsection{Multi-Layer Map Structure}
\label{subsec:multilayer-map}

The complete elevation map comprises multiple synchronized layers stored in GPU memory:

\begin{itemize}
    \item \textbf{Elevation layer} ($\hat{h}_{ij}$): Mean height estimate per cell
    \item \textbf{Variance layer} ($\sigma_{h,ij}^2$): Height uncertainty for each cell
    \item \textbf{Traversability layer} ($T_{ij}$): CNN-predicted navigability scores
    \item \textbf{Time layer}: Timestamp of last measurement update per cell
\end{itemize}

This multi-layer structure enables the navigation system to query different terrain properties as needed: the planner uses traversability for mode selection decisions, variance information for path confidence estimation, and raw elevation data for geometric constraint estimation. The entire map updates at 15-20 Hz on the external laptop (NVIDIA RTX 3070), providing the navigation stack with current terrain awareness for real-time planning. Figure~\ref{fig:elevation_sample} illustrates a representative elevation map generated during operation.

The CNN architecture for traversability prediction is depicted in Figure~\ref{fig:cnn_arch}, showing the network structure from input elevation patches to output traversability scores.

\subsection{Implementation and Validation Workflow}
\label{subsec:implementation-workflow}

\begin{figure}[htbp]
    \centering
    \includegraphics[width=0.7\textwidth]{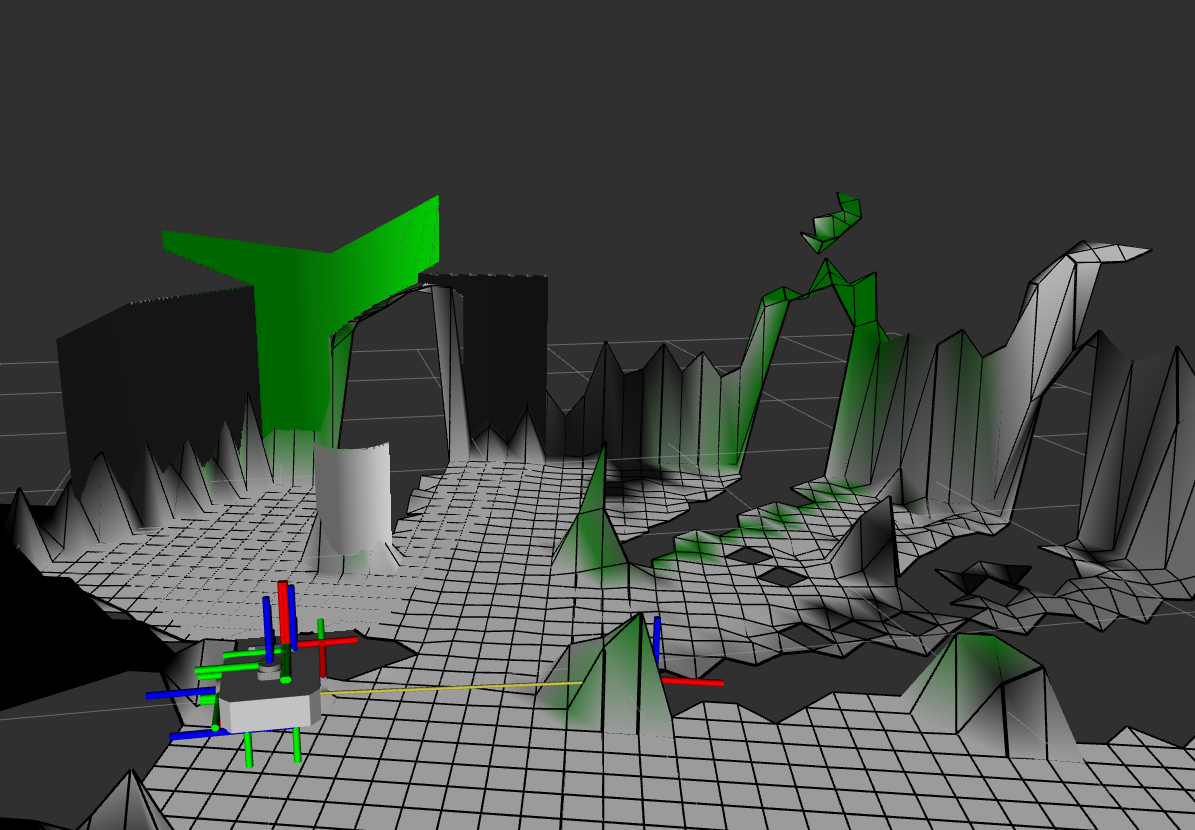}
    \caption{Elevation map generation during Gazebo simulation testing.}
    \label{fig:elevation_sim}
\end{figure}

Prior to deployment on the physical M4 platform, the elevation mapping framework underwent validation in a Gazebo simulation environment~\cite{koenig2004gazebo}. The simulation approach before actual hardware deployment helped the verification of ROS2 compatibility, parameter tuning, and integration testing with the navigation stack without risk to hardware. The simulated M4 model, equipped with a virtual LiDAR sensor plugin, traversed test scenarios including flat ground, obstacles of varying heights to validate map generation accuracy and real-time performance. Figure~\ref{fig:elevation_sim} demonstrates the elevation mapping output during simulation testing, validating framework functionality before hardware deployment.

Following successful simulation validation, the framework was deployed on the physical M4 robot with identical ROS2 topic interfaces and parameter configurations. The Docker-based deployment approach enables a smooth transition from simulation to hardware, as the containerized elevation mapping node required no modifications between platforms.

\subsection{Costmap Generation for Navigation}
\label{subsec:costmap-generation}

The traversability obtained from the elevation map must be converted into a format compatible with the Nav2 navigation stack~\cite{macenski2020nav2}, which expects costmaps represented in an acceptable format as \texttt{nav\_msgs/OccupancyGrid} messages. A dedicated costmap converter node bridges this gap, transforming the multi-layer GridMap structure into a persistent occupancy grid suitable for path planning.

Unlike traditional robot-centric costmaps that move with the robot, the system employs a large persistent costmap with fixed world coordinates. The map spans 100 m $\times$ 100 m at 0.1 m resolution (1000$\times$1000 cells), with origin at (-50, -50) m to center the robot's initial position. This persistent representation accumulates terrain knowledge over time, enabling the planner to reason about previously explored regions even when outside the current sensor field of view.

The map initializes optimistically with all cells set to cost = 50 (free/semi-traversable space), under the assumption that unmapped terrain is traversable until proven otherwise. As the elevation mapping system processes new sensor data, the costmap converter selectively updates cells within the high-confidence region of the elevation map.

To mitigate boundary artifacts and ensure costmap quality, the converter implements a selective update policy that only incorporates data from the central 6 m $\times$ 6 m region of the elevation map, the area with highest point cloud density and lowest uncertainty. An additional 1 m buffer is maintained from the crop boundary to exclude edge cells susceptible to interpolation artifacts. The first 20 messages from the elevation mapping node are discarded as a warmup period, allowing FAST-LIO localization to stabilize before map accumulation begins.

\subsubsection{Traversability-to-Cost Mapping}
\label{subsubsec:trav-to-cost}

The conversion from continuous traversability values $T \in [0,1]$ to discrete occupancy costs $C \in [0,100]$ employs a piecewise function designed to provide clear delineation between safe, marginal, and hazardous terrain:

\begin{equation}
C(T) = \begin{cases}
0 & T \geq T_{\text{high}} \\
\left\lfloor 20 \cdot \frac{T_{\text{high}} - T}{T_{\text{high}} - T_{\text{crit}}} \right\rfloor & T_{\text{crit}} \leq T < T_{\text{high}} \\
\left\lfloor 100 - 80 \cdot \frac{T}{T_{\text{crit}}} \right\rfloor & T < T_{\text{crit}}
\end{cases}
\label{eq:trav-to-cost}
\end{equation}

where $T_{\text{high}} = 0.85$ represents the threshold for fully traversable terrain, and $T_{\text{crit}} = 0.6$ delineates the boundary below which terrain is considered hazardous. This mapping ensures that terrain with CNN-predicted traversability above 0.85 receives zero cost (freely navigable), while terrain below 0.6 incurs costs ranging from 20-100, with lethal obstacles approaching cost = 100.

It is important to note that this costmap contains only geometric and terrain-based traversability information. Higher-level costs related to locomotion mode switching, energy consumption, and path optimality are incorporated subsequently by the global planner (Section~\ref{sec:navigation-planning}).

\begin{figure}[htbp]
    \centering
    \includegraphics[width=0.7\textwidth]{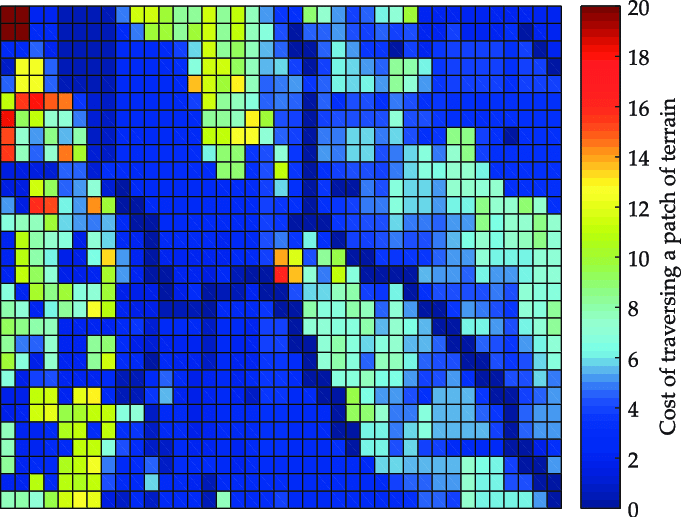}
    \caption{Sample navigation costmap}
    \label{fig:costmap_sample}
\end{figure}

Algorithm~\ref{alg:costmap_conversion} presents the costmap conversion procedure. For each incoming elevation map message, the algorithm extracts the traversability layer, applies the center crop with edge buffering, transforms valid traversability values to costs using the piecewise mapping function, and updates the corresponding cells in the persistent world-frame costmap. The resulting \texttt{OccupancyGrid} is published at 2 Hz for consumption by the navigation stack. Figure~\ref{fig:costmap_sample} illustrates a sample costmap generated during operation, showing the cost distribution corresponding to terrain traversability.

\begin{algorithm}[htbp]
\caption{Costmap Conversion from Traversability Layer}
\label{alg:costmap_conversion}
\begin{algorithmic}[1]
\State \textbf{Input:} Traversability map $T$, thresholds $T_{\text{high}} = 0.85$, $T_{\text{crit}} = 0.6$
\State \textbf{Output:} Costmap $\mathcal{C}$ with costs $\in [0, 100]$
\State
\For{each traversability score $t$ in $T$}
    \If{$t$ is NaN or $t < 0$}
        \State $\text{cost} \leftarrow -1$ \Comment{Unknown}
    \ElsIf{$t \geq T_{\text{high}}$}
        \State $\text{cost} \leftarrow 0$ \Comment{Fully traversable}
    \ElsIf{$t \geq T_{\text{crit}}$}
        \State $\alpha \leftarrow (t - T_{\text{crit}}) / (T_{\text{high}} - T_{\text{crit}})$
        \State $\text{cost} \leftarrow \lfloor 20 \cdot (1 - \alpha) \rfloor$ \Comment{Transition: 0-20}
    \Else
        \State $\alpha \leftarrow t / T_{\text{crit}}$
        \State $\text{cost} \leftarrow \lfloor 100 - 80 \cdot \alpha \rfloor$ \Comment{Less Traversible: 20-100}
    \EndIf
    \State $\mathcal{C}[\text{cell}] \leftarrow \text{cost}$
\EndFor
\end{algorithmic}
\end{algorithm}

\section{Navigation System and Path Planning}
\label{sec:navigation-planning}

The multimodal navigation system integrates the Nav2 framework for local trajectory control with a custom traversability-aware global planner that incorporates both terrain difficulty and energy consumption into path optimization. This architecture enables the robot to generate paths that balance geometric efficiency, terrain negotiability, and battery conservation.

\subsection{Nav2 Local Controller Configuration}
\label{subsec:nav2-local}

The system employs the Dynamic Window Approach (DWA) controller~\cite{fox1997dwa} from the Nav2 stack for local trajectory tracking and dynamic obstacle avoidance. The DWA controller operates at 20 Hz, sampling velocity commands within the robot's kinematic constraints to select trajectories that minimize a weighted combination of goal alignment, path deviation, and obstacle clearance. Key parameters include maximum linear velocity (0.5 m/s), maximum angular velocity (1.0 rad/s), and a minimum obstacle clearance radius of 0.2 m. The controller consumes the global path from the A* planner as a reference trajectory while reactively adjusting to local costmap changes detected within a 2.0 m lookahead window.

\subsection{Energy-Aware A* Global Planner}
\label{subsec:astar-planner}

While the DWA controller handles short-term reactive navigation, global path planning requires reasoning about terrain traversability and energy expenditure over longer distances. The custom A* implementation~\cite{hart1968astar} addresses this through a hybrid cost function that integrates geometric distance, terrain difficulty from the CNN-predicted costmap, and energy consumption based on the robot's power characteristics.

\paragraph{Cost Function Formulation:}
\label{subsubsec:cost-function}

For ground-based navigation, the cost of traversing one meter is determined by the robot's driving power consumption $P_d$ and velocity $v_d$:

\begin{equation}
C_d = \frac{P_d}{v_d}
\label{eq:ground_energy_cost}
\end{equation}

This energy cost per meter represents the battery drain rate during ground locomotion. For the M4 platform operating at nominal ground velocity ($v_d = 0.3$ m/s) with average driving power ($P_d \approx 45$ W), the energy cost is approximately $C_d = 150$ J/m.

The complete edge cost for transitioning from grid cell $i$ to neighboring cell $j$ combines three components:

\begin{equation}
c_{ij} = d_{ij} \left(1 + w_t \cdot T_{ij} + C_d \right)
\label{eq:hybrid_cost}
\end{equation}

where $d_{ij}$ represents the Euclidean distance between cell centers (1.0 or 1.414 for cardinal/diagonal moves at 0.1 m resolution), $T_{ij}$ is the normalized traversability cost from the costmap ($\in [0,1]$), $w_t = 20$ is the terrain penalty weight, and $C_d$ is the energy cost per meter from Equation~\ref{eq:ground_energy_cost}.

This formulation ensures that the planner simultaneously minimizes path length, avoids high-cost terrain eventually reducing mechanical stress, and reduces energy consumption. The terrain weight $w_t$ is tuned to balance these competing objectives—higher values prioritize smoother terrain over shorter paths.

\subsubsection{A* Search Implementation}
\label{subsubsec:astar-implementation}

Algorithm~\ref{alg:astar_planner} presents the A* planning procedure. The search maintains a priority queue ordered by $f(n) = g(n) + h(n)$, where $g(n)$ accumulates the hybrid cost from the start to node $n$, and $h(n)$ provides an admissible heuristic (Euclidean distance to goal)~\cite{hart1968astar}. The algorithm expands nodes with lowest $f$-value, computing edge costs via Equation~\ref{eq:hybrid_cost} and terminating when the goal is reached or a timeout occurs.

\begin{algorithm}[htbp]
\caption{Energy-Aware A* Path Planning}
\label{alg:astar_planner}
\begin{algorithmic}[1]
\State \textbf{Input:} Start $s$, Goal $g$, Costmap $\mathcal{C}$, Energy cost $C_d$, Terrain weight $w_t$
\State \textbf{Output:} Path $\pi$ from $s$ to $g$
\State
\State Initialize priority queue $Q \leftarrow \{(f(s), s, 0, [s])\}$ where $f(s) = h(s)$
\State Initialize closed set $V \leftarrow \emptyset$
\State
\While{$Q \neq \emptyset$}
    \State Pop node $(f, n, g_n, \text{path})$ with minimum $f$ from $Q$
    \If{$n \in V$}
        \State \textbf{continue}
    \EndIf
    \State $V \leftarrow V \cup \{n\}$
    \State
    \If{$n = g$}
        \State \Return $\text{path}$ \Comment{Goal reached}
    \EndIf
    \State
    \For{each neighbor $n'$ of $n$}
        \If{$n' \in V$ or $\mathcal{C}[n'] \geq 100$}
            \State \textbf{continue} \Comment{Skip visited/lethal cells}
        \EndIf
        \State
        \State $d \leftarrow \|n' - n\| \cdot r$ \Comment{Distance in meters, resolution $r$}
        \State $T \leftarrow \mathcal{C}[n'] / 100$ \Comment{Normalized terrain cost}
        \State $c \leftarrow d \cdot (1 + w_t \cdot T + C_d)$ \Comment{Hybrid edge cost}
        \State
        \State $g_{n'} \leftarrow g_n + c$
        \State $h_{n'} \leftarrow \|n' - g\| \cdot r$ \Comment{Euclidean heuristic}
        \State Push $(g_{n'} + h_{n'}, n', g_{n'}, \text{path} + [n'])$ into $Q$
    \EndFor
\EndWhile
\State
\State \Return \texttt{NULL} \Comment{No path found}
\end{algorithmic}
\end{algorithm}

The planner operates at 5 Hz with continuous replanning enabled, allowing the robot to adapt to dynamic obstacles or newly mapped terrain as the costmap updates. The resulting path undergoes line-of-sight pruning to remove redundant intermediate waypoints, reducing trajectory length while maintaining path validity. This pruning step iteratively checks if direct line segments between non-adjacent waypoints remain collision-free, eliminating unnecessary corners from the grid-based A* output. Local trajectory smoothing is subsequently handled by the DWA controller, which interpolates continuous velocity commands between pruned waypoints while respecting kinematic constraints.

\subsection{DWB Local Planner}

The Dynamic Window B (DWB) controller selects optimal velocity commands by sampling trajectories within a dynamic window and minimizing a weighted objective function. The dynamic window defines reachable velocities given current velocity and acceleration limits:

\begin{align}
\mathcal{V}_{dw} &= [v - a_{max} \Delta t, \ v + a_{max} \Delta t] \cap [v_{min}, v_{max}] \\
\Omega_{dw} &= [\omega - \alpha_{max} \Delta t, \ \omega + \alpha_{max} \Delta t] \cap [\omega_{min}, \omega_{max}]
\end{align}

\noindent where $v$ and $\omega$ are current linear and angular velocities, $a_{max}$ and $\alpha_{max}$ are acceleration limits, and $\Delta t$ is the control period.

For each sampled trajectory $\tau \in \mathcal{T}$, the total cost is computed as:

\begin{equation}
J(\tau) = \sum_{i=1}^{n} w_i \cdot C_i(\tau)
\end{equation}

\noindent where $w_i$ represents the weight for critic $i$ and $C_i(\tau)$ is the cost assigned by that critic. The individual critics used in our implementation are:

\begin{align}
C_{path}(\tau) &= \min_{p \in P} \|\tau_{end} - p\|_2 \label{eq:path_cost} \\
C_{goal}(\tau) &= \|\tau_{end} - g\|_2 \label{eq:goal_cost} \\
C_{align}(\tau) &= |\theta_\tau - \text{atan2}(g_y - \tau_y, g_x - \tau_x)| \label{eq:align_cost} \\
C_{obs}(\tau) &= \max_{t \in [0,T]} \mathcal{M}(x_t, y_t) \label{eq:obs_cost}
\end{align}

\noindent where $P$ is the global path from the A* planner, $g$ is the goal position, $\theta_{\tau}$ is the trajectory end heading, and $\mathcal{M}(x,y)$ returns the local costmap value at position $(x,y)$.

The optimal trajectory is selected as:

\begin{equation}
\tau^* = \underset{\tau \in \mathcal{T}}{\arg\min} \ J(\tau)
\end{equation}

\noindent and the corresponding velocity command $(v^*, \omega^*)$ is sent to the robot controller.

\subsection{Multimodal Decision Framework}
\label{subsec:multimodal-decision}

For scenarios where ground navigation encounters insurmountable obstacles or prohibitively high terrain costs, the system incorporates a decision framework that evaluates whether aerial locomotion provides a more efficient alternative. This decision node operates as a supervisory layer monitoring both the A* planner output and the Nav2 execution status.

\subsubsection{Cost-Based Mode Selection}
\label{subsubsec:cost-mode-selection}

The primary decision criterion compares the total energy cost of ground navigation against aerial bypass. The ground path cost $C_{\text{ground}}$ is computed by summing the hybrid edge costs along the A* trajectory:

\begin{equation}
C_{\text{ground}} = \sum_{i=1}^{|\pi|} d_{i,i+1} \cdot \frac{c_{i,i+1}}{2}
\label{eq:ground_cost}
\end{equation}

The aerial cost considers direct flight distance, flight power consumption, and transformation overhead:

\begin{equation}
C_{\text{aerial}} = d_{\text{direct}} \cdot \frac{P_f}{v_f} + 2C_t
\label{eq:aerial_cost}
\end{equation}

where $P_f = 600$ W is the flight power, $v_f = 1.5$ m/s is the flight velocity, and $C_t = 300$ J is the transformation cost. The factor of 2 in the transformation term accounts for both the takeoff transformation and the landing transformation required for a complete aerial segment.

The decision framework triggers aerial mode recommendation under the following conditions:

\textbf{Failure Recovery Conditions:}
\begin{equation}
\text{Switch to aerial if: } \begin{cases}
|\omega_z| > 0.3 \land |v_x| < 0.05 \text{ for } t > 1\text{s} \text{ (twice)} & \text{(spin detection)} \\
\Delta d_{\text{goal}} < 0.1\text{m for } t > 10\text{s} & \text{(stuck detection)} \\
|\mathcal{A}_{20\text{s}}| \geq 2 & \text{(retry loop)} \\
|\pi_{\text{ground}}| = 0 \text{ (twice)} & \text{(no path found)} \\
\max(c_i) > 80 & \text{(impassable terrain)}
\end{cases}
\label{eq:failure_conditions}
\end{equation}

\textbf{Cost Optimization:}
\begin{equation}
m_{k+1} = \begin{cases}
\text{aerial} & m_k = \text{ground} \land C_{\text{aerial}} < C_{\text{ground}} \\
\text{ground} & m_k = \text{aerial} \land C_{\text{ground}} < 0.8 \cdot C_{\text{aerial}} \\
m_k & \text{otherwise}
\end{cases}
\label{eq:mode_selection}
\end{equation}

The hysteresis factor ($\gamma = 0.8$) prevents oscillation between modes when costs are similar.

The decision node subscribes to the Nav2 action server status and continuously monitors for \texttt{STATUS\_ABORTED} messages, which indicate that the DWA controller encountered an unrecoverable situation (e.g., trapped by dynamic obstacles, local minima in the costmap). Upon detecting a failure, the framework immediately suggests aerial mode as a recovery strategy, bypassing the cost comparison since ground navigation has demonstrably failed.

\subsubsection{Current Implementation Status}
\label{subsubsec:implementation-status}

The decision framework is currently implemented as a monitoring and recommendation system. When aerial mode is determined to be advantageous, the node outputs a terminal message indicating the suggested mode switch along with the cost differential or failure reason, and the servos transform to aerial mode. The actual execution of aerial locomotion, quadcopter motor control, and aerial trajectory tracking remains outside the scope of this thesis and is reserved for future integration work. This design separation allows the decision logic to be validated and tuned independently before full multimodal hardware deployment.

% testing and discussion
% Chapter 5: Experimental Results and System Validation

\chapter{RESULTS}
\label{chap:results}

\section{IMU Calibration Results}
\label{sec:results}

The Allan variance analysis provided accelerometer and gyroscope noise parameters that align with expected characteristics for the Ouster REV3 integrated IMU. The curves showed the standard white-noise region, followed by bias-instability behavior, and finally the long-term random walk region.

The extracted parameters used for SLAM configuration were:

\begin{table}[htbp]
\centering
\caption{IMU noise parameters extracted from Allan variance analysis.}
\label{tab:imu-noise-params}
\begin{tabular}{|l|c|c|}
\hline
\textbf{Parameter} & \textbf{Value} & \textbf{Unit} \\
\hline
Accelerometer noise density & 0.01161 & m/s$^2$/$\sqrt{\text{Hz}}$ \\
\hline
Accelerometer random walk & $2.15 \times 10^{-4}$ & m/s$^3$/$\sqrt{\text{Hz}}$ \\
\hline
Gyroscope noise density & $4.05 \times 10^{-4}$ & rad/s/$\sqrt{\text{Hz}}$ \\
\hline
Gyroscope random walk & $1.26 \times 10^{-5}$ & rad/s$^2$/$\sqrt{\text{Hz}}$ \\
\hline
\end{tabular}
\end{table}

\begin{figure}[htbp]
    \centering
    \includegraphics[width=0.78\textwidth, trim=10 10 10 75, clip]{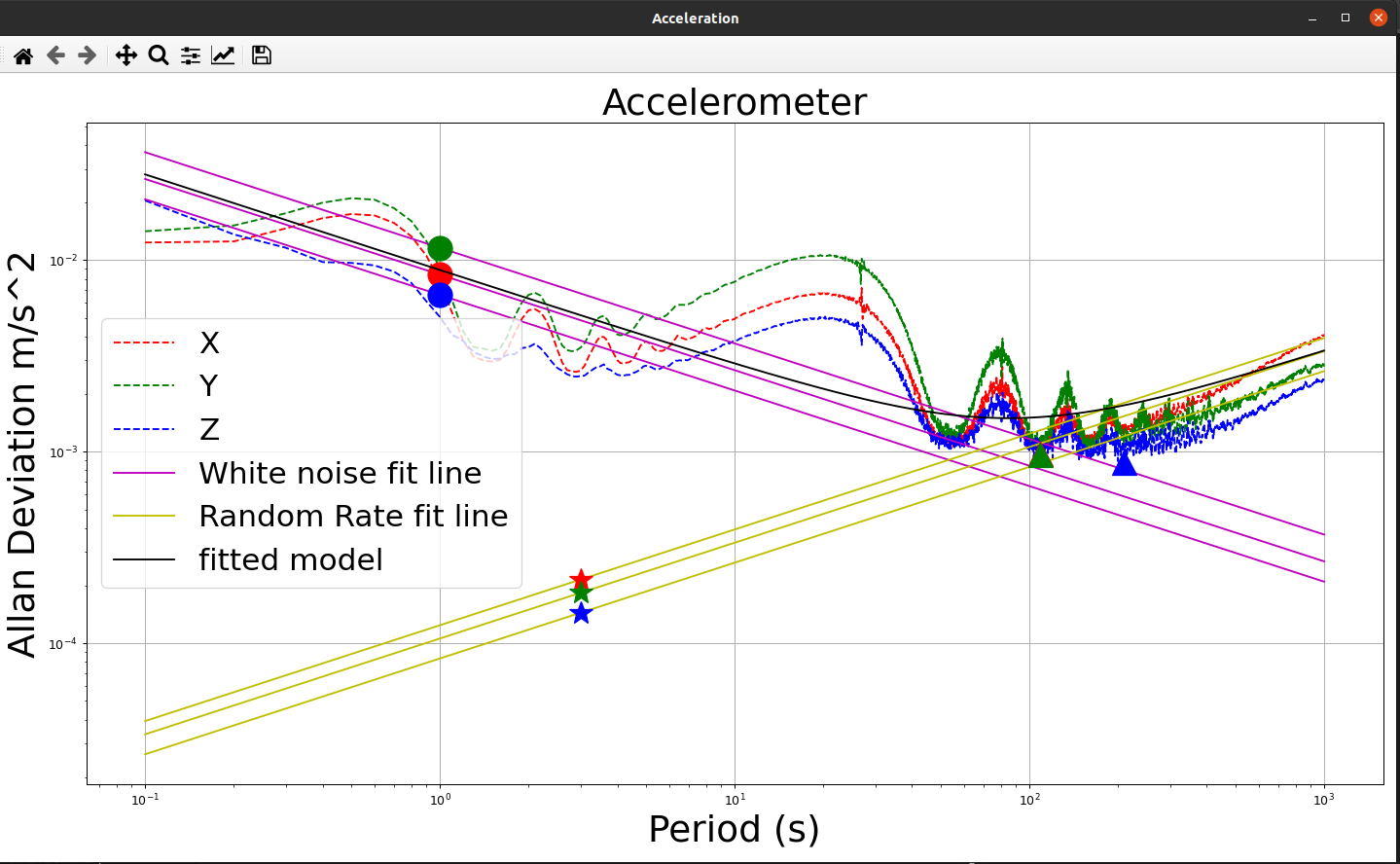}
    \caption{Accelerometer Allan deviation for Ouster REV3 IMU.}
\end{figure}

\begin{figure}[htbp]
    \centering
    \includegraphics[width=0.78\textwidth, trim=10 10 10 75, clip]{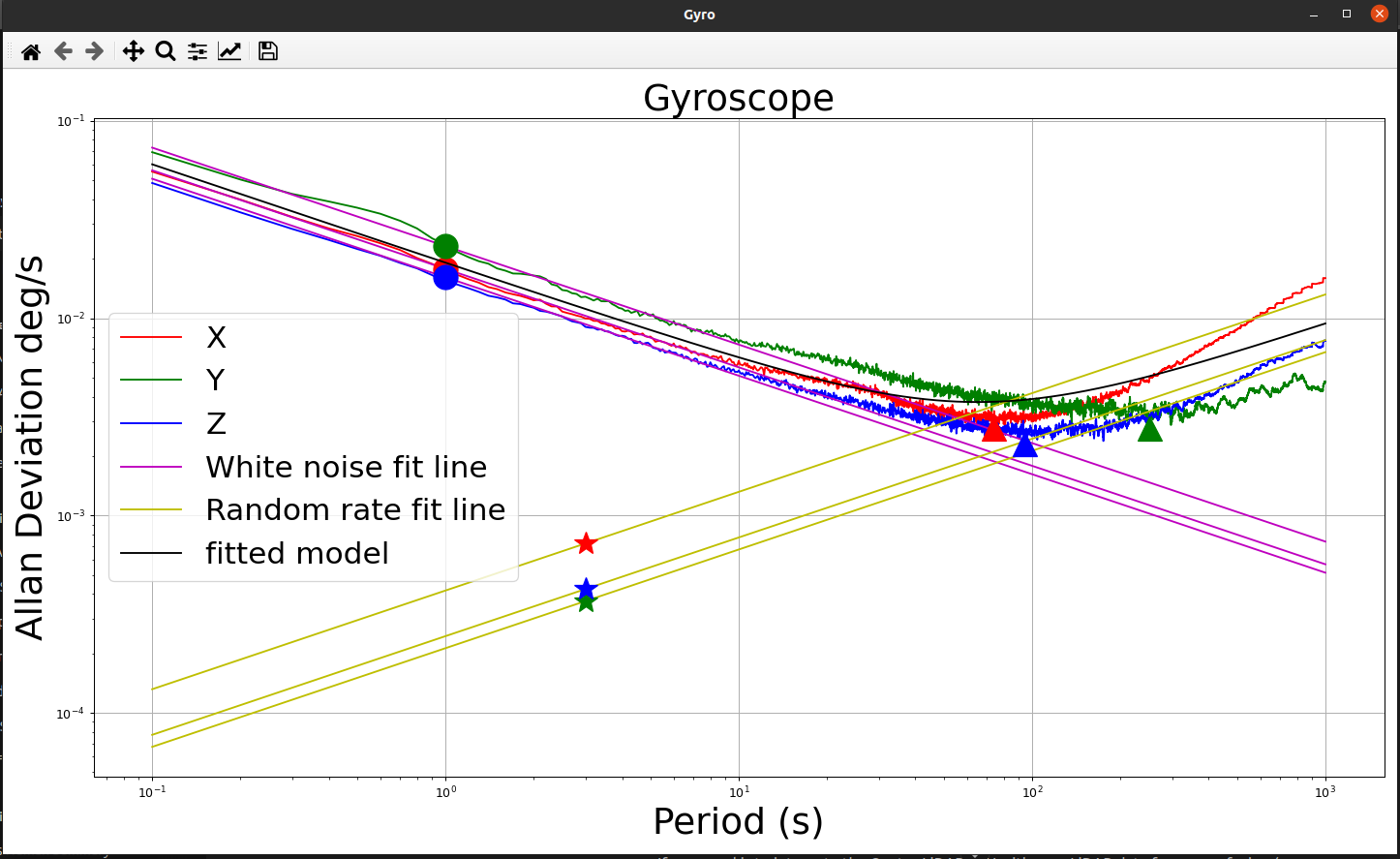}
    \caption{Gyroscope Allan deviation for Ouster REV3 IMU.}
\end{figure}

\section{Trajectory Accuracy Results}
\label{sec:results-trajectories}

Trajectory error metrics were computed using \texttt{evo} with SE(3) Umeyama alignment. LIO-SAM tracked the OptiTrack trajectory with stable drift characteristics. The overall RMSE was approximately 10.4\,cm across all runs.

\begin{figure}[htbp]
    \centering
    \includegraphics[width=0.55\textwidth]{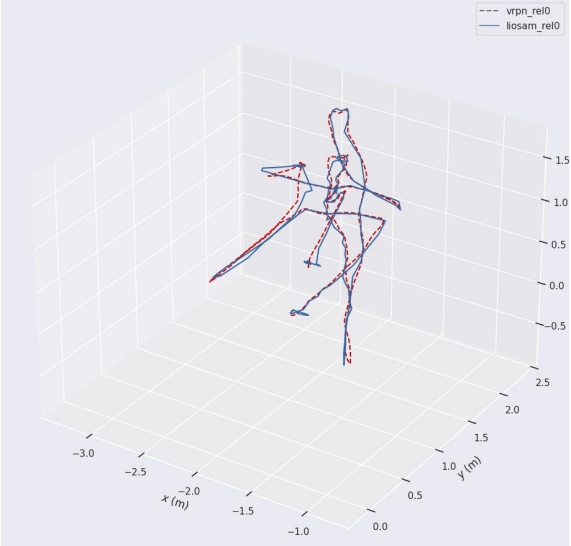}
    \caption{Comparison of LIO-SAM (blue) and OptiTrack ground truth (red).}
    \label{fig:lidas-optitrack}
\end{figure}

RTAB-Map produced a globally consistent map when loop closure was enabled; however, local drift appeared in rapid rotations or visually sparse regions.

\section{Point Cloud Geometric Quality}
\label{sec:results-c2c}

\begin{figure}[htbp]
    \centering
    \includegraphics[width=0.75\textwidth]{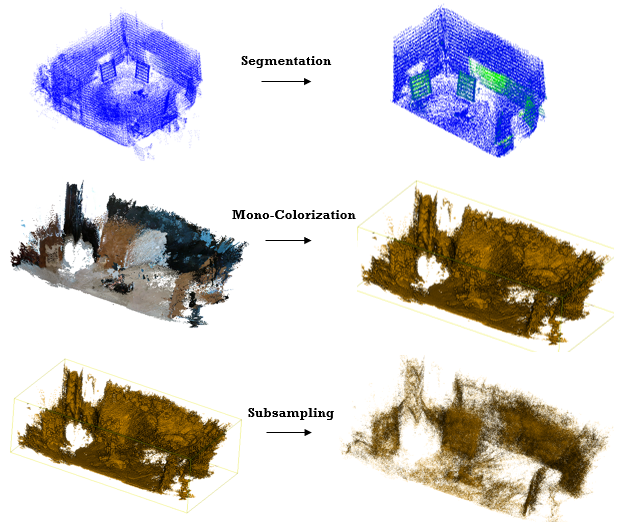}
    \caption{Pre-processing operations on pointclouds}
    \label{fig:preprocessing}
\end{figure}

Before performing cloud-to-cloud distance analysis, both point clouds underwent a three-stage preprocessing pipeline illustrated in Figure~\ref{fig:preprocessing}. First, segmentation isolated specific regions of interest from the full environment scan, enabling targeted geometric comparison without interference from distant structures. Second, mono-colorization removed RGB color information from the RTAB-Map reconstructions, ensuring that subsequent ICP registration relied purely on geometric features rather than appearance. Finally, subsampling reduced point density to a uniform resolution, balancing computational efficiency with geometric fidelity for the alignment and distance computation stages.

Cloud-to-cloud distance analysis quantified deviation between RTAB-Map’s reconstruction and LIO-SAM’s LiDAR reference. Voxel downsampling reduced point count from over 2 million to approximately 541k (62\% reduction), preserving essential structure.

\begin{figure}[htbp]
    \centering
    \includegraphics[width=0.55\textwidth]{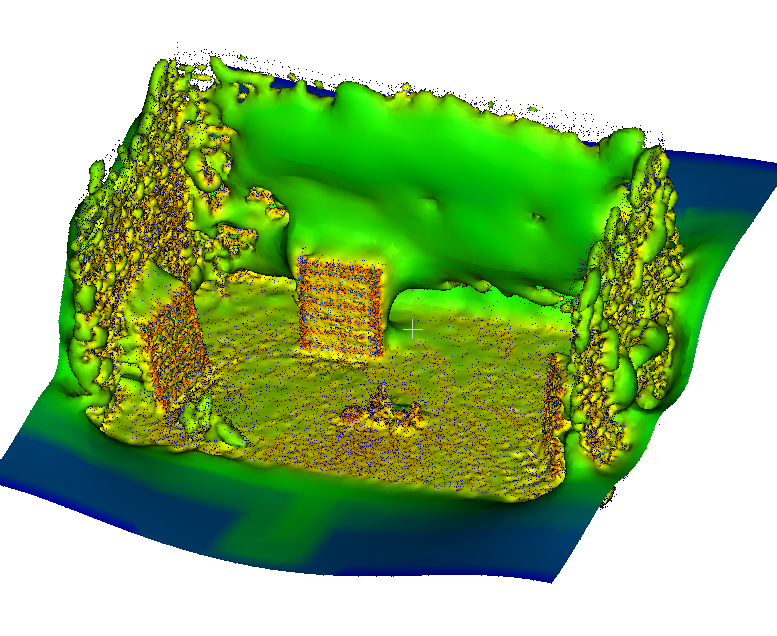}
    \caption{Delaunay triangulation mesh generated for Rtabmap to show an example.}
    \label{fig:triangulation-mesh}
\end{figure}

Following preprocessing, the LIO-SAM point cloud was converted into a triangulated mesh using Delaunay triangulation. This mesh served as the reference surface against which RTAB-Map point distances were computed, enabling geometry-aware comparison rather than simple point-to-point heuristics.

\section{LiDAR vs Camera: Suitability for Elevation Mapping}
\label{sec:results-lidar-vs-camera}

Across all scenarios, RTAB-Map’s reconstruction showed average deviations of 5–7\,cm and peak deviations above 30\,cm in occluded regions. These values approach or exceed the 5–10\,cm height differences M4 must reliably detect to determine ground traversability.

LIO-SAM consistently achieved \textasciitilde4\,cm error on planar surfaces and retained good geometric structure even in cluttered scenes. Since elevation mapping requires stable height measurements independent of lighting and texture, LiDAR was selected as the primary sensing modality for the M4 traversability pipeline.

\begin{figure}[htbp]
    \centering
    \includegraphics[width=0.7\textwidth]{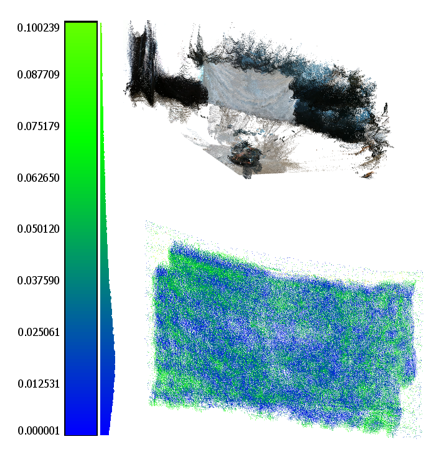}
    \caption{Mesh-to-cloud distance heatmap for planar surface.}
    \label{fig:c2c-planar}
\end{figure}

\begin{figure}[htbp]
    \centering
    \includegraphics[width=0.7\textwidth]{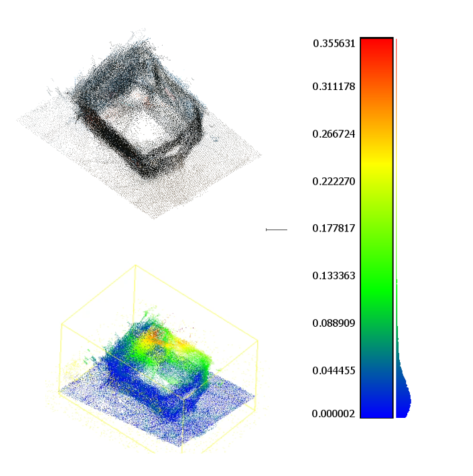}
    \caption{Mesh-to-cloud distance heatmap for isolated box.}
    \label{fig:c2c-box}
\end{figure}

\begin{figure}[htbp]
    \centering
    \includegraphics[width=0.7\textwidth]{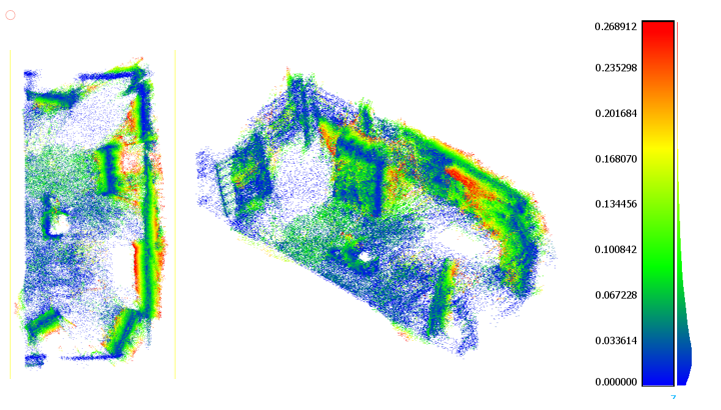}
    \caption{Mesh-to-cloud distance heatmap for occluded area.}
    \label{fig:c2c-occluded}
\end{figure}

\begin{figure}[htbp]
    \centering
    \includegraphics[width=0.7\textwidth]{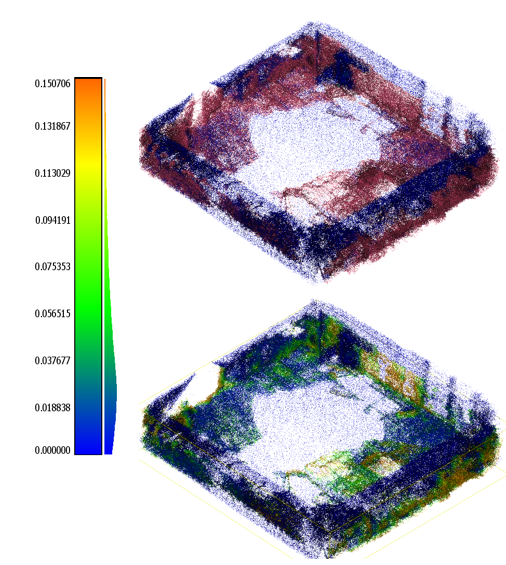}
    \caption{Mesh-to-cloud distance heatmap for full cage environment. Blue = close alignment (0--3\,cm), red = high deviation.}
    \label{fig:c2c-fullcage}
\end{figure}

Quantitative error statistics are summarized in Table~\ref{tab:c2c-summary}.

\begin{table}[htbp]
\centering
\caption{Summary of RTAB-Map deviation from LIO-SAM.}
\label{tab:c2c-summary}
\begin{tabular}{|l|c|c|c|}
\hline
\textbf{Scenario} & \textbf{Mean (cm)} & \textbf{Std Dev (cm)} & \textbf{Max Error (cm)} \\
\hline
Planar surface & 3.9 & 3.0 & 10 \\
Isolated box & 5.9 & 6.5 & 35 \\
Occluded area & 7.0 & 6.7 & 26 \\
Full cage map & 6.6 & 6.9 & 15 \\
\hline
\end{tabular}
\end{table}

\section{FAST-LIO Localization Performance}

\subsection{Point Cloud Map Quality}

\begin{figure}[htbp]
    \centering
    \includegraphics[width=0.55\textwidth]{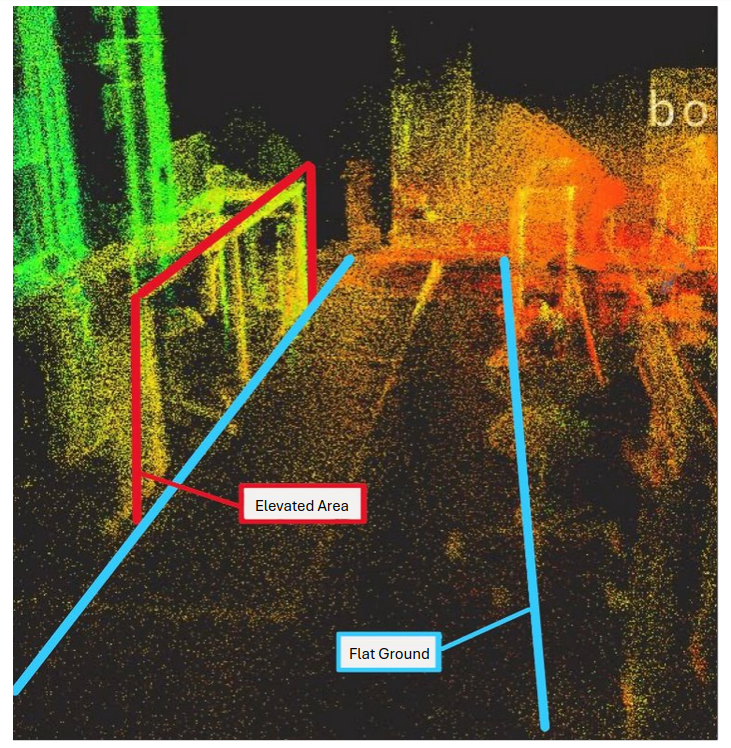}
    \caption{FAST-LIO generated point cloud map from Highbay .}
    \label{fig:fastlio_map}
\end{figure}

Figure~\ref{fig:fastlio_map} presents the point cloud map generated during navigation experiments in the Highbay Laboratory. The reconstruction exhibits clean, consistent geometry of environment features including walls, floor surfaces, and obstacle boundaries. The map demonstrates minimal noise and clear representation of planar surfaces, which is essential for accurate elevation extraction. Ground plane surfaces show uniform point density with no visible drift-induced deformations or feature duplication, confirming that the localization remains stable throughout the mapping process.

The quality of floor surface reconstruction is particularly important, as elevation mapping relies on accurate ground height measurements to establish the baseline reference from which obstacle heights and terrain variations are computed. The observed point cloud density and vertical consistency in floor regions provide confidence that the subsequent elevation mapping pipeline receives high-fidelity geometric input, enabling reliable traversability assessment.

\subsection{Trajectory Consistency and Loop Closure}

\begin{figure}[htbp]
    \centering
    \begin{subfigure}[b]{0.48\textwidth}
        \centering
        \includegraphics[width=\textwidth]{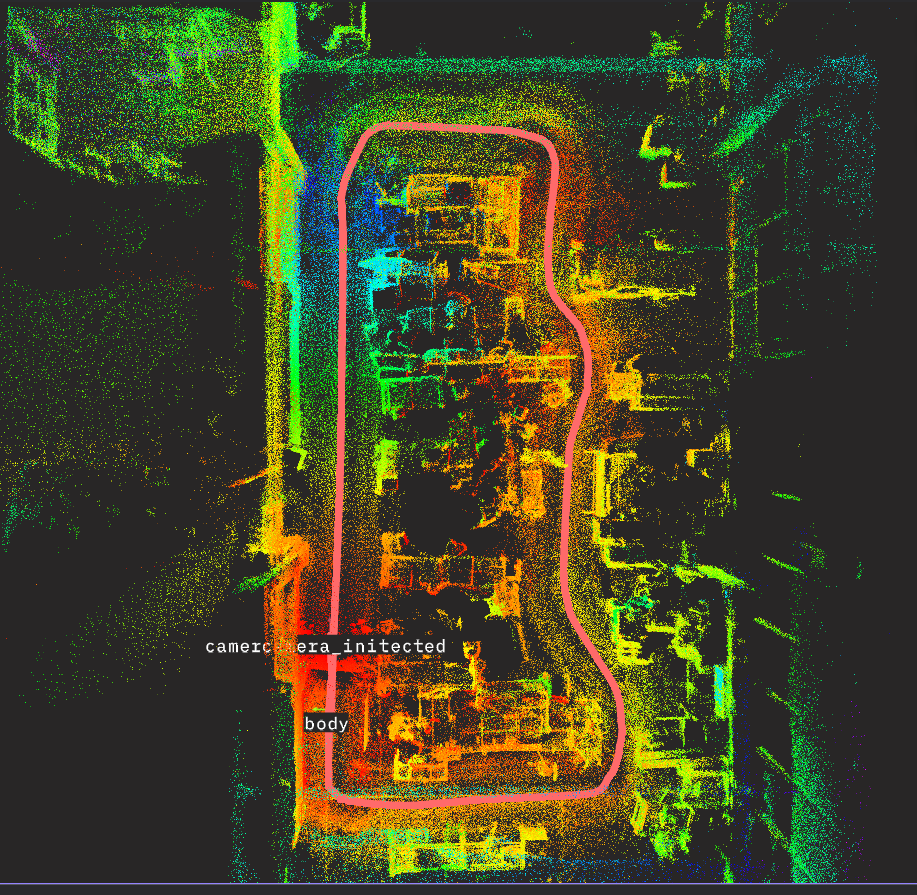}
        \caption{Trajectory with map}
        \label{fig:traj_with_map}
    \end{subfigure}
    \hfill
    \begin{subfigure}[b]{0.48\textwidth}
        \centering
        \includegraphics[width=\textwidth]{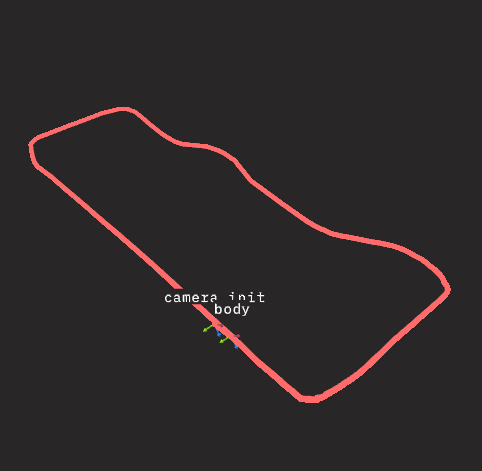}
        \caption{Isolated trajectory path}
        \label{fig:traj_only}
    \end{subfigure}
    \caption{Robot trajectory showing closed-loop path through Highbay.}
    \label{fig:fastlio_trajectory_loop}
\end{figure}

To evaluate localization consistency over extended operation, the robot was commanded to execute loop trajectories returning to previously mapped regions. Figure~\ref{fig:fastlio_trajectory_loop} shows the recorded trajectory with the robot traversing an approximate closed loop path through the laboratory environment. Visual inspection of the trajectory reveals smooth, continuous motion tracking without discontinuities or unrealistic jumps, indicating stable pose estimation throughout the sequence.

Critically, when the robot returns to previously mapped areas, the point cloud map exhibits no detectable feature duplication or misalignment. This indicates that accumulated drift over the loop duration remains negligible, confirming that FAST-LIO provides a stable world reference frame suitable for persistent elevation mapping.

\section{Elevation Mapping and Traversability Estimation}

The elevation mapping framework was tested in the Highbay Laboratory to validate terrain representation and traversability classification during real-time operation.

\subsection{Multi-Layer Map Output}

\begin{figure}[htbp]
    \centering
    \begin{subfigure}[b]{0.48\textwidth}
        \centering
        \includegraphics[width=\textwidth]{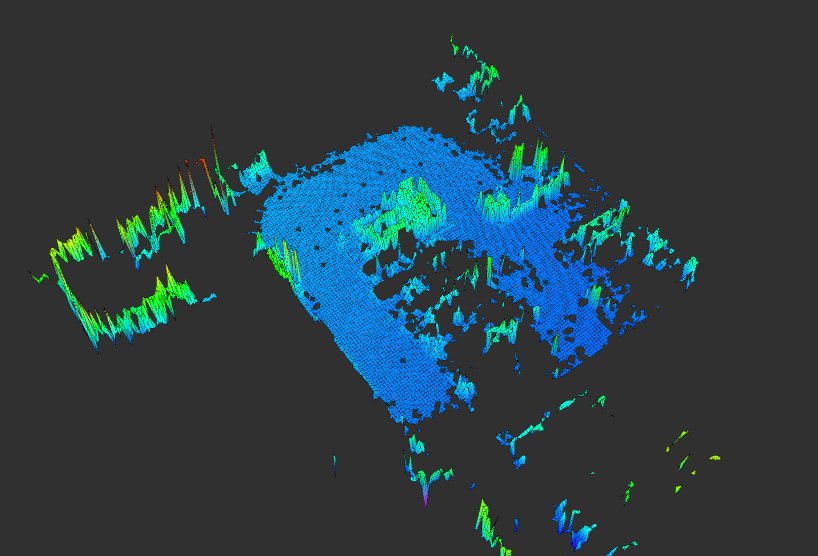}
        \caption{Elevation layer (height)}
        \label{fig:elev_height}
    \end{subfigure}
    \hfill
    \begin{subfigure}[b]{0.48\textwidth}
        \centering
        \includegraphics[width=\textwidth]{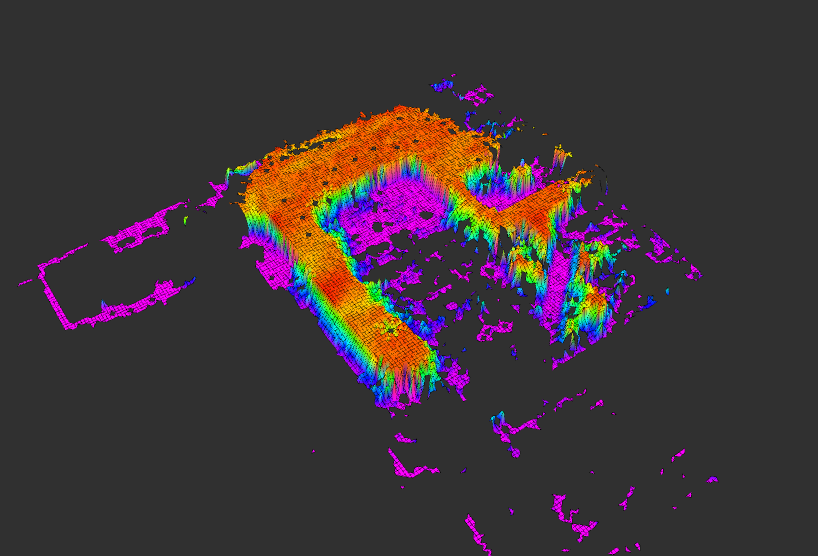}
        \caption{Traversability layer (CNN)}
        \label{fig:trav_cnn}
    \end{subfigure}
    \caption{Multi-layer elevation map from Highbay Laboratory.}
    \label{fig:elevation_layers}
\end{figure}

Figure~\ref{fig:elevation_layers} shows the elevation and traversability layers generated during testing. The elevation layer captures terrain height at 0.04 m resolution, clearly showing floor surfaces and obstacle boundaries. The traversability layer displays CNN predictions where open floor areas receive high scores (red/yellow), indicating safe ground navigation, while obstacles and elevation changes receive low scores (blue/purple), marking them as non-traversable.

The system processed the MID-360 LiDAR output at 15-18 Hz on the laptop GPU, and was able to update the map layers at the same rate, maintaining real-time terrain awareness for navigation planning.

\section{Offline Path Planning Validation}

The traversability-aware path planning system was validated offline using map data collected in the Highbay Laboratory before deployment on the physical M4 platform.

A test environment was set up with a black cloth placed on the floor to simulate low-traversability terrain, along with small obstacles and wooden planks creating a constrained pathway. The environment was mapped using the elevation mapping pipeline, generating a costmap with both obstacle information and CNN-predicted traversability scores. Two planning scenarios were defined where the shortest geometric path passed over the cloth region, while alternative routes required longer distances.

\begin{figure}[htbp]
    \centering
    \includegraphics[width=0.80\textwidth]{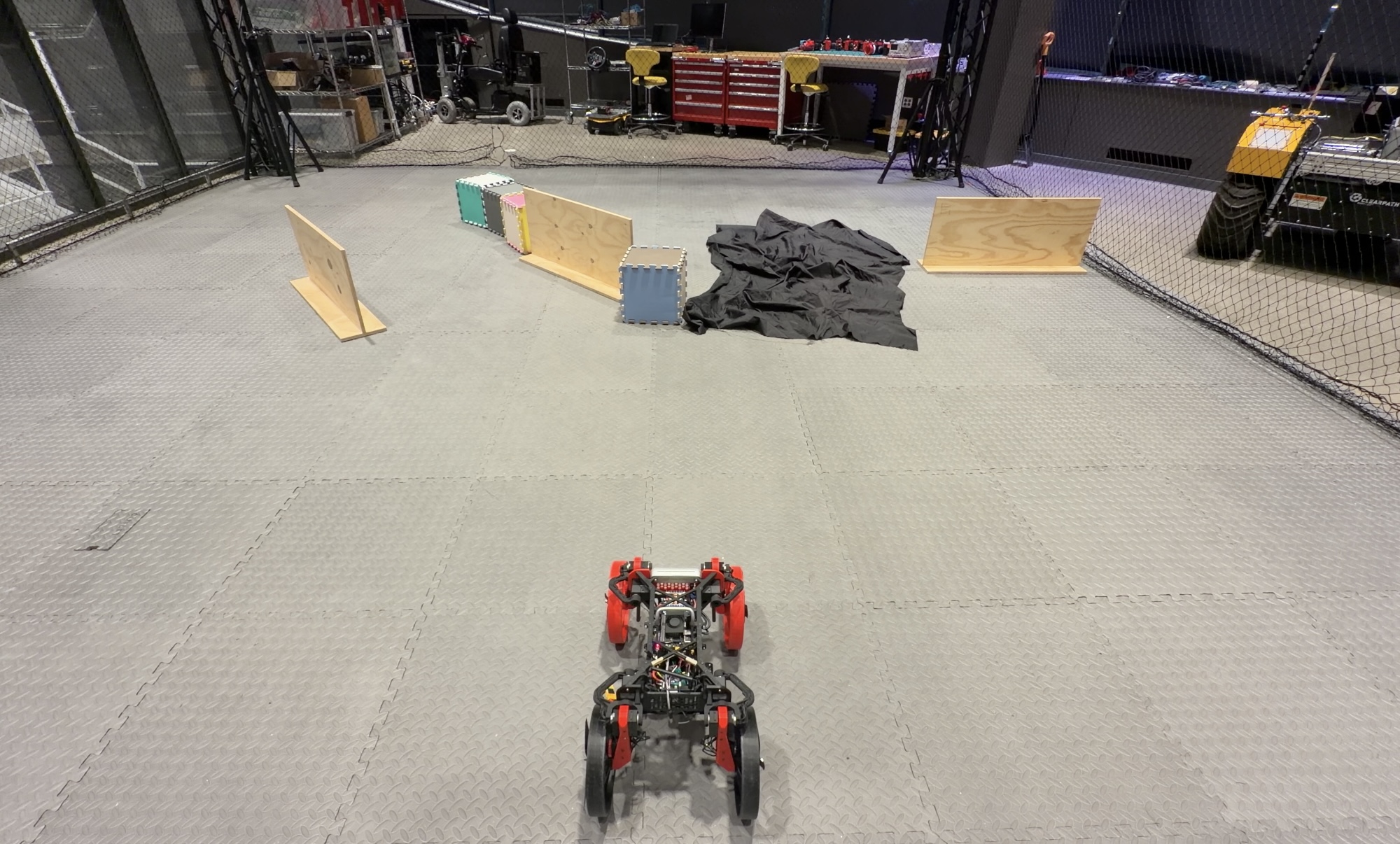}
    \caption{Offline path planning environment setup.}
    \label{fig:offline_setup}
\end{figure}

For each scenario, paths were computed using two approaches: a traditional occupancy-based planner treating all non-obstacle cells as equally traversable, and the proposed traversability-aware planner using the hybrid cost function with terrain weight $w_t = 8.0$.

\subsection{Results}

\begin{figure}[htbp]
    \centering
    \includegraphics[width=1.0\textwidth]{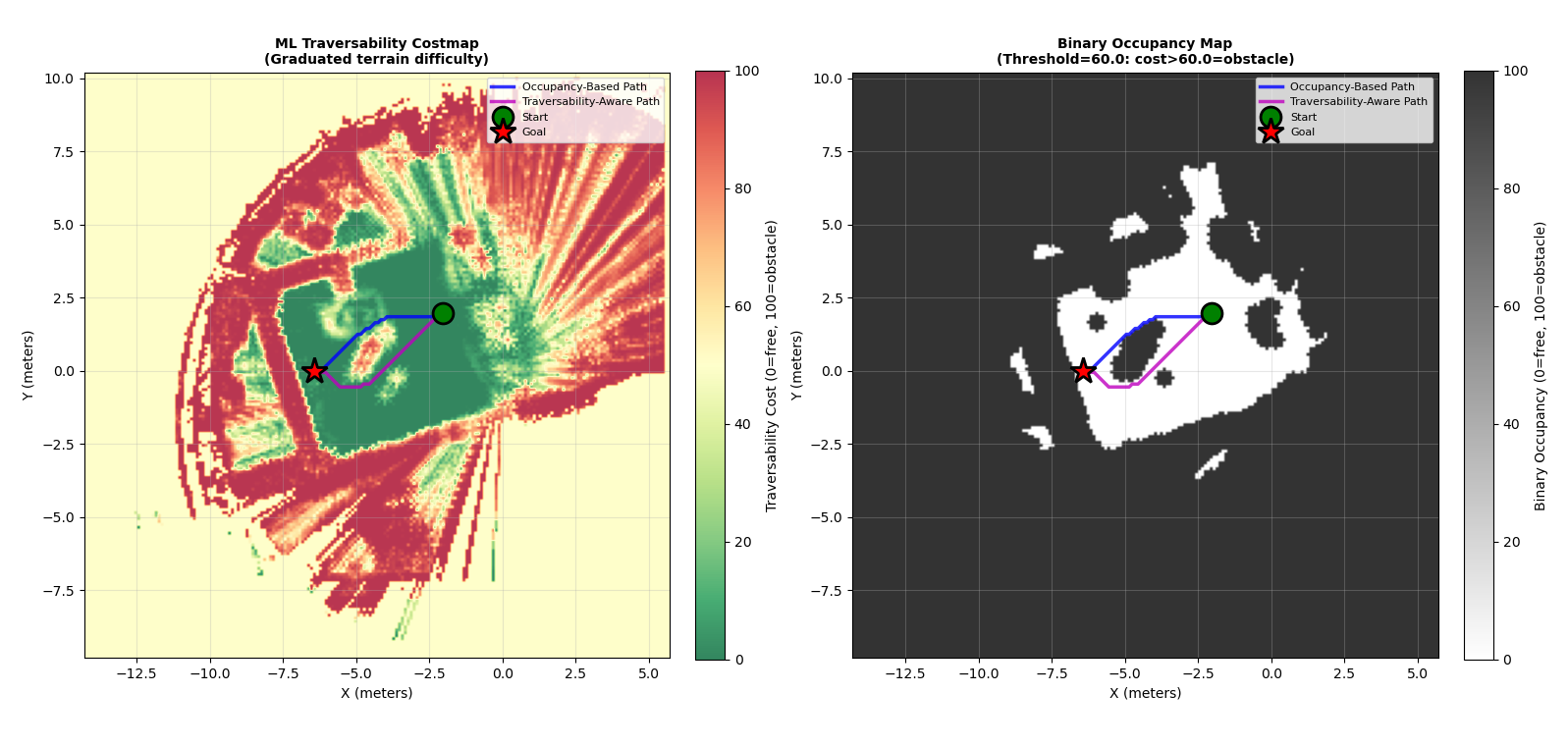}
    \caption{A* on a goal.}
    \label{fig:offline_test1}
\end{figure}

Figure~\ref{fig:offline_test1} shows results from the first scenario. The occupancy-based planner produced a 5.23 m path over the cloth with terrain cost of 16.71. The traversability-aware planner generated a 5.64 m path (+7.9\% distance) routing around the cloth, achieving terrain cost of 0.30, a 98.2\% reduction.

\begin{figure}[htbp]
    \centering
    \includegraphics[width=1.0\textwidth]{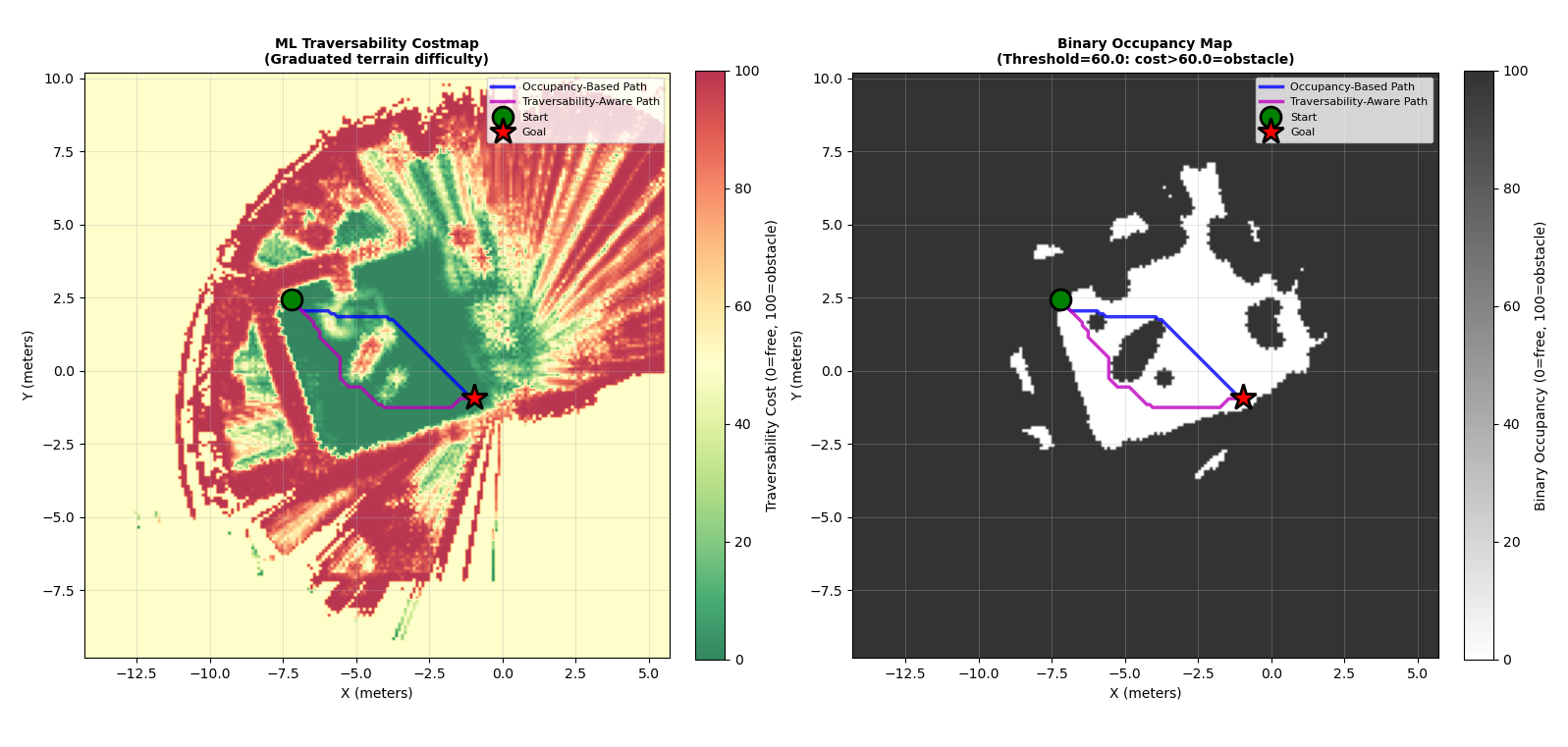}
    \caption{A* on different goal}
    \label{fig:offline_test2}
\end{figure}

Figure~\ref{fig:offline_test2} shows the second scenario with a different start-goal configuration. The occupancy-based path (7.71 m, terrain cost 38.99) again prioritizes geometric efficiency, while the traversability-aware path (8.54 m, terrain cost 0.99) shows consistent behavior: +10.8\% distance for 97.5\% terrain cost reduction.

These offline tests validate that the planner appropriately trades modest distance increases (8--11\%) for substantial terrain cost reductions (\textgreater 97\%), confirming the system behaves as designed before physical deployment.

\section{Nav2 Baseline Navigation}

Before implementing the traversability-aware global planner, the navigation system was tested using Nav2's default SMAC planner to verify basic obstacle avoidance and path following capabilities. This baseline validation ensured that the costmap generation, localization, and local controller integration were functioning correctly.
\begin{figure}[htbp]
    \centering
    \includegraphics[width=0.45\textwidth]{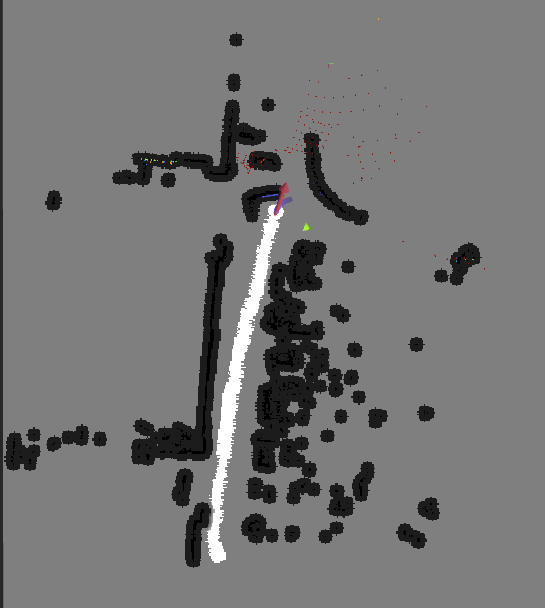}
    \caption{Nav2 baseline navigation cost-map.}
    \label{fig:nav2_baseline}
\end{figure}

Figure~\ref{fig:nav2_baseline} shows the navigation system operating with the default planner. The costmap correctly identifies obstacles from the traversability layer, and the global path (green) routes around occupied regions. The local controller successfully tracks the planned path while reactively adjusting for minor deviations and dynamic changes.

With baseline navigation validated, the default SMAC planner was replaced with the custom traversability-aware A* planner described in Section~\ref{subsec:astar-planner}, enabling terrain-quality-based path optimization in subsequent experiments.

\section{Integrated Navigation Performance in Complex Environment}

To validate the complete multimodal navigation system under realistic operating conditions, the M4 platform was deployed in a highly cluttered region of the Highbay Laboratory featuring dense obstacle arrangements representative of unstructured indoor environments.

\subsection{Test Environment and Experimental Setup}

\begin{figure}[htbp]
    \centering
    \includegraphics[width=0.7\textwidth]{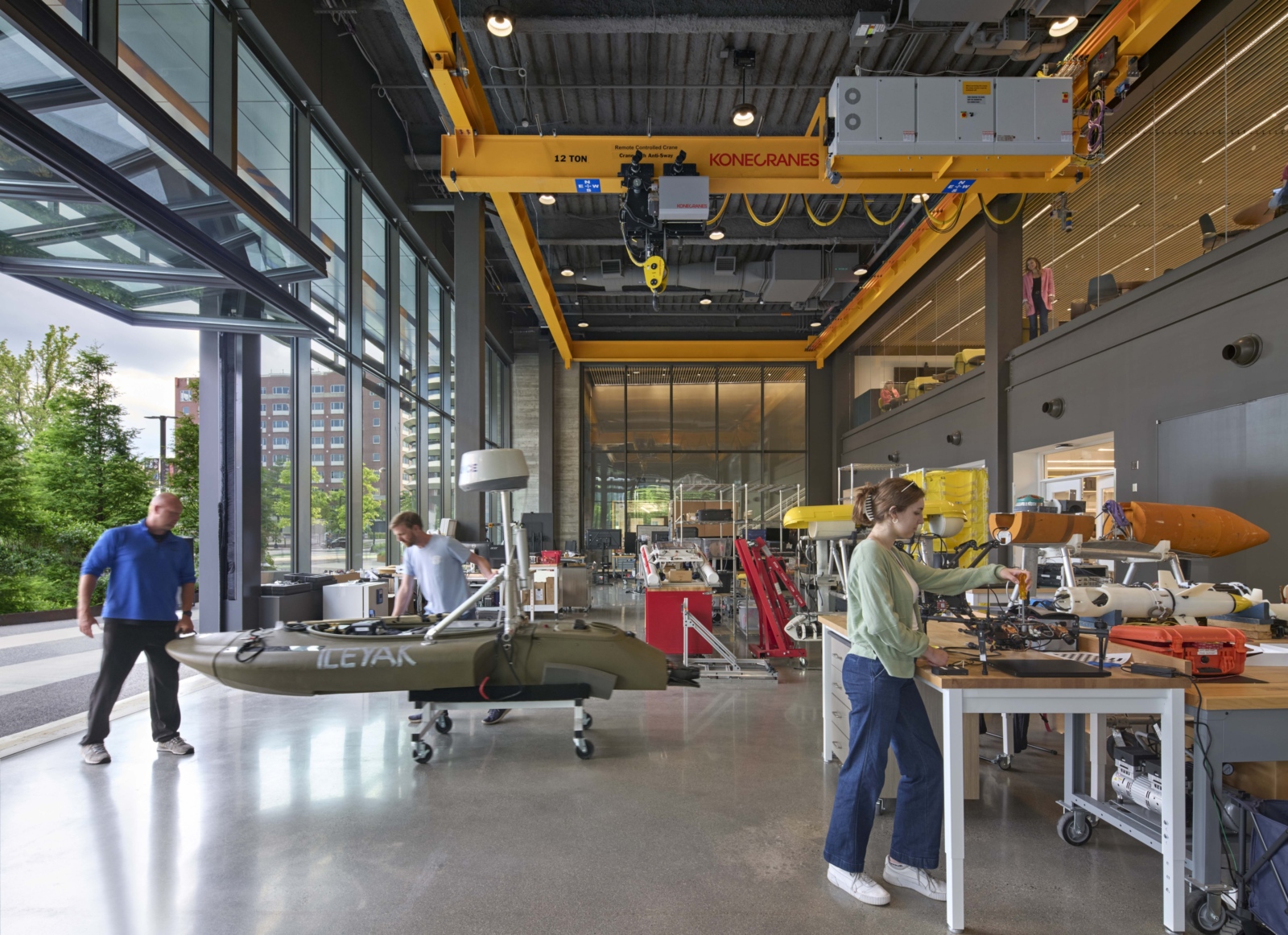}
    \caption{Test environment in Highbay Laboratory at Northeastern University.}
    \label{fig:complex_env}
\end{figure}

The test environment comprised a 15 m $\times$ 10 m area populated with tables, chairs, wire management ramps, cardboard boxes, and various laboratory equipment positioned at irregular intervals. Figure~\ref{fig:complex_env} shows a wide-angle view of the test space, illustrating the challenging navigation scenario with narrow passages, occlusions, and heterogeneous obstacle types that exercise both the traversability estimation and path planning components.

Navigation goals were specified interactively using the RViz 2D Goal Pose tool, which publishes target poses on the \texttt{/goal\_pose} topic. Upon receiving a goal, the A* planner computes an initial path based on the current traversability costmap, and the Nav2 DWA controller begins trajectory execution. The planner operates in continuous replanning mode at 5 Hz, allowing the system to adapt as the robot explores and the elevation mapping system updates the costmap with newly observed terrain.

\subsection{Elevation Mapping and Traversability Assessment in Test Environment}

\begin{figure}[htbp]
    \centering
    \begin{subfigure}[b]{0.48\textwidth}
        \centering
        \includegraphics[width=\textwidth]{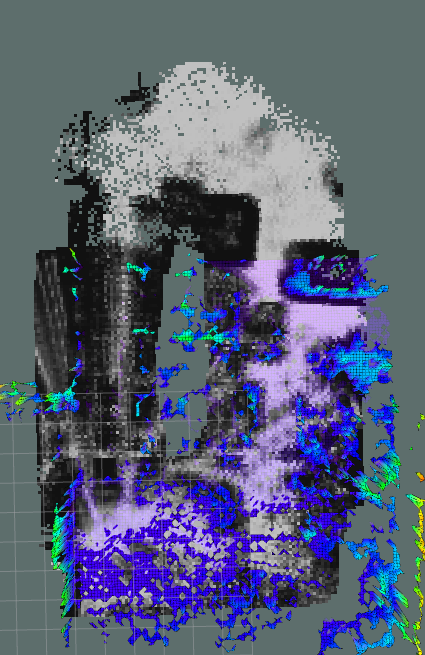}
        \caption{Elevation layer}
        \label{fig:test_elev}
    \end{subfigure}
    \hfill
    \begin{subfigure}[b]{0.48\textwidth}
        \centering
        \includegraphics[width=\textwidth]{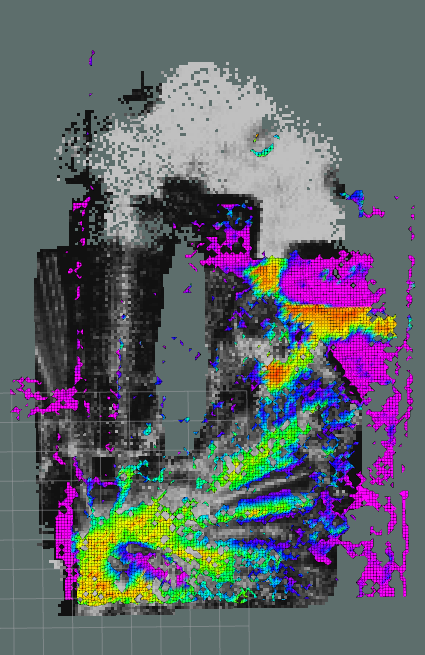}
        \caption{Traversability layer}
        \label{fig:test_trav}
    \end{subfigure}
    \caption{Elevation and traversability maps from Highbay navigation test}
    \label{fig:test_elevation_trav}
\end{figure}

Figure~\ref{fig:test_elevation_trav} presents the elevation and traversability maps generated during the Highbay navigation experiment. The elevation layer captures the geometric structure of the environment including floor surfaces, obstacle boundaries, and elevation discontinuities with 0.04 m resolution. The traversability layer demonstrates the CNN's real-time classification of terrain navigability, with open floor areas receiving high scores (green, $T > 0.85$) and obstacle regions marked as non-traversable (red, $T < 0.4$). This multi-layer representation provides the foundation for costmap generation and subsequent path planning decisions.

\subsection{Costmap-Based Path Planning}

\begin{figure}[htbp]
    \centering
    \includegraphics[width=0.65\textwidth]{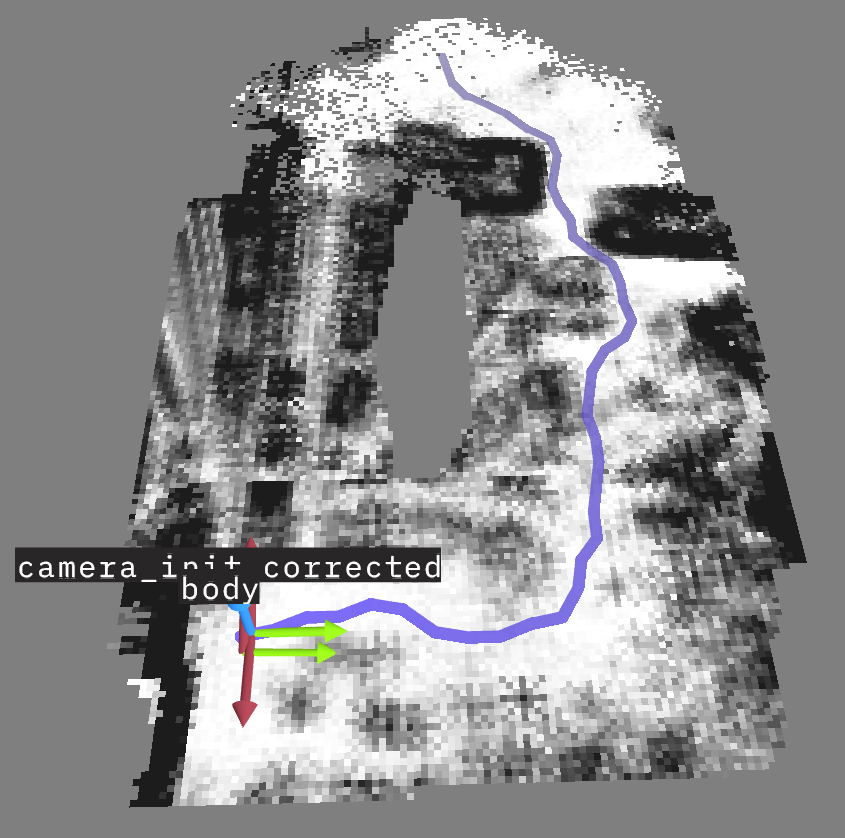}
    \caption{Illustrates a navigation costmap with A* planned path.}
    \label{fig:costmap_planned_path}
\end{figure}

Figure~\ref{fig:costmap_planned_path} presents the navigation costmap with the A* planned path during the test run. The costmap clearly delineates high-traversability regions (blue, cost $<$ 20) corresponding to open floor areas, intermediate-cost zones (yellow-orange, cost 20-60) near obstacles where the inflation radius creates safety margins, and high-cost or lethal regions (red-black, cost $>$ 80) marking obstacle footprints and constrained spaces.

The planned path (visualized in green) exhibits clear preference for high-traversability corridors, routing through open spaces and avoiding proximity to obstacles even when geometrically shorter paths exist through narrow gaps. This validates the hybrid cost function formulation (Equation 4.8), which penalizes both geometric distance and terrain difficulty. As the robot progresses and new regions enter the sensor field of view, the costmap updates with fresh traversability estimates at 15-18 Hz, triggering replanning events that refine the path based on newly available information.

\subsection{Region 1: Narrow Passage Navigation}

\begin{figure}[htbp]
    \centering
    \includegraphics[width=0.85\textwidth]{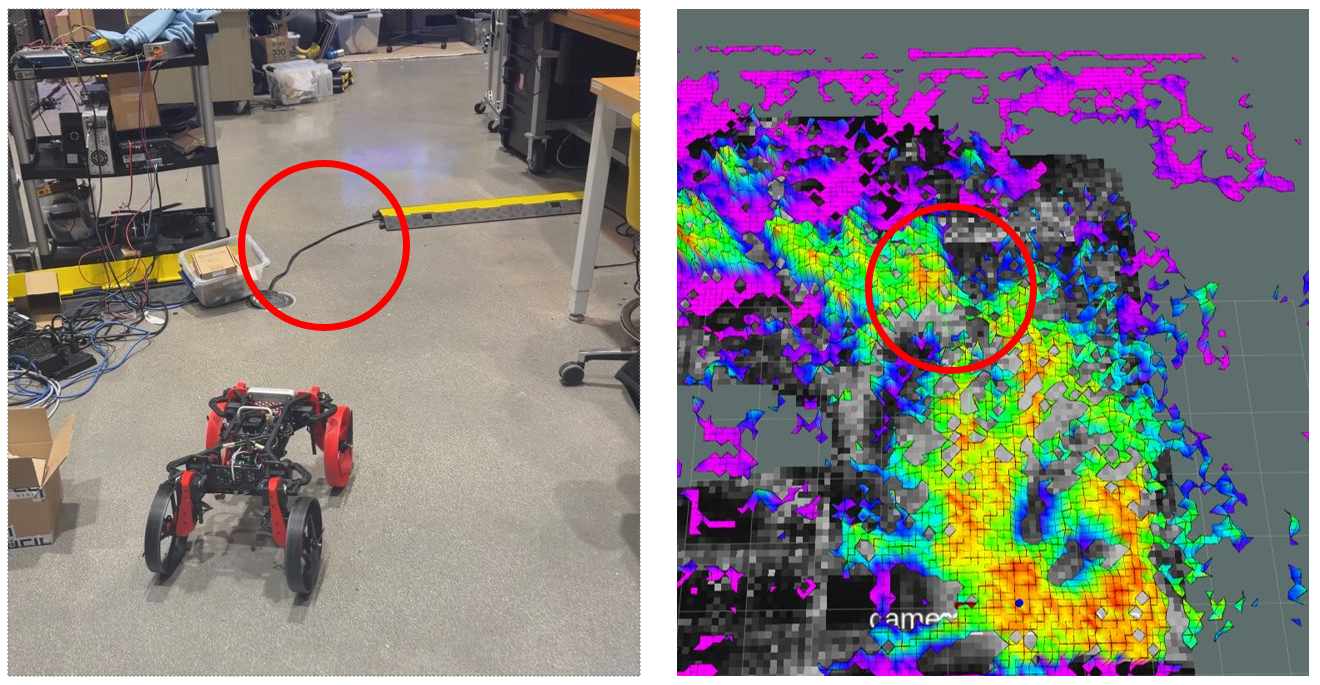}
    \caption{Illustrates Navigation through narrow passage.}
    \label{fig:narrow_passage}
\end{figure}

During the test run, the robot encountered a constrained passage between a floor-mounted wire management ramp and a cluster of boxes and tables. Figure~\ref{fig:narrow_passage} shows this region where the traversability map identified the narrow corridor as marginally acceptable (traversability approximately 0.65-0.75), with the wire ramp and obstacle edges marked as low-traversability boundaries. The A* planner selected this path as the optimal route to the goal, correctly assessing that the gap width exceeded the robot's footprint with sufficient clearance.

The robot successfully executed this maneuver with the DWA local controller maintaining safe clearance from both boundaries while tracking the global path, validating both the planner's risk assessment and the controller's ability to execute precise trajectories in confined spaces.

\subsection{Region 2: Traversability-Aware Obstacle Avoidance}

\begin{figure}[htbp]
    \centering
    \includegraphics[width=0.85\textwidth]{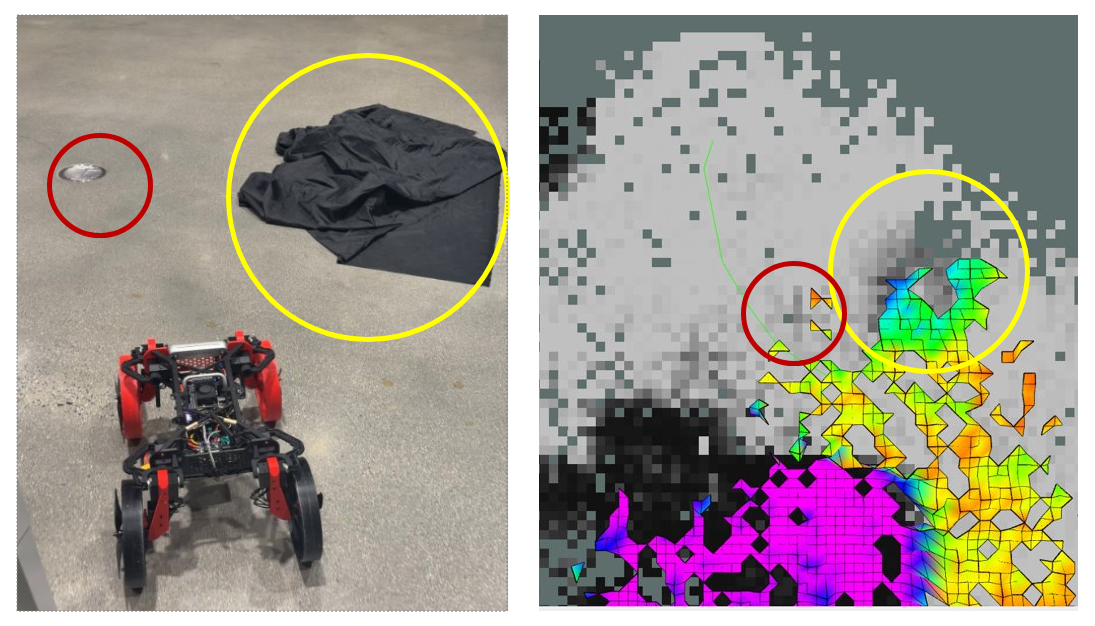}
    \caption{Illustrates M4 avoiding a textured and comparatively less traversable floor.}
    \label{fig:cloth_avoidance}
\end{figure}

In another region of the test environment, a black cloth placed on the floor provided an opportunity to evaluate the system's traversability estimation for novel terrain features. Figure~\ref{fig:cloth_avoidance} presents an annotated visualization showing the physical scene, traversability assessment, and resulting navigation decision.

The elevation mapping system detected the cloth as a subtle elevation discontinuity, and the CNN traversability estimator assigned low scores ($T < 0.4$) to the cloth region, appropriately classifying it as undesirable terrain. This classification triggered the costmap converter to mark the area with high cost ($C > 70$), and the continuous replanning loop generated an updated path routing around the obstacle. The robot smoothly avoided the cloth region, demonstrating the system's traversability-aware decision making and ability to adapt to environment variations through the integrated perception-planning pipeline.

Throughout the test run, the robot successfully navigated to the commanded goal position while avoiding both structured obstacles (tables, chairs, ramps) and unstructured terrain variations (cloth), validating the end-to-end functionality of the traversability-aware navigation system in a realistic, unstructured environment.
\subsection{Mode Transition Triggered by Path Planning Failure}
\label{subsec:mode-transition-results}

To validate the multimodal decision framework described in Section~\ref{subsubsec:cost-mode-selection}, we tested the system's response to scenarios where ground navigation becomes infeasible. In this experiment, the robot was commanded to navigate to a goal location blocked by an obstacle configuration that prevented any valid ground path.

The A* planner failed to find a traversable path on two consecutive planning cycles, triggering the failure condition $|\pi_{\text{ground}}| = 0$ (twice) from Equation~\ref{eq:failure_conditions}. Upon detecting this condition, the decision node recommended aerial mode transition, and the robot initiated the morphology transformation sequence.

\begin{figure}[htbp]
    \centering
    \includegraphics[width=\textwidth]{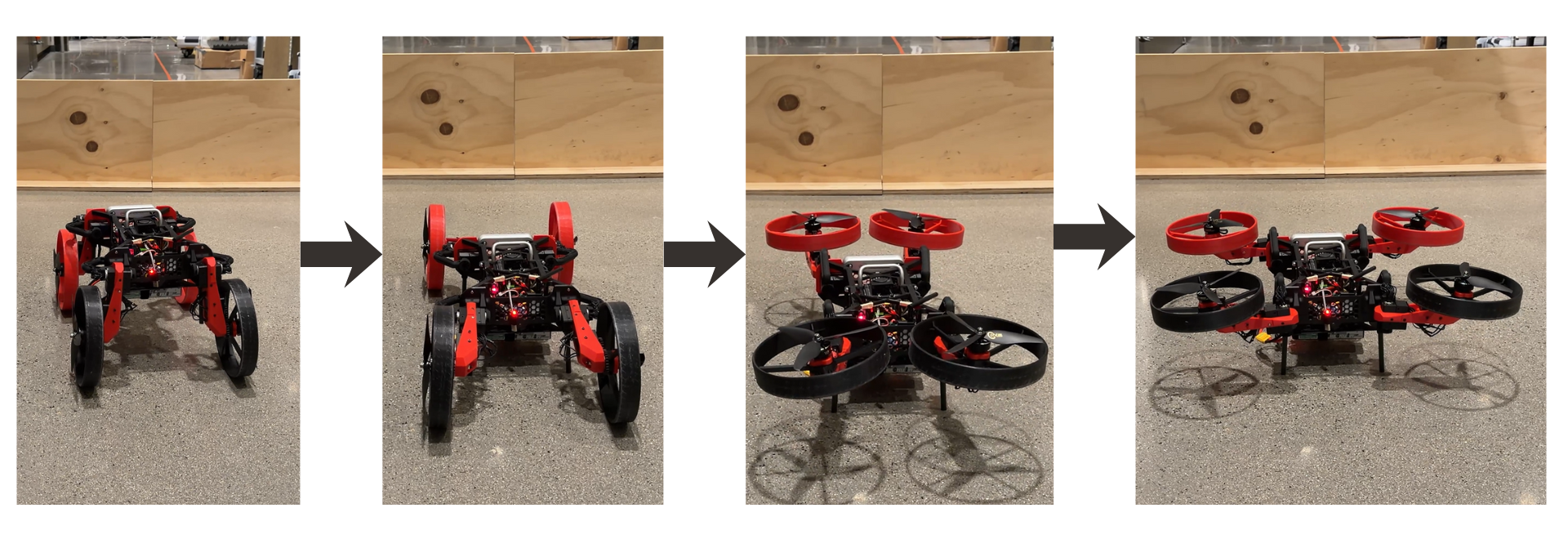}
    \caption{Illustrates Autonomous mode changing functionality.}
    \label{fig:mode_transition}
\end{figure}

Figure~\ref{fig:mode_transition} illustrates the complete ground-to-aerial transformation sequence. The robot begins in drive mode with wheels deployed for ground locomotion. Upon receiving the aerial mode recommendation, the appendages rotate upward, repositioning the wheels to serve as propeller guards. The transformation completes with the robot in quadcopter configuration, ready for aerial traversal to bypass the obstacle.

The transformation was triggered within 0.5 seconds of the second failed planning attempt, demonstrating responsive failure detection. This result validates that the decision framework correctly identifies impassable terrain scenarios and initiates appropriate mode switching to maintain mission continuity.

% conclusion and Future work
 % Chapter 6: Conclusion

\chapter{CONCLUSION}

\section{Summary}

This thesis developed a traversability-aware navigation system for the M4 multimodal ground-aerial robot that enables autonomous operation in complex, unstructured environments. The system combines FAST-LIO localization, GPU-accelerated elevation mapping with CNN-based traversability estimation, and an energy-aware A* planner to generate paths that balance distance efficiency with terrain quality.

The navigation framework uses a hybrid cost function incorporating geometric distance, terrain difficulty, and energy consumption. This allows the robot to make intelligent trade-offs between path length and traversability. Experimental validation showed the approach consistently chooses paths 8-11\% longer to achieve over 97\% reduction in terrain cost compared to traditional occupancy-based planning. Real robot tests in the cluttered Highbay Laboratory confirmed successful navigation through narrow passages and around both structured obstacles and unstructured terrain variations.

The system includes a decision framework that compares ground navigation energy costs against aerial bypass options, accounting for both locomotion power and morphology transformation costs. This foundation supports future deployment of fully autonomous multimodal navigation.

\section{Challenges and Limitations}

The CNN traversability model was trained on generic terrain data rather than M4-specific characteristics. While it generalizes adequately to tested environments, training with platform-specific data would improve edge case performance. The path planner operates in 2D, optimizing only ground trajectories without considering 3D aerial paths. This requires separate comparison between ground and aerial options rather than unified multimodal planning. Testing was limited to indoor laboratory environments with flat surfaces and discrete obstacles. Performance on outdoor terrain with slopes, vegetation, and varying conditions remains unvalidated.

\section{Future Work}

Future development should focus on three key areas. First, collecting an M4-specific traversability dataset that captures wheel slippage, traction characteristics, and kinematic constraints during actual terrain traversal. Retraining the CNN with ground truth labels reflecting the M4's physical capabilities, including tip-over risk, ground clearance limitations, and energy expenditure patterns, would improve prediction accuracy. Second, extending the planner to 3D for aerial mode trajectory optimization with dynamic constraints and obstacle avoidance would enable unified multimodal path planning. Third, completing hardware integration of aerial capabilities including servo control for morphology transformation and quadcopter motor control would validate the energy cost models and enable end-to-end multimodal autonomous operation.

% --- Bibliography ----
\bibliographystyle{IEEEtran}  %'plain' for standard, 'unsrt' for correct order

% include bibliography definition
\bibliography{bib/thesis}

@article{sihite2023multi,
  title={Multi-Modal Mobility Morphobot (M4) with appendage repurposing for locomotion plasticity enhancement},
  author={Sihite, Eric and Kalantari, Arash and Nemovi, Reza and Ramezani, Alireza and Gharib, Morteza},
  journal={Nature communications},
  volume={14},
  number={1},
  pages={3323},
  year={2023},
  publisher={Nature Publishing Group UK London}
}

@inproceedings{sihite2023demonstrating,
  title={Demonstrating autonomous 3d path planning on a novel scalable ugv-uav morphing robot},
  author={Sihite, Eric and Slezak, Filip and Mandralis, Ioannis and Salagame, Adarsh and Ramezani, Milad and Kalantari, Arash and Ramezani, Alireza and Gharib, Morteza},
  booktitle={2023 IEEE/RSJ International Conference on Intelligent Robots and Systems (IROS)},
  pages={3064--3069},
  year={2023},
  organization={IEEE}
}

@article{gherold2025self,
  title={Self-supervised cost of transport estimation for multimodal path planning},
  author={Gherold, Vincent and Mandralis, Ioannis and Sihite, Eric and Salagame, Adarsh and Ramezani, Alireza and Gharib, Morteza},
  journal={IEEE Robotics and Automation Letters},
  year={2025},
  publisher={IEEE}
}

@article{mandralis2025atmo,
  title={ATMO: an aerially transforming morphobot for dynamic ground-aerial transition},
  author={Mandralis, Ioannis and Nemovi, Reza and Ramezani, Alireza and Murray, Richard M and Gharib, Morteza},
  journal={Communications Engineering},
  volume={4},
  number={1},
  pages={74},
  year={2025},
  publisher={Nature Publishing Group UK London}
}

@inproceedings{sihite2024dynamic,
  title={Dynamic modeling of wing-assisted inclined running with a morphing multi-modal robot},
  author={Sihite, Eric and Ramezani, Alireza and Gharib, Morteza},
  booktitle={2024 IEEE International Conference on Robotics and Automation (ICRA)},
  pages={2339--2345},
  year={2024},
  organization={IEEE}
}

@article{salagame2025crater,
  title={Crater Observing Bioinspired Rolling Articulator (COBRA)},
  author={Salagame, Adarsh and Noyes, Henry and Sihite, Eric and Kalantari, Arash and Ramezani, Alireza},
  journal={Advanced Intelligent Systems},
  pages={2500352},
  year={2025},
  publisher={Wiley Online Library}
}

@article{wang2025thruster,
  title={Thruster-Enhanced Locomotion: A Decoupled Model Predictive Control with Learned Contact Residuals},
  author={Wang, Chenghao and Ramezani, Alireza},
  journal={arXiv preprint arXiv:2508.03003},
  year={2025}
}

@article{wang2025dynamic,
  title={Dynamic Quadrupedal Legged and Aerial Locomotion via Structure Repurposing},
  author={Wang, Chenghao and Krishnamurthy, Kaushik Venkatesh and Pitroda, Shreyansh and Salagame, Adarsh and Mandralis, Ioannis and Sihite, Eric and Ramezani, Alireza and Gharib, Morteza},
  journal={arXiv preprint arXiv:2510.09526},
  year={2025}
}

@article{salagame2025optimal,
  title={Optimal Trajectory Planning in a Vertically Undulating Snake Locomotion using Contact-implicit Optimization},
  author={Salagame, Adarsh and Sihite, Eric and Ramezani, Alireza},
  journal={arXiv preprint arXiv:2508.02953},
  year={2025}
}

@article{salagame2025vision,
  title={Vision-Guided Loco-Manipulation with a Snake Robot},
  author={Salagame, Adarsh and Potluri, Sasank and Vaidyanathan, Keshav Bharadwaj and Gangaraju, Kruthika and Sihite, Eric and Ramezani, Milad and Ramezani, Alireza},
  journal={arXiv preprint arXiv:2503.18308},
  year={2025}
}

@inproceedings{krishnamurthy2025thruster,
  title={Thruster-Assisted Incline Walking of a Legged-Aerial Robot Using Reduced Order Model and Collocation Method},
  author={Krishnamurthy, Kaushik Venkatesh and Wang, Chenghao and Pitroda, Shreyansh and Salagame, Adarsh and Sihite, Eric and Nemovi, Reza and Ramezani, Alireza and Gharib, Morteza},
  booktitle={2025 American Control Conference (ACC)},
  pages={3738--3743},
  year={2025},
  organization={IEEE}
}

@inproceedings{sihite2024posture,
  title={Posture manipulation of thruster-enhanced bipedal robot performing dynamic wall-jumping using model predictive control},
  author={Sihite, Eric and Pitroda, Shreyansh and Liu, Taoran and Wang, Chenghao and Krishnamurthy, Kaushik Venkatesh and Salagame, Adarsh and Nemovi, Reza and Ramezani, Alireza and Gharib, Morteza},
  booktitle={2024 IEEE-RAS 23rd International Conference on Humanoid Robots (Humanoids)},
  pages={491--496},
  year={2024},
  organization={IEEE}
}

@article{sihite2025morphology,
  title={A morphology-centered view towards describing bats dynamically versatile wing conformations},
  author={Sihite, Eric and Ramezani, Alireza},
  journal={The International Journal of Robotics Research},
  volume={44},
  number={3},
  pages={431--464},
  year={2025},
  publisher={SAGE Publications Sage UK: London, England}
}

@inproceedings{jiang2024hierarchical,
  title={Hierarchical RL-guided large-scale navigation of a snake robot},
  author={Jiang, Shuo and Salagame, Adarsh and Ramezani, Alireza and Wong, Lawson LS},
  booktitle={2024 IEEE International Conference on Advanced Intelligent Mechatronics (AIM)},
  pages={1347--1352},
  year={2024},
  organization={IEEE}
}

@inproceedings{salagame2024non,
  title={Non-impulsive contact-implicit motion planning for morpho-functional loco-manipulation},
  author={Salagame, Adarsh and Gangaraju, Kruthika and Nallaguntla, Harin Kumar and Gupta, Bibek and Sihite, Eric and Schirner, Gunar and Ramezani, Alireza},
  booktitle={2024 IEEE International Conference on Advanced Intelligent Mechatronics (AIM)},
  pages={309--314},
  year={2024},
  organization={IEEE}
}

@inproceedings{gupta2024banking,
  title={Banking turn of high-dof dynamic morphing wing flight by shifting structure response using optimization},
  author={Gupta, Bibek and Shah, Yogi and Liu, Taoran and Sihite, Eric and Ramezani, Alireza},
  booktitle={2024 IEEE International Conference on Advanced Intelligent Mechatronics (AIM)},
  pages={94--99},
  year={2024},
  organization={IEEE}
}

@inproceedings{gupta2024bounding,
  title={Bounding flight control of dynamic morphing wings},
  author={Gupta, Bibek and Dhole, Aniket and Salagame, Adarsh and Niu, Xuejian and Xu, Yizhe and Venkatesh, Kaushik and Ghanem, Paul and Mandralis, Ioannis and Sihite, Eric and Ramezani, Alireza},
  booktitle={2024 IEEE International Conference on Advanced Intelligent Mechatronics (AIM)},
  pages={100--105},
  year={2024},
  organization={IEEE}
}

@inproceedings{salagame2024dynamic,
  title={Dynamic posture manipulation during tumbling for closed-loop heading angle control},
  author={Salagame, Adarsh and Sihite, Eric and Schirner, Gunar and Ramezani, Alireza},
  booktitle={2024 IEEE International Conference on Advanced Intelligent Mechatronics (AIM)},
  pages={64--69},
  year={2024},
  organization={IEEE}
}

@inproceedings{jiang2024snake,
  title={Snake robot with tactile perception navigates on large-scale challenging terrain},
  author={Jiang, Shuo and Salagame, Adarsh and Ramezani, Alireza and Wong, Lawson LS},
  booktitle={2024 IEEE International Conference on Robotics and Automation (ICRA)},
  pages={5090--5096},
  year={2024},
  organization={IEEE}
}

@inproceedings{sihite2023actuation,
  title={Actuation and flight control of high-dof dynamic morphing wing flight by shifting structure response},
  author={Sihite, Eric and Salagame, Adarsh and Ghanem, Paul and Ramezani, Alireza},
  booktitle={2023 62nd IEEE Conference on Decision and Control (CDC)},
  pages={8824--8829},
  year={2023},
  organization={IEEE}
}

@article{ghanem2021efficient,
  title={Efficient modeling of morphing wing flight using neural networks and cubature rules},
  author={Ghanem, Paul and Bicer, Yunus and Erdogmus, Deniz and Ramezani, Alireza},
  journal={arXiv preprint arXiv:2110.01057},
  year={2021}
}

@inproceedings{dhole2023hovering,
  title={Hovering control of flapping wings in tandem with multi-rotors},
  author={Dhole, Aniket and Gupta, Bibek and Salagame, Adarsh and Niu, Xuejian and Xu, Yizhe and Venkatesh, Kaushik and Ghanem, Paul and Mandralis, Ioannis and Sihite, Eric and Ramezani, Alireza},
  booktitle={2023 IEEE/RSJ International Conference on Intelligent Robots and Systems (IROS)},
  pages={6639--6644},
  year={2023},
  organization={IEEE}
}

@inproceedings{sihite2022efficient,
  title={Efficient path planning and tracking for multi-modal legged-aerial locomotion using integrated probabilistic road maps (prm) and reference governors (rg)},
  author={Sihite, Eric and Mottis, Benjamin and Ghanem, Paul and Ramezani, Alireza and Gharib, Morteza},
  booktitle={2022 IEEE 61st Conference on Decision and Control (CDC)},
  pages={764--770},
  year={2022},
  organization={IEEE}
}

@article{ramezani2022thruster,
  title={Thruster-assisted legged mobility for explorations on mars},
  author={Ramezani, A and Sreenath, K},
  journal={Low-Cost Science Mission Concepts for Mars Exploration},
  volume={2655},
  pages={5074},
  year={2022}
}

@article{ramezani2022efficient,
  title={Efficient and endured aerial mobility on mars using novel morphing micro aerial vehicle designs},
  author={Ramezani, A and Sihite, E and Devey, S and Gharib, M},
  journal={Low-Cost Science Mission Concepts for Mars Exploration},
  volume={2655},
  pages={5051},
  year={2022}
}

@inproceedings{ramezani2021generative,
  title={Generative design of nu's husky carbon, a morpho-functional, legged robot},
  author={Ramezani, Alireza and Dangol, Pravin and Sihite, Eric and Lessieur, Andrew and Kelly, Peter},
  booktitle={2021 IEEE International Conference on Robotics and Automation (ICRA)},
  pages={4040--4046},
  year={2021},
  organization={IEEE}
}

@inproceedings{liang2021rough,
  title={Rough-terrain locomotion and unilateral contact force regulations with a multi-modal legged robot},
  author={Liang, Kaier and Sihite, Eric and Dangol, Pravin and Lessieur, Andrew and Ramezani, Alireza},
  booktitle={2021 American Control Conference (ACC)},
  pages={1762--1769},
  year={2021},
  organization={IEEE}
}

@inproceedings{sihite2021integrated,
  title={An integrated mechanical intelligence and control approach towards flight control of aerobat},
  author={Sihite, Eric and Darabi, Atefe and Dangol, Pravin and Lessieur, Andrew and Ramezani, Alireza},
  booktitle={2021 American Control Conference (ACC)},
  pages={84--91},
  year={2021},
  organization={IEEE}
}

@inproceedings{sihite2020enforcing,
  title={Enforcing nonholonomic constraints in Aerobat, a roosting flapping wing model},
  author={Sihite, Eric and Ramezani, Alireza},
  booktitle={2020 59th IEEE Conference on Decision and Control (CDC)},
  pages={5321--5327},
  year={2020},
  organization={IEEE}
}

@inproceedings{hoff2019trajectory,
  title={Trajectory planning for a bat-like flapping wing robot},
  author={Hoff, Jonathan and Syed, Usman and Ramezani, Alireza and Hutchinson, Seth},
  booktitle={2019 IEEE/RSJ International Conference on Intelligent Robots and Systems (IROS)},
  pages={6800--6805},
  year={2019},
  organization={IEEE}
}

@article{hoff2018optimizing,
  author = {Jonathan Hoff and Alireza Ramezani and Soon-Jo Chung and Seth Hutchinson},
  title = {Optimizing the structure and movement of a robotic bat with biological kinematic synergies},
  journal = {The International Journal of Robotics Research},
  volume = {37},
  number = {10},
  pages = {1233-1252},
  year = {2018},
  doi = {10.1177/0278364918804654}
}

@inproceedings{sihite2021unilateral,
  title={Unilateral ground contact force regulations in thruster-assisted legged locomotion},
  author={Sihite, Eric and Dangol, Pravin and Ramezani, Alireza},
  booktitle={2021 IEEE/ASME International Conference on Advanced Intelligent Mechatronics (AIM)},
  pages={389--395},
  year={2021},
  organization={IEEE}
}

@article{ramezani2017biomimetic,
  title={A biomimetic robotic platform to study flight specializations of bats},
  author={Ramezani, Alireza and Chung, Soon-Jo and Hutchinson, Seth},
  journal={Science Robotics},
  volume={2},
  number={3},
  pages={eaal2505},
  year={2017},
  publisher={American Association for the Advancement of Science}
}

@inproceedings{salagame2024quadrupedal,
  title={Quadrupedal locomotion control on inclined surfaces using collocation method},
  author={Salagame, Adarsh and Gianello, Maria and Wang, Chenghao and Venkatesh, Kaushik and Pitroda, Shreyansh and Rajput, Rohit and Sihite, Eric and Leeser, Miriam and Ramezani, Alireza},
  booktitle={2024 American Control Conference (ACC)},
  pages={2838--2843},
  year={2024},
  organization={IEEE}
}

@inproceedings{krishnamurthy2024narrow,
  title={Narrow-path, dynamic walking using integrated posture manipulation and thrust vectoring},
  author={Krishnamurthy, Kaushik Venkatesh and Wang, Chenghao and Pitroda, Shreyansh and Salagame, Adarsh and Sihite, Eric and Nemovi, Reza and Ramezani, Alireza and Gharib, Morteza},
  booktitle={2024 IEEE International Conference on Advanced Intelligent Mechatronics (AIM)},
  pages={898--903},
  year={2024},
  organization={IEEE}
}

@inproceedings{salagame2024heading,
  title={Heading control for obstacle avoidance using dynamic posture manipulation during tumbling locomotion},
  author={Salagame, Adarsh and Gangaraju, Kruthika and Sihite, Eric and Schirner, Gunar and Ramezani, Alireza},
  booktitle={2024 IEEE/RSJ International Conference on Intelligent Robots and Systems (IROS)},
  pages={13555--13560},
  year={2024},
  organization={IEEE}
}

@inproceedings{salagame2024loco,
  title={Loco-manipulation with nonimpulsive contact-implicit planning in a slithering robot},
  author={Salagame, Adarsh and Gangaraju, Kruthika and Nallaguntla, Harin Kumar and Sihite, Eric and Schirner, Gunar and Ramezani, Alireza},
  booktitle={2024 IEEE/RSJ International Conference on Intelligent Robots and Systems (IROS)},
  pages={309--314},
  year={2024},
  organization={IEEE}
}

@article{salagame2025nmpc,
  title={NMPC-based Unified Posture Manipulation and Thrust Vectoring for Fault Recovery},
  author={Salagame, Adarsh and Pandya, Shashwat and Sihite, Eric and Ramezani, Alireza and Gharib, Morteza},
  journal={IEEE Control Systems Letters},
  year={2025},
  publisher={IEEE}
}

@article{salagame2025reduced,
  title={Reduced-Order Model-Based Gait Generation for Snake Robot Locomotion using NMPC},
  author={Salagame, Adarsh and Sihite, Eric and Ramezani, Milad and Ramezani, Alireza},
  journal={arXiv preprint arXiv:2503.06402},
  year={2025}
}

@inproceedings{pitroda2024capture,
  title={Capture point control in thruster-assisted bipedal locomotion},
  author={Pitroda, Shreyansh and Bondada, Aditya and Venkatesh, Kaushik and Salagame, Adarsh and Wang, Chenghao and Liu, Taoran and Gupta, Bibek and Sihite, Eric and Nemovi, Reza and Ramezani, Alireza and others},
  booktitle={2024 IEEE International Conference on Advanced Intelligent Mechatronics (AIM)},
  pages={1139--1144},
  year={2024},
  organization={IEEE}
}

@article{kim2021bipedal,
  title={A bipedal walking robot that can fly, slackline, and skateboard},
  author={Kim, Kyunam and Spieler, Patrick and Lupu, Elena-Sorina and Ramezani, Alireza and Chung, Soon-Jo},
  journal={Science Robotics},
  volume={6},
  number={59},
  pages={eabf8136},
  year={2021},
  publisher={American Association for the Advancement of Science}
}

@article{shi2024mtabot,
  title={MTABot: An efficient morphable terrestrial-aerial robot with two transformable wheels},
  author={Shi, Ke and Jiang, Zainan and Ma, Liyan and Qi, Le and Jin, Minghe},
  journal={IEEE Robotics and Automation Letters},
  volume={9},
  number={2},
  pages={1875--1882},
  year={2024},
  publisher={IEEE}
}

@ARTICLE{kalantari2014hytaq,
  author={Kalantari, Arash and Spenko, Matthew},
  journal={IEEE Transactions on Robotics}, 
  title={Modeling and Performance Assessment of the HyTAQ, a Hybrid Terrestrial/Aerial Quadrotor}, 
  year={2014},
  volume={30},
  number={5},
  pages={1278-1285},
  doi={10.1109/TRO.2014.2337555}
}

@article{farley2020mars,
  title={Mars 2020 mission overview},
  author={Farley, Kenneth A and Williford, Kenneth H and Stack, Kathryn M and Bhartia, Rohit and Chen, Al and de la Torre, Manuel and Hand, Kevin and Goreva, Yulia and Herd, Christopher DK and Hueso, Ricardo and others},
  journal={Space Science Reviews},
  volume={216},
  number={8},
  pages={142},
  year={2020},
  publisher={Springer}
}

@inproceedings{tzanetos2022ingenuity,
  title={Ingenuity mars helicopter: From technology demonstration to extraterrestrial scout},
  author={Tzanetos, Theodore and Aung, MiMi and Balaram, J and Grip, Havard Fjrer and Karras, Jaakko T and Canham, Timothy K and Kubiak, Gerik and Anderson, Joshua and Merewether, Gene and Starch, Michael and others},
  booktitle={2022 IEEE Aerospace Conference (AERO)},
  pages={01--19},
  year={2022},
  organization={IEEE}
}

@INPROCEEDINGS{shan2020liosam,
  author={Shan, Tixiao and Englot, Brendan and Meyers, Drew and Wang, Wei and Ratti, Carlo and Rus, Daniela},
  booktitle={2020 IEEE/RSJ International Conference on Intelligent Robots and Systems (IROS)}, 
  title={LIO-SAM: Tightly-coupled Lidar Inertial Odometry via Smoothing and Mapping}, 
  year={2020},
  pages={5135-5142},
  doi={10.1109/IROS45743.2020.9341176}
}

@article{labbe2019rtabmap,
  title={RTAB-Map as an open-source lidar and visual simultaneous localization and mapping library for large-scale and long-term online operation},
  author={Labb{\'e}, Mathieu and Michaud, Fran{\c{c}}ois},
  journal={Journal of Field Robotics},
  volume={36},
  number={2},
  pages={416--446},
  year={2019},
  publisher={Wiley}
}

@ARTICLE{xu2021fastlio,
  author={Xu, Wei and Zhang, Fu},
  journal={IEEE Robotics and Automation Letters}, 
  title={FAST-LIO: A Fast, Robust LiDAR-Inertial Odometry Package by Tightly-Coupled Iterated Kalman Filter}, 
  year={2021},
  volume={6},
  number={2},
  pages={3317-3324},
  doi={10.1109/LRA.2021.3064227}
}

@article{xu2022fastlio2,
  title={Fast-lio2: Fast direct lidar-inertial odometry},
  author={Xu, Wei and Cai, Yixi and He, Dongjiao and Lin, Jiarong and Zhang, Fu},
  journal={IEEE Transactions on Robotics},
  volume={38},
  number={4},
  pages={2053--2073},
  year={2022},
  publisher={IEEE}
}

@inproceedings{zhang2014loam,
  title={LOAM: Lidar odometry and mapping in real-time},
  author={Zhang, Ji and Singh, Sanjiv},
  booktitle={Robotics: Science and Systems},
  volume={2},
  number={9},
  pages={1--9},
  year={2014},
  organization={Berkeley, CA}
}

@inproceedings{shan2018lego,
  title={Lego-loam: Lightweight and ground-optimized lidar odometry and mapping on variable terrain},
  author={Shan, Tixiao and Englot, Brendan},
  booktitle={2018 IEEE/RSJ International Conference on Intelligent Robots and Systems (IROS)},
  pages={4758--4765},
  year={2018},
  organization={IEEE}
}

@article{campos2021orb,
  title={Orb-slam3: An accurate open-source library for visual, visual--inertial, and multimap slam},
  author={Campos, Carlos and Elvira, Richard and Rodr{\'\i}guez, Juan J G{\'o}mez and Montiel, Jos{\'e} MM and Tard{\'o}s, Juan D},
  journal={IEEE Transactions on Robotics},
  volume={37},
  number={6},
  pages={1874--1890},
  year={2021},
  publisher={IEEE}
}

@ARTICLE{cadena2016slam,
  author={Cadena, Cesar and Carlone, Luca and Carrillo, Henry and Latif, Yasir and Scaramuzza, Davide and Neira, Jos{\'e} and Reid, Ian and Leonard, John J.},
  journal={IEEE Transactions on Robotics}, 
  title={Past, Present, and Future of Simultaneous Localization and Mapping: Toward the Robust-Perception Age}, 
  year={2016},
  volume={32},
  number={6},
  pages={1309-1332},
  doi={10.1109/TRO.2016.2624754}
}

@article{hornung2013octomap,
  author = {Hornung, Armin and Wurm, Kai M. and Bennewitz, Maren and Stachniss, Cyrill and Burgard, Wolfram},
  title = {OctoMap: an efficient probabilistic 3D mapping framework based on octrees},
  journal = {Autonomous Robots},
  volume = {34},
  number = {3},
  pages = {189--206},
  year = {2013},
  doi = {10.1007/s10514-012-9321-0}
}

@INPROCEEDINGS{miki2022elevation,
  author={Miki, Takahiro and Wellhausen, Lorenz and Grandia, Ruben and Jenelten, Fabian and Homberger, Timon and Hutter, Marco},
  booktitle={2022 IEEE/RSJ International Conference on Intelligent Robots and Systems (IROS)}, 
  title={Elevation Mapping for Locomotion and Navigation using GPU}, 
  year={2022},
  pages={2273-2280},
  doi={10.1109/IROS47612.2022.9981507}
}

@inbook{fankhauser2014elevation,
  author = {Fankhauser, P. and Bloesch, M. and Gehring, C. and Hutter, M. and Siegwart, R.},
  title = {Robot-Centric Elevation Mapping with Uncertainty Estimates},
  booktitle = {Mobile Service Robotics},
  pages = {433-440},
  year = {2014},
  doi = {10.1142/9789814623353_0051}
}

@inbook{fankhauser2016grid,
  author = {Fankhauser, P{\'e}ter and Hutter, Marco},
  year = {2016},
  title = {A Universal Grid Map Library: Implementation and Use Case for Rough Terrain Navigation},
  volume = {625},
  journal = {Robot Operating System (ROS)},
  doi = {10.1007/978-3-319-26054-9_5}
}

@article{papadakis2013traversability,
  title = {Terrain traversability analysis methods for unmanned ground vehicles: A survey},
  journal = {Engineering Applications of Artificial Intelligence},
  volume = {26},
  number = {4},
  pages = {1373-1385},
  year = {2013},
  doi = {10.1016/j.engappai.2013.01.006},
  author = {Panagiotis Papadakis}
}

@INPROCEEDINGS{gasparino2024wayfaster,
  author={Gasparino, Mateus V. and Sivakumar, Arun N. and Chowdhary, Girish},
  booktitle={2024 IEEE International Conference on Robotics and Automation (ICRA)}, 
  title={WayFASTER: a Self-Supervised Traversability Prediction for Increased Navigation Awareness}, 
  year={2024},
  pages={8486-8492},
  doi={10.1109/ICRA57147.2024.10610436}
}

@ARTICLE{gasparino2022wayfast,
  author={Gasparino, Mateus V. and Sivakumar, Arun N. and Liu, Yixiao and Velasquez, Andres E. B. and Higuti, Vitor A. H. and Rogers, John and Tran, Huy and Chowdhary, Girish},
  journal={IEEE Robotics and Automation Letters}, 
  title={WayFAST: Navigation With Predictive Traversability in the Field}, 
  year={2022},
  volume={7},
  number={4},
  pages={10651-10658},
  doi={10.1109/LRA.2022.3193464}
}

@INPROCEEDINGS{seraji2001traversability,
  author={Seraji, H.},
  booktitle={10th IEEE International Conference on Fuzzy Systems}, 
  title={Terrain-based robot navigation using multi-scale traversability indices}, 
  year={2001},
  volume={3},
  pages={1309-1314},
  doi={10.1109/FUZZ.2001.1008899}
}

@article{howard2001vision,
  title={Vision-based terrain characterization and traversability assessment},
  author={Howard, Ayanna and Seraji, Homayoun},
  journal={Journal of Robotic Systems},
  volume={18},
  number={10},
  pages={577--587},
  year={2001},
  publisher={Wiley Online Library}
}

@article{waibel2022rough,
  title={How rough is the path? Terrain traversability estimation for local and global path planning},
  author={Waibel, Gabriel G{\"u}nter and L{\"o}w, Tobias and Nass, Mathieu and Howard, David and Bandyopadhyay, Tirthankar and Borges, Paulo Vinicius Koerich},
  journal={IEEE Transactions on Intelligent Transportation Systems},
  volume={23},
  number={9},
  pages={16462--16473},
  year={2022},
  publisher={IEEE}
}

@article{valada2017deep,
  title={Deep spatiotemporal models for robust proprioceptive terrain classification},
  author={Valada, Abhinav and Burgard, Wolfram},
  journal={The International Journal of Robotics Research},
  volume={36},
  number={13-14},
  pages={1521--1539},
  year={2017},
  publisher={SAGE Publications}
}

@article{zurn2020self,
  title={Self-supervised visual terrain classification from unsupervised acoustic feature learning},
  author={Z{\"u}rn, Jannik and Burgard, Wolfram and Valada, Abhinav},
  journal={IEEE Transactions on Robotics},
  volume={37},
  number={2},
  pages={466--481},
  year={2020},
  publisher={IEEE}
}

@article{chavez2018learning,
  title={Learning ground traversability from simulations},
  author={Chavez-Garcia, R Omar and Guzzi, J{\'e}r{\^o}me and Gambardella, Luca M and Giusti, Alessandro},
  journal={IEEE Robotics and Automation Letters},
  volume={3},
  number={3},
  pages={1695--1702},
  year={2018},
  publisher={IEEE}
}

@article{frey2023fast,
  title={Fast traversability estimation for wild visual navigation},
  author={Frey, Jonas and Mattamala, Matias and Chebrolu, Nived and Cadena, Cesar and Fallon, Maurice and Hutter, Marco},
  journal={arXiv preprint arXiv:2305.08510},
  year={2023}
}

@article{vecchio2024terrain,
  title={Terrain traversability prediction through self-supervised learning and unsupervised domain adaptation on synthetic data},
  author={Vecchio, Giuseppe and Palazzo, Simone and Guastella, Dario C and Giordano, Daniela and Muscato, Giovanni and Spampinato, Concetto},
  journal={Autonomous Robots},
  volume={48},
  number={2},
  pages={4},
  year={2024},
  publisher={Springer}
}

@article{choi2023learning,
  title={Learning quadrupedal locomotion on deformable terrain},
  author={Choi, Suyoung and Ji, Gwanghyeon and Park, Jeongsoo and Kim, Hyeongjun and Mun, Juhyeok and Lee, Jeong Hyun and Hwangbo, Jemin},
  journal={Science Robotics},
  volume={8},
  number={74},
  pages={eade2256},
  year={2023},
  publisher={American Association for the Advancement of Science}
}

@article{kim2025high,
  title={High-speed control and navigation for quadrupedal robots on complex and discrete terrain},
  author={Kim, Hyeongjun and Oh, Hyunsik and Park, Jeongsoo and Kim, Yunho and Youm, Donghoon and Jung, Moonkyu and Lee, Minho and Hwangbo, Jemin},
  journal={Science Robotics},
  volume={10},
  number={102},
  pages={eads6192},
  year={2025},
  publisher={American Association for the Advancement of Science}
}

@article{lee2024learning,
  title={Learning robust autonomous navigation and locomotion for wheeled-legged robots},
  author={Lee, Joonho and Bjelonic, Marko and Reske, Alexander and Wellhausen, Lorenz and Miki, Takahiro and Hutter, Marco},
  journal={Science Robotics},
  volume={9},
  number={89},
  pages={eadi9641},
  year={2024},
  publisher={American Association for the Advancement of Science}
}

@ARTICLE{hart1968astar,
  author={Hart, Peter E. and Nilsson, Nils J. and Raphael, Bertram},
  journal={IEEE Transactions on Systems Science and Cybernetics}, 
  title={A Formal Basis for the Heuristic Determination of Minimum Cost Paths}, 
  year={1968},
  volume={4},
  number={2},
  pages={100-107},
  doi={10.1109/TSSC.1968.300136}
}

@ARTICLE{fox1997dwa,
  author={Fox, D. and Burgard, W. and Thrun, S.},
  journal={IEEE Robotics \& Automation Magazine}, 
  title={The dynamic window approach to collision avoidance}, 
  year={1997},
  volume={4},
  number={1},
  pages={23-33},
  doi={10.1109/100.580977}
}

@article{kavraki1998prm,
  title={Analysis of probabilistic roadmaps for path planning},
  author={Kavraki, Lydia E and Kolountzakis, Mihail N and Latombe, J-C},
  journal={IEEE Transactions on Robotics and Automation},
  volume={14},
  number={1},
  pages={166--171},
  year={1998},
  publisher={IEEE}
}

@article{shu2012dijkstra,
  title={The improved Dijkstra's shortest path algorithm and its application},
  author={Shu-Xi, Wang},
  journal={Procedia Engineering},
  volume={29},
  pages={1186--1190},
  year={2012},
  publisher={Elsevier}
}

@article{tranzatto2022cerberus,
  author = {Tranzatto, Marco and Dharmadhikari, Mihir and Bernreiter, Lukas and others},
  year = {2022},
  title = {Team CERBERUS Wins the DARPA Subterranean Challenge: Technical Overview and Lessons Learned},
  doi = {10.48550/arXiv.2207.04914}
}

@article{daler2015bioinspired,
  title={A bioinspired multi-modal flying and walking robot},
  author={Daler, Ludovic and Mintchev, Stefano and Stefanini, Cesare and Floreano, Dario},
  journal={Bioinspiration \& biomimetics},
  volume={10},
  number={1},
  pages={016005},
  year={2015},
  publisher={IOP Publishing}
}

@article{shin2024fast,
  title={Fast ground-to-air transition with avian-inspired multifunctional legs},
  author={Shin, Won Dong and Phan, Hoang-Vu and Daley, Monica A and Ijspeert, Auke J and Floreano, Dario},
  journal={Nature},
  volume={636},
  number={8041},
  pages={86--91},
  year={2024},
  publisher={Nature Publishing Group UK London}
}

@article{tang2025duawlfin,
  title={Duawlfin: A drone with unified actuation for wheeled locomotion and flight operation},
  author={Tang, Jerry and Zhang, Ruiqi and Beyduz, Kaan and Jiang, Yiwei and Wiebe, Cody and Zhang, Haoyu and Asoro, Osaruese and Mueller, Mark W},
  journal={arXiv preprint arXiv:2505.13836},
  year={2025}
}

@inproceedings{sabet2019rollocopter,
  title={Rollocopter: An energy-aware hybrid aerial-ground mobility for extreme terrains},
  author={Sabet, Sahand and Agha-Mohammadi, Ali-Akbar and Tagliabue, Andrea and Elliott, D Sawyer and Nikravesh, Parviz E},
  booktitle={2019 IEEE Aerospace Conference},
  pages={1--8},
  year={2019},
  organization={IEEE}
}

@mastersthesis{rajput2023m4,
  title={Towards Autonomous Multi-Modal Mobility Morphobot (M4) Robot: Traversability Estimation and 3D Path Planning},
  author={Rajput, Rohit Hiraman},
  year={2023},
  school={Northeastern University},
  address={Boston, Massachusetts},
  note={Order No. 30637204, ProQuest}
}

@misc{optitrack2023,
  author = {{NaturalPoint, Inc.}},
  title = {{OptiTrack} Motion Capture Systems},
  year = {2023},
  url = {https://optitrack.com/},
  note = {Accessed: 2024}
}

@article{besl1992icp,
  author = {Besl, Paul J. and McKay, Neil D.},
  title = {A Method for Registration of 3-D Shapes},
  journal = {IEEE Transactions on Pattern Analysis and Machine Intelligence},
  volume = {14},
  number = {2},
  pages = {239--256},
  year = {1992},
  doi = {10.1109/34.121791}
}

@misc{cloudcompare,
  author = {Girardeau-Montaut, Daniel},
  title = {{CloudCompare} - 3D Point Cloud and Mesh Processing Software},
  year = {2023},
  url = {https://www.cloudcompare.org/},
  note = {Open Source Project}
}

@misc{grupp2017evo,
  author = {Grupp, Michael},
  title = {evo: {P}ython package for the evaluation of odometry and {SLAM}},
  year = {2017},
  url = {https://github.com/MichaelGrupp/evo},
  note = {Accessed: 2024}
}

@article{allan1966statistics,
  author = {Allan, David W.},
  title = {Statistics of Atomic Frequency Standards},
  journal = {Proceedings of the IEEE},
  volume = {54},
  number = {2},
  pages = {221--230},
  year = {1966},
  doi = {10.1109/PROC.1966.4634}
}

@misc{dynamixel2023,
  author = {{ROBOTIS}},
  title = {{Dynamixel XM540-W270-R} Servo Motor},
  year = {2023},
  url = {https://emanual.robotis.com/docs/en/dxl/x/xm540-w270/},
  note = {Accessed: 2024}
}

@misc{livox2020mid360,
  author = {{Livox Technology}},
  title = {{Livox MID-360} LiDAR Sensor},
  year = {2020},
  url = {https://www.livoxtech.com/mid-360},
  note = {Accessed: 2024}
}

@article{macenski2022ros2,
  author = {Macenski, Steven and Foote, Tully and Gerkey, Brian and Lalancette, Chris and Woodall, William},
  title = {Robot Operating System 2: Design, architecture, and uses in the wild},
  journal = {Science Robotics},
  volume = {7},
  number = {66},
  pages = {eabm6074},
  year = {2022},
  doi = {10.1126/scirobotics.abm6074}
}

@inproceedings{macenski2020nav2,
  author = {Macenski, Steve and Mart{\'\i}n, Francisco and White, Ruffin and Gin{\'e}s Clavero, Jonatan},
  title = {The Marathon 2: A Navigation System},
  booktitle = {2020 IEEE/RSJ International Conference on Intelligent Robots and Systems (IROS)},
  pages = {2718--2725},
  year = {2020},
  doi = {10.1109/IROS45743.2020.9341207}
}

@book{siegwart2011robotics,
  author = {Siegwart, Roland and Nourbakhsh, Illah R. and Scaramuzza, Davide},
  title = {Introduction to Autonomous Mobile Robots},
  edition = {2nd},
  publisher = {MIT Press},
  year = {2011},
  isbn = {978-0262015356}
}

@article{borenstein1996odometry,
  author = {Borenstein, Johann and Feng, Liqiang},
  title = {Measurement and Correction of Systematic Odometry Errors in Mobile Robots},
  journal = {IEEE Transactions on Robotics and Automation},
  volume = {12},
  number = {6},
  pages = {869--880},
  year = {1996},
  doi = {10.1109/70.544770}
}

@article{cai2021ikdtree,
  author = {Cai, Yixi and Xu, Wei and Zhang, Fu},
  title = {ikd-Tree: An Incremental KD Tree for Robotic Applications},
  journal = {arXiv preprint arXiv:2102.10808},
  year = {2021}
}

@inproceedings{koenig2004gazebo,
  author = {Koenig, Nathan and Howard, Andrew},
  title = {Design and Use Paradigms for Gazebo, an Open-Source Multi-Robot Simulator},
  booktitle = {2004 IEEE/RSJ International Conference on Intelligent Robots and Systems (IROS)},
  pages = {2149--2154},
  year = {2004},
  doi = {10.1109/IROS.2004.1389727}
}

@inproceedings{seraji1999traversability,
  author = {Seraji, Homayoun},
  title = {Traversability Index: A New Concept for Planetary Rovers},
  booktitle = {Proceedings of the 1999 IEEE International Conference on Robotics and Automation},
  pages = {2006--2013},
  year = {1999},
  organization = {IEEE},
  doi = {10.1109/ROBOT.1999.770402}
}

% --- Appendix ---
\appendix
%include anything you need in the appendix
%\include{tex/appendixA}

% --- Index ----
%\printindex

% --- that's it ---
\end{document}